\pdfoutput=1

\documentclass[11pt]{article}

\usepackage[preprint]{acl}

\usepackage{times}
\usepackage{latexsym}

\usepackage[T1]{fontenc}

\usepackage[utf8]{inputenc}

\usepackage{microtype}

\usepackage{inconsolata}

\usepackage{graphicx}

\usepackage{fancybox}

\usepackage[dvipsnames]{xcolor}
\usepackage{soul}

\usepackage{caption}
\usepackage{subcaption}

\usepackage{xurl}

\title{The QCET Taxonomy of \\Standard Quality Criterion Names and Definitions for\\the Evaluation of NLP Systems}

\author{Anya Belz, \hspace{.05cm} Simon Mille \and Craig Thomson\\
  DCU Natural Language Generation Research Group\\
  ADAPT, Dublin City University, Dublin, Ireland \\
  \texttt{anya.belz@dcu.ie}
}

\begin{document}

\newcommand{\qcetblue}[1]{\sethlcolor{RoyalBlue!20}\hl{#1}\sethlcolor{yellow}}
\newcommand{\qcetgreen}[1]{\sethlcolor{OliveGreen!20}\hl{#1}\sethlcolor{yellow}}
\newcommand{\qcetpurple}[1]{\sethlcolor{violet!20}\hl{#1}\sethlcolor{yellow}}
\newcommand{\qcetred}[1]{\sethlcolor{red!20}\hl{#1}\sethlcolor{yellow}}
\newcommand{\qcetyellow}[1]{\sethlcolor{Goldenrod!40}\hl{#1}\sethlcolor{yellow}}
\newcommand{\qcetorange}[1]{\sethlcolor{orange!30}\hl{#1}\sethlcolor{yellow}}
\newcommand{\qcetgrey}[1]{\sethlcolor{gray!20}\hl{#1}\sethlcolor{yellow}}
\newcommand{\qcetlightgrey}[1]{\sethlcolor{gray!10}\hl{#1}\sethlcolor{yellow}}

\maketitle
\begin{abstract}
Prior work has shown that two NLP evaluation experiments that report results for the same quality criterion \textit{name} (e.g.\ Fluency) do not necessarily evaluate the same aspect of quality, %
and the comparability implied by the name can be misleading. 
Not knowing when two evaluations are comparable in this sense means we currently lack the ability to draw reliable conclusions about system quality on the basis of multiple, independently conducted evaluations. %
This in turn hampers the ability of the field to progress scientifically as a whole, a pervasive issue in NLP since its beginning \cite{sparkjones:k:1981}. 
It is hard to see how the issue of unclear comparability can be fully addressed other than by the creation of a standard set of quality criterion names and definitions that the several hundred 
quality criterion names actually in use in the field can be mapped to, and grounded in. %
Taking a strictly descriptive approach, the QCET \textit{Q}uality \textit{C}riteria for \textit{E}valuation \textit{T}axonomy derives a standard set of quality criterion names and definitions from three surveys of evaluations reported in NLP, and structures them into a hierarchy where each parent node captures common aspects of its child nodes. %
We present QCET and the resources it consists of, and discuss its three main uses in (i) establishing comparability of existing evaluations, (ii) guiding the design of new evaluations, and (iii) assessing regulatory compliance.  %
\end{abstract}

\tableofcontents

\section{Introduction}\label{sec:intro} %

\citet{howcroft:etal:2020} and \citet{belz-etal-2020-disentangling} described a situation in human evaluation of language-generating systems where over 200 different quality criteria (QC) names were in use, definitions were patchy, and it was mostly impossible to tell from papers if the same aspect of quality was being assessed in different evaluations, resulting in unclear comparability and low repeatability \cite{cohen-etal-2018-thee-dimensions}. Consider the following examples which share the same QC name, {Fluency}, but each with a different definition (or closest thing to a definition provided):

\begin{enumerate}\itemsep=-0.05cm
    \item \citet{yu-etal-2020-review}: ``judging the question fluency.''
    \item \citet{van2020automatic}: ``grammatical and syntactically well-formed.''
    \item \citet{pan-etal-2020-semantic}: ``follows the grammar and accords with the correct logic."
\end{enumerate}

\noindent %
From the above, we can tell that the first two evaluations assessed single but different criteria (utterances can be grammatical but not very fluent), whereas the third assesses two criteria (an utterance can be grammatical yet illogical). In other words, none of the three assess the same QC. 

\citet{belz-etal-2020-disentangling} and \citet{howcroft:etal:2020} mapped the 200+ different QCs found in their 20Years survey to a set of 71 standard QC names and definitions designed to reflect all and only the true differences found between evaluations.  This reduction in QC numbers implies that two thirds of the original set did not reflect actual differences between aspects of quality assessed, merely a lack of standardisation in the naming of QCs. The three QCs in the above example were mapped to the following QCs from the 20Years standard inventory:

\begin{enumerate}\itemsep=-0.05cm
    \item \citet{yu-etal-2020-review}: Fluency $\rightarrow$ Fluency\footnote{Strictly speaking, in the case of Yu et al., it's unclear what was actually assessed, as we're given barely more than the QC name. If we could see the question/prompt put to evaluators we might arrive at a different conclusion about which standard QC is being assessed in the work.}
    \item \citet{van2020automatic}: Fluency $\rightarrow$ Grammaticality
    \item \citet{pan-etal-2020-semantic}: Fluency $\rightarrow$ Grammaticality $+$Coherence
\end{enumerate}

\noindent The 20Years mapping exercise revealed that it was very common for what was actually evaluated to be at odds with the name (or even the definition) of the quality criterion given in papers. For example, among the three definitions above, the second claims to be evaluating Fluency but actually evaluates Grammaticality. In this situation, not only is it problematic to report, say, that a system improves Fluency, but any comparisons with Fluency assessments of other systems are also unsafe. For human evaluations, the whole situation is compounded by the fact that very few details of evaluation experiments are usually reported beyond the quality criterion name \cite{belz-et-al-2023-missing,belz2023non,ruan-etal-2024-defining,schmidtova-etal-2024-automatic-metrics}.
As has been argued \cite{van-der-lee-etal-2019-best, howcroft:etal:2020,belz-etal-2020-disentangling, van2021human, gehrmann2023repairing}, this is an unacceptable state of affairs in particular for human evaluation which has always been regarded as the Litmus test of quality in NLP.

It is hard to see how the current misalignments between (i) what is actually evaluated vs.\ what it is named, and (ii) what different researchers mean by the same quality criterion name, can be addressed other than by a standard reference set of quality criteria names and definitions that those actually in use can be grounded in. Providing such a resource, in collaboration and exchange with the field, is the aim behind QCET. In the remainder of this section, we sketch out the way we approach this task (Section~\ref{sec:descr-stand-approach}); what previous contributions we build on (Section~\ref{sec:credits}); the resources we have created for the taxonomy (Section~\ref{sec:resources}); and how the rest of the paper is structured (Section~\ref{sec:paper-overview}).

\subsection{Descriptive standardisation approach}\label{sec:descr-stand-approach}

Our approach in all aspects of designing a standardised set of quality criteria, our philosophy if you like, is \textit{descriptive}. In other words, we take as our starting point the quality criteria that practitioners are already using in NLP, determine what is assessed in each case, group those together that assess the same thing (regardless of what they're called), and select a name (where possible, the original name used by the majority of practitioners) for each group. In this we follow the same steps that led to the 71 QC names for human evaluation of NLG in \citet{howcroft:etal:2020}.

The descriptive approach seeks to capture \textit{how people are doing something}, in contrast to the prescriptive approach which aims to set out \textit{how people should be doing it}.\footnote{The descriptive/prescriptive distinction is of course familiar from linguistics, in particular with regard to grammar.} The former is grounded in existing practice and has a strongly empirical flavour, whereas the latter is very hard to motivate on objective grounds, and would in the present context at best be based on theory outside of NLP, and at worst on individual intuition and opinion.

\subsection{Credits}\label{sec:credits}

The QCET taxonomy takes as a starting point the structure and set of 71 quality criteria that were created by the first two authors of this paper 
for the 20-year survey of human evaluations of NLG systems by \citet{howcroft:etal:2020}, with the top level distinction between correctness, goodness and feature-type QCs, as well as evaluation modes, as the basis for classifying quality criteria, introduced by \citet{belz-etal-2020-disentangling}.

Given the descriptive standardisation purpose explained in the preceding section, the QCET taxonomy is based on surveys of reported human evaluations in NLP/ML. More specifically, QCET is based on three surveys the first of which is the 20Years/NLG survey \cite{howcroft:etal:2020} whose first nine authors painstakingly analysed and annotated a systematically selected set of 165 papers with human evaluations of NLG systems published from 2000--2019. 

Taking the top-level classes from \citet{belz-etal-2020-disentangling} and the 71 standardised QCs from \citet{howcroft:etal:2020}, we created a new taxonomy with a more consistent internal structure, many more leaf and some more internal class nodes, and extended coverage (i) from NLG to NLP, and (ii) from human evaluation to both human and automatic evaluation.

During the extension of the taxonomy we added branches and leaf-nodes logically and based on our knowledge of the field, for each hypothesised node then finding attestations in the literature, and removing it if we couldn't find one. This resulted in an increase to 92 QC nodes.

To better cover automatic metrics and non-text generating NLP systems in QCET, we then carried out two additional surveys \cite{belz-et-al-2025-120papers} of  %
$2\times60$ randomly selected NLP papers published in ACL main proceedings over the last three years (2022--2024), collecting provided information for the following properties (in addition to bibliographic information) for every evaluation method found in the papers:

\begin{enumerate}\itemsep=0cm
    \item Metric vs.\ human evaluation, or `none found'
    \item Location in paper or elsewhere of evidence
    \item Verbatim QC name in paper, or `none given`
    \item Verbatim QC definition (or closest thing to it) in paper
    \item QCET node a QC belongs to, or `not found'
\end{enumerate}

\noindent Here our scope was all of NLP, not just NLG as in the 20Years survey
(see also Section~\ref{sec:nlp}). We identified altogether 455 individual occurrences of QCs. %
If QCET did not already cover a QC, we created a new QC node for it. Nearly all new QC names and definitions were derived from evaluation methods using automatic metrics. In this way, we increased the total number of QC terminal nodes in QCET to 111. The small number of new QCs found in these two surveys, combined with the fact that they were nearly all from metric evaluations, serves as a strong validation of the QCET taxonomy and its coverage of QCs in use in NLP.

In addition to extending taxonomic coverage, we extended and revised the information available at each node: (i) we revised and standardised all QC names and definitions; (ii) we added explanatory notes; (iii) we added example elicitation questions for use in human evaluations; and (iv) we added examples of usage (attestations) for each QC.

The QCET Taxonomy is compatible with the interactive taxonomy tool we created previously \citet{belz2024qcet} and was able to be plugged into it as is, supporting search and perusal of the current set of 111 QC nodes.

\subsection{QCET resources}\label{sec:resources}

The QCET \textit{Q}uality \textit{C}riteria for \textit{E}valuation \textit{T}axonomy, comprises the following resources (browser at \url{https://nlp-qcet.github.io/}; other items at \url{https://github.com/DCU-NLG/qcet_code}):

\vspace{-.1cm}
\begin{small}
\begin{enumerate}\itemsep=-0.05cm
    \item An interactive taxonomy browser;
    \item Extended description and usage guidance;
    \item At-a-glance diagram of taxonomy (see also in Figure~\ref{fig:diag}).
\end{enumerate}
\end{small}
\vspace{-.25cm}

\subsection{Paper overview}\label{sec:paper-overview}

The remainder of the paper is structured as follows. %
Section~\ref{sec:scope} describes the scope of the taxonomy. Section~\ref{sec:disentangling} discusses what quality criteria (QCs) are and how they can be identified in evaluations, as well as why it's important to treat them separately from evaluation modes and design. %

We then introduce the taxonomy structure and naming
conventions (Section~\ref{sec:tax-structure}), before describing each of the three main branching levels: frame-of-reference branches (Section~\ref{sec:for-branches}), type-of-quality branches (Section~\ref{ssec:type-of-quality-branches}), and aspect-of-quality branches (Section~\ref{sec:aspect-of-quality-branches}).

Next 
we discuss three examples of envisaged use of QCET, including worked examples 
(Section~\ref{sec:use}). We conclude with some discussion and a look to future work (Section~\ref{sec:concl}). In Appendix~A we list each of the 111 QC names and definitions in QCET, grouped together by main frame of reference: quality of outputs in their own right (Section~\ref{qcet:QO}), quality of outputs relative to inputs (Section~\ref{qcet:QI}), quality of outputs relative to in-distribution target outputs (Section~\ref{qcet:QT}), and quality of outputs relative to external frame of reference (Section~\ref{qcet:QE}).

\section{Scope}\label{sec:scope}

\subsection{Metric and human evaluation} 

As a taxonomy of quality criteria (QCs, which are distinct from evaluation mode and experiment design, see Section~\ref{sec:disentangling}), QCET is agnostic about what evaluation method is used to evaluate a QC, meaning it covers human evaluation methods and automatically computed metrics equally. The original set of 71 QCs from \citet{howcroft:etal:2020} was limited to human evaluations of NLG systems, and we have expanded QCs to include those more typically assessed by metrics (e.g.\ Perplexity, Similarity to Target Outputs) and in other subfields of NLP.

\subsection{NLP systems and components}\label{sec:nlp}

We use the term \textbf{NLP system} to refer to components (e.g.\ a dialogue system embedded in a flight booking system) or stand-alone applications (e.g.\ a machine translation system) that implement an NLP task. When we say `system quality' in the present context we mean NLP system quality, so if a system has non-NLP components, such as a user interface, back-end data management etc., these may need to be evaluated for quality criteria (such as user experience criteria) not included in the QCET Taxonomy.

To support this expansion, we have carried out two additional surveys each of 60 NLP papers published in ACL main proceedings over the last three years (2022--2024), collecting provided information for the following properties (in addition to bibliographic information) for every QC found in the paper:

\begin{enumerate}\itemsep=0cm
    \item Metric vs.\ human evaluation
    \item Location in paper or elsewhere of evidence for QC identification
    \item Verbatim QC name in paper
    \item Verbatim QC definition (or closest thing to it) in paper
    \item QCET node a QC belongs to, or `not found'
\end{enumerate}

\noindent If QCET did not already cover the QC, we created a new QC node for it. In this way, we identified and added 19 new QC names and definitions.

\subsection{Different output modalities}

While the 20Years survey \cite{howcroft:etal:2020} focused on NLP tasks with textual output, the QCET Taxonomy is agnostic about output modality, with the exception of certain individual leaf nodes (e.g.\ Spelling Accuracy is defined only for textual output). Other output modalities including speech, structured representations, labels, and data are thus covered too.

\subsection{Outputs and output sets/sequences}

In the present context we're dealing with evaluation of NLP systems. Because you can't take a system and assign a score or grade to it directly, this in reality means evaluating what might be termed \textit{manifestations of system behaviour}. The scope of QCET extends to the \textit{quality} of such manifestations, i.e.\ all QC nodes sit under the same top node, Quality. As Figure~\ref{fig:high-level-taxonomy} indicates, other evaluation criterion taxonomies could be constructed for Efficiency criteria (compute, memory, processing time, etc.), and Social, Legal and Ethical criteria (bias, adherence to regulations, responsible treatment of users, etc.).

When assessing quality, the manifestations of system behaviour of interest are (with possible edge-case exceptions) system outputs in the broadest sense, and user interactions with them. For simplicity, in this paper and other QCET resources, we use the term \textbf{output} to refer to (i) individual outputs (produced for individual inputs), and (ii) sets or sequences of outputs (produced for sets or sequences of inputs) that are evaluated as a whole (as e.g.\ in dialogue, interactive task completion, or question/answer generation).

\newcommand{\phhl}[1]{\sethlcolor{green!10}\hl{\textsf{\small #1}}\sethlcolor{yellow}}

\begin{table*}[ht]
    \centering\small
    \begin{tabular}{|p{0.9\textwidth}|}
    \hline
\textit{A better system... }\\
    \hline
    \hline
...is more \phhl{\$property}, where compared systems differ only in their outputs \\
    \hline
...produces \phhl{\$output\_type}s with fewer \phhl{\$error\_type}s \\
    \hline
...produces \phhl{\$output\_type}s with fewer \phhl{\$error\_type}s relative to the input\\
    \hline
...produces \phhl{\$output\_type}s that lack fewer of the \phhl{\$control\_attribute}s provided in the input \\
    \hline
...produces \phhl{\$output\_type}s with \phhl{\$\{more, less\}} \phhl{\$property} \phhl{*\$\{in their form, in their content/meaning, overall\}} \phhl{*\$additional\_constraint} \\
    \hline
...produces \phhl{\$output\_type}s that are \phhl{\$\{more, less\}} \phhl{\$property} \phhl{*\$\{in their form, in their content/meaning, overall\}} \phhl{*\$additional\_constraint} \\
    \hline
...produces \phhl{\$output\_type}s that are \phhl{\$\{more, less\}} \phhl{\$property} relative to the input \phhl{*\$additional\_constraint} \\
    \hline
...produces \phhl{\$output\_type}s of the input with fewer \phhl{\$error\_type}s \\
    \hline
...produces \phhl{\$output\_type}s that \phhl{*\$\{in their form, in their content/meaning, overall\}} less often differ from the given target \phhl{\$output\_type}s \phhl{*\$additional\_constraint} \\
    \hline
...produces \phhl{\$output\_type}s that are \phhl{*\$\{in their form, in their content/meaning, overall\}} more at the target level of \phhl{\$property} provided in the input \\
    \hline
...produces \phhl{\$output\_type}s that are \phhl{*\$\{in their form, in their content/meaning, overall\}} more similar to given target \phhl{\$output\_type}s \phhl{*\$additional\_constraint} \\
    \hline
...produces \phhl{\$output\_type}s that are \phhl{*\$\{in their form, in their content/meaning, overall\}} either (a) more \phhl{\$property}, or (b) less \phhl{\$property} \\
    \hline
...produces \phhl{\$output\_type}s that are \phhl{*\$\{in their form, in their content/meaning, overall\}} (a) more \phhl{\$property}, (b) less \phhl{\$property}, or (c) more at the target level of \phhl{\$property} provided in the input \phhl{*\$additional\_constraint} \\
    \hline
...produces \phhl{\$output\_type}s that \phhl{\$affect} \phhl{\$object}s (a) more, (b) less, or (c) as specified in the input, in terms of a given range of possible \phhl{\$object} values \\
    \hline
...produces \phhl{\$output\_type}s that make the entity producing the text come across to an observer as \phhl{\$property} (a) more, (b) less, or (c) to the degree specified in the input \\
    \hline
...produces \phhl{\$output\_type}s that result in \phhl{\$output\_type}s with \phhl{\$\{more, less\}} \phhl{\$property} in a given  \phhl{\$evaluation\_type} \\
\hline
    \end{tabular}
    \caption{The 16 QC definition templates used in QCET. Each template starts with the words \textit{A better system...} which we omit in each template for brevity.}
    \label{tab:definition_templates}
\end{table*}

\section{Standardisation and Wording Conventions for Quality Criteria}\label{sec:disentangling}

QCET is based on the notion of quality criterion, i.e.\ the specific aspect of system quality that is assessed in an evaluation. This is the basic
level at which we would expect two high-quality evaluations (that assess the same QC) to support the same conclusions about which of two systems is better.  We would not expect implementational details, operationalisation, whether we're assessing one or two systems at a time, etc., to affect such coarse-grained conclusions, and this is what makes the assessed aspect of quality the right level for comparability grounding.

However, it is often not straightforward to look at an evaluation and immediately identify the specific aspect of quality it assesses. For example, compare the following descriptions of two evaluations:

\vspace{-.15cm}
\begin{quote}
    ``[...] eight users were given flight reservation tasks that required them to access the airline schedule database [...]. The system logged the total completion time of a dialogue (Total Completion Time) [...]." %
    \cite{qu-green-2002-constraint}
    
    ``[we] compare [search] task completion times for two search algorithms [...] we had a number of paid participants describe a difficult [search] task which they had recently attempted. [...] Once 100 tasks were obtained in this manner, a separate group of 200 paid participants [acted] as users to attempt these tasks. [...] The resulting task times [...]'' \cite{xu2009evaluating}
\vspace{-.15cm}

\end{quote}

\noindent There are considerable differences between the two evaluation experiments: 1 system and 8 users in Q\&G vs.\ 2 systems and 200 users in X\&M; researcher-composed narrow tasks in Q\&G vs.\ user-generated broad tasks in X\&M; interaction via dialogue with the system in Q\&G vs.\ key-word entry and clicking on returned search results in X\&M, to name a few. However, none of these differences change the aspect of system quality, i.e.\ the quality criterion, that is assessed. %

To support such assessments, we decompose evaluation methods $M$ into QC, \textbf{evaluation modes} and \textbf{experiment design} which relate as follows:

\vspace{.2cm}
\setlength\tabcolsep{3.6pt} %
\renewcommand{\arraystretch}{1.2} %
\noindent\begin{tabular}{ll|p{6.5cm}|}
   \cline{3-3}
   && Quality criterion + evaluation mode = \textbf{evaluation measure}; \\
   && Evaluation measure + experiment design = \textbf{evaluation method}. \\
   \cline{3-3}
\end{tabular}%
\vspace{.15cm}

\noindent \textbf{Evaluation mode} \cite{belz-etal-2020-disentangling} has three dimensions: \textbf{extrinsic} (impact on something external to the system is assessed) vs.\ \textbf{intrinsic} (otherwise); \textbf{objective} (repeated measurements yield the same results to within some margin of precision) vs.\ \textbf{objective} (otherwise); and \textbf{absolute} (one system evaluated at a time), vs.\ \textbf{relative} (more than one system evaluated at a time).

In order to disentangle the QC, we need to peel away everything to do with evaluation mode, experiment design and implementation, to find the answer to the following question: 

\vspace{-.15cm}
\begin{quote}
\textit{Q: What does this evaluation consider a better %
system to be?} 
\end{quote}
\vspace{-.15cm}

\noindent In the above evaluations, a better system is not one that is found to be better in an experiment with 8 users that access an airline schedule database to book a specific flight while talking to the system and take less time to do it. Whether there are 8 users or a different number, where they get hold of the flight information, etc., is not part of system quality, but of experiment design or system implementation. 
~
~
Once we disregard such factors, we are  left with the following answer to the above question:

\begin{quote}
     \textit{A: A better system is one that enables the user to complete a task more quickly.}
\end{quote}

\noindent At this level we can see that both evaluations above assess the same aspect of quality. We would expect two well-designed evaluations that assess the same pair of systems, using the same task and data, in terms of this aspect of quality, to come to the same conclusion about which is better.\footnote{Or more precisely, we would expect a majority of evaluations of this type to come to the same conclusion, because each is associated with a probability of wrong results.}

The above process is how we derive QC definitions, as explained in more detail next (re wording of definitions see Section~\ref{sec:word-conv-qc-definitions}). %

\subsection{Wording conventions for QC names}\label{sec:word-conv-qc-names}

For QC names, we follow the convention that we name QCs after the `good end' of the corresponding quality spectrum, or both ends where there isn't one (see also Section~\ref{sec:tax-structure}),  aiming to also closely reflect the definition. %
E.g.\ in the context of the above example, `task completion time' is overly specific and quality-neutral. Instead, we choose the QC name Task Completion Speed.

Moreover, QC names need to be both unique and self-explanatory as far as reasonably possible. However, the QCET node ID should be considered part of the name as it traces which type of quality and which aspect of the output a QC assesses (Section~\ref{sec:node-ids-names}).

If the above conventions are met and there is no risk of confusion with other QCs, we then use the most commonly used, clear QC name; e.g.\ Grammaticality, Coherence. 

Where needed for clarity, we explicitly refer to what a QC compares outputs against in the QC name, e.g.: Absence of Omissions (relative to input). In such contexts, we call the gold-standard reference outputs often used as comparators in evaluation, \textbf{target outputs}; and any other comparators (e.g.\ compared against for similarity of style), \textbf{non-target references}. 
Finally, QC names are title-cased, except for clarifying additions in parentheses which are in lower case. 

Occasionally, the convention that the QC name should reflect what the QC name considers the good end of the scale leads to an awkward QC name; e.g.\ Absence of Omissions, Low Perplexity. 
Striving for clarity has resulted in some inconveniently long QC names, one of the longer ones being Performance of an embedding\footnote{An embedding system here is one that contains the component under evaluation.} or downstream system/component.

\subsection{Wording conventions for QC definitions}\label{sec:word-conv-qc-definitions}

All QC definitions are worded in terms of what makes a better system (i.e.\ as an answer to the question at the start of this section). This is in part because it provides a good starting place for evaluation design, pinpointing both what are desirable properties in the system under development, and how developers know they're making progress. This facilitates comparisons between systems as well as evaluation of a single system against a baseline or its previous versions. %
E.g.\ in terms of Speed of Understanding, \textit{a better system produces outputs that are faster to understand}.

An important principle is that we avoid introducing words and concepts into definitions that are used in the names of other QC criteria. E.g.\ Fluency is defined simply as \textit{A better system is one that produces more fluent output}, rather than using words like grammatical or readable. %

We use the term \textbf{outputs} where a QC can potentially apply to any \textbf{modality}, but use a narrower term, e.g.\ speech or text, if the QC is applicable to a more constrained set of outputs.

For uniformity, we use definition templates as shown in Table~\ref{tab:definition_templates}, all starting with \textit{A better system...}. Note that in some cases the templates produce definitions that are ungrammatical or structurally too complex for readability, in which case we minimally adjust the wording to address the issue(s) without affecting the meaning.

\begin{figure*}[ht]
    \centering
    \begin{subfigure}[b]{0.5\textwidth}
         \includegraphics[width=0.95\linewidth,trim={3.1cm 3cm 6cm 1cm}, clip]{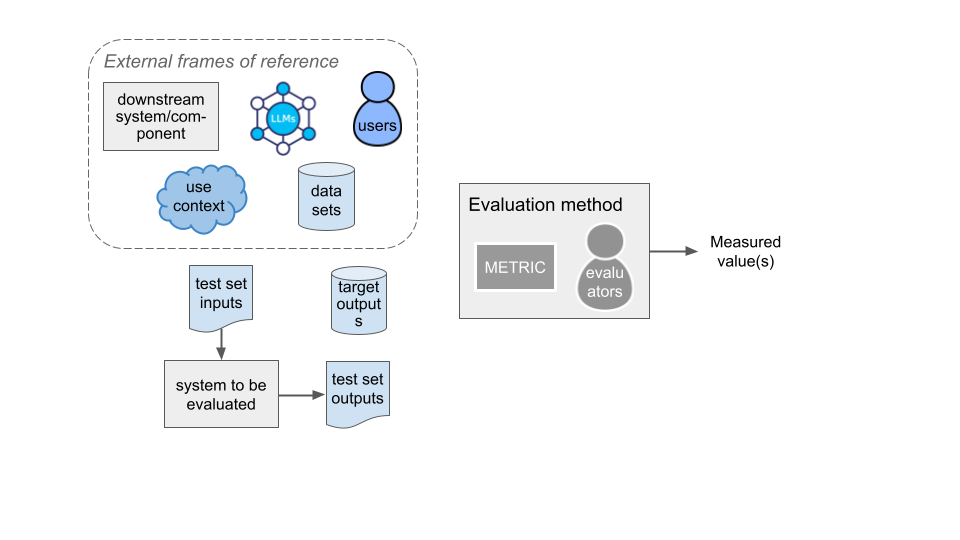} 
         \caption{Possible inputs to evaluation methods.}\label{sfig:template}
    \end{subfigure}%
    \begin{subfigure}[b]{0.5\textwidth}
    \begin{tabular}{cc}
    \includegraphics[width=0.5\linewidth,trim={3.1cm 3cm 6cm 1cm}, clip]{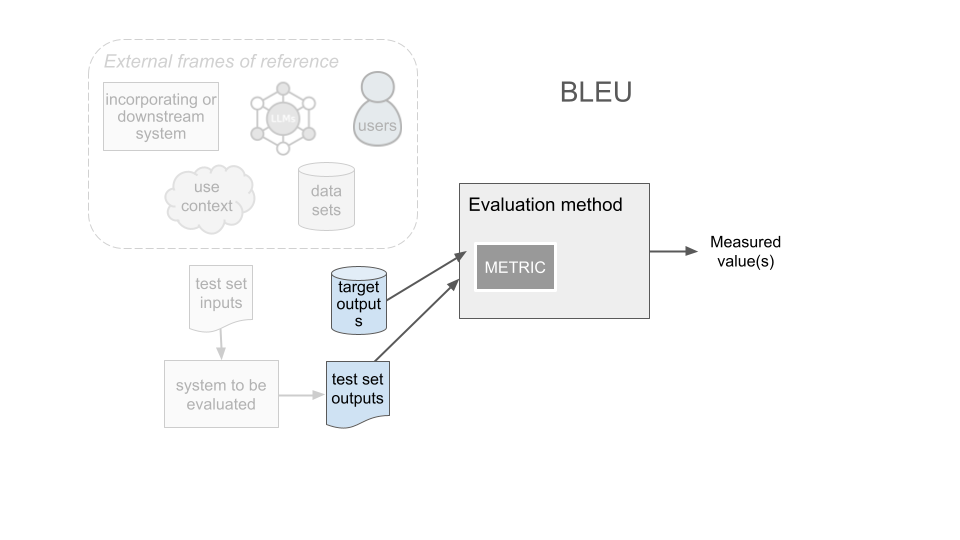} & \includegraphics[width=0.5\linewidth,trim={3.1cm 3cm 6cm 1cm}, clip]{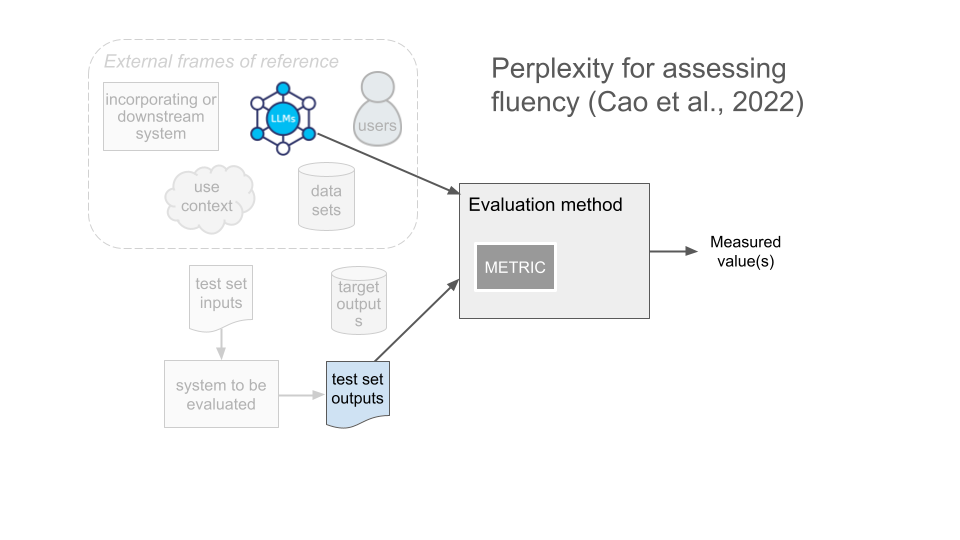} \\
    \includegraphics[width=0.5\linewidth,trim={3.1cm 3cm 6cm 1cm}, clip]{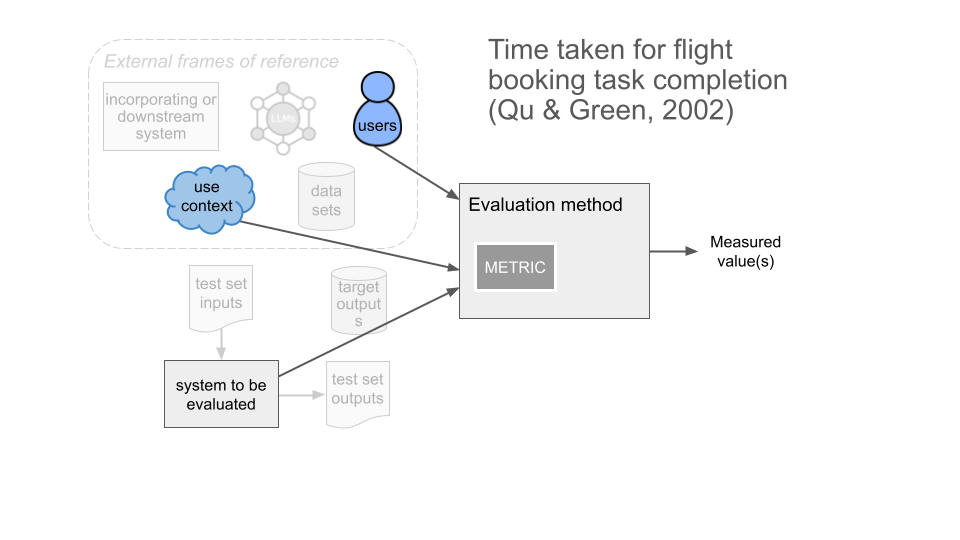} & \includegraphics[width=0.5\linewidth,trim={3.1cm 3cm 6cm 1cm}, clip]{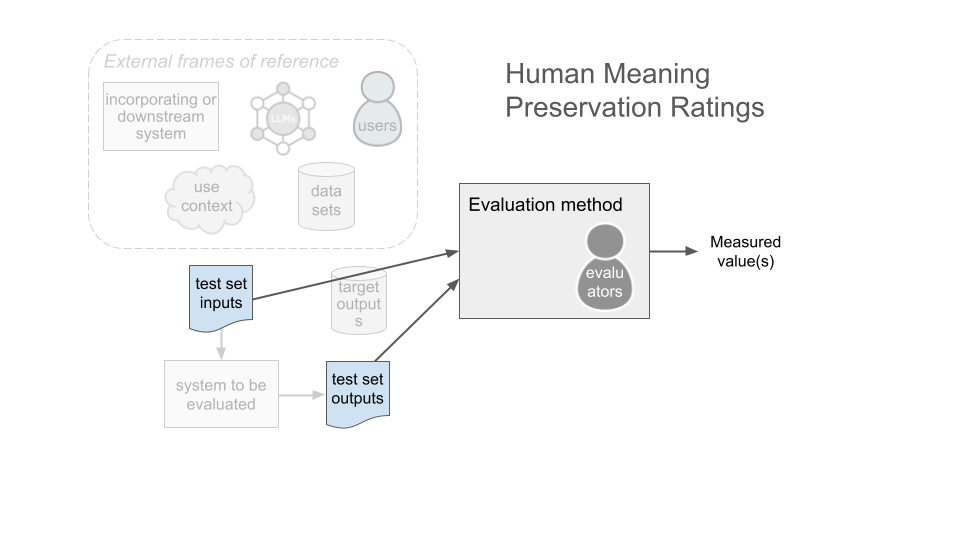} \\
    \end{tabular}
    \caption{Four example evaluation methods and their inputs.}
    \end{subfigure}
    \caption{Left: Overview of different external frames of reference, the system inputs, system, and system output that are typically used, in different combinations, in evaluation methods. Right: Individual example evaluation methods and their inputs superimposed on the overview.}
    \label{fig:what-goes-into-eval-methods}
\end{figure*}

\subsection{Wording conventions for QC elicition prompts}\label{sec:word-conv-qc-elicitation}

At each node we provide example elicitation questions that can be used in a human evaluation that assesses the quality criterion. There are six different combinations of evaluation modes (start of Section~\ref{sec:disentangling}) each of which would call for a different operationalisation of the QC. We provide questions for two of the most common combinations (see \citeauthor{belz-etal-2020-disentangling}, \citeyear{belz-etal-2020-disentangling} for details of evaluation modes), as in the following example for Spelling Accuracy:

\begin{enumerate}
    \item \textit{Intrinsic, subjective, relative}: Which of these texts has fewest spelling errors, looking at their form only and ignoring their content/meaning?
    \item \textit{Intrinsic, subjective, absolute}: To what degree is this text free of spelling errors, looking at its form only and ignoring its content/meaning?
\end{enumerate}

\noindent We use the same two templates for all QC nodes, here too making minimal adjustments for readability: 
\begin{enumerate}
    \item \textit{Which of these texts [has, is] [description of property], looking at their [form only, meaning only, form and meaning both] [and ignoring their [form, content/meaning]]?}
    \item \textit{To what degree [does, is] this text [description of property], looking at its [form only, meaning only, form and meaning both] [and ignoring its [form, content/meaning]]?}
\end{enumerate}

\noindent Note that Feature QCs have two versions of each prompt type, e.g.\ for Diversity/Non-diveristy, one that asks which output is more diverse, and one that asks which is less diverse, depending on which end of the scale is the `good' one in a given evaluation context. In the taxonomy, as in the examples in the list of QC leaf nodes in Appendix~\ref{sec:list-qcs}, however, we only show the `more' option.

\section{Overall Taxonomy Structure and Representational Conventions}\label{sec:tax-structure}

\begin{figure*}[ht]
    \centering
    \includegraphics[width=\textwidth]{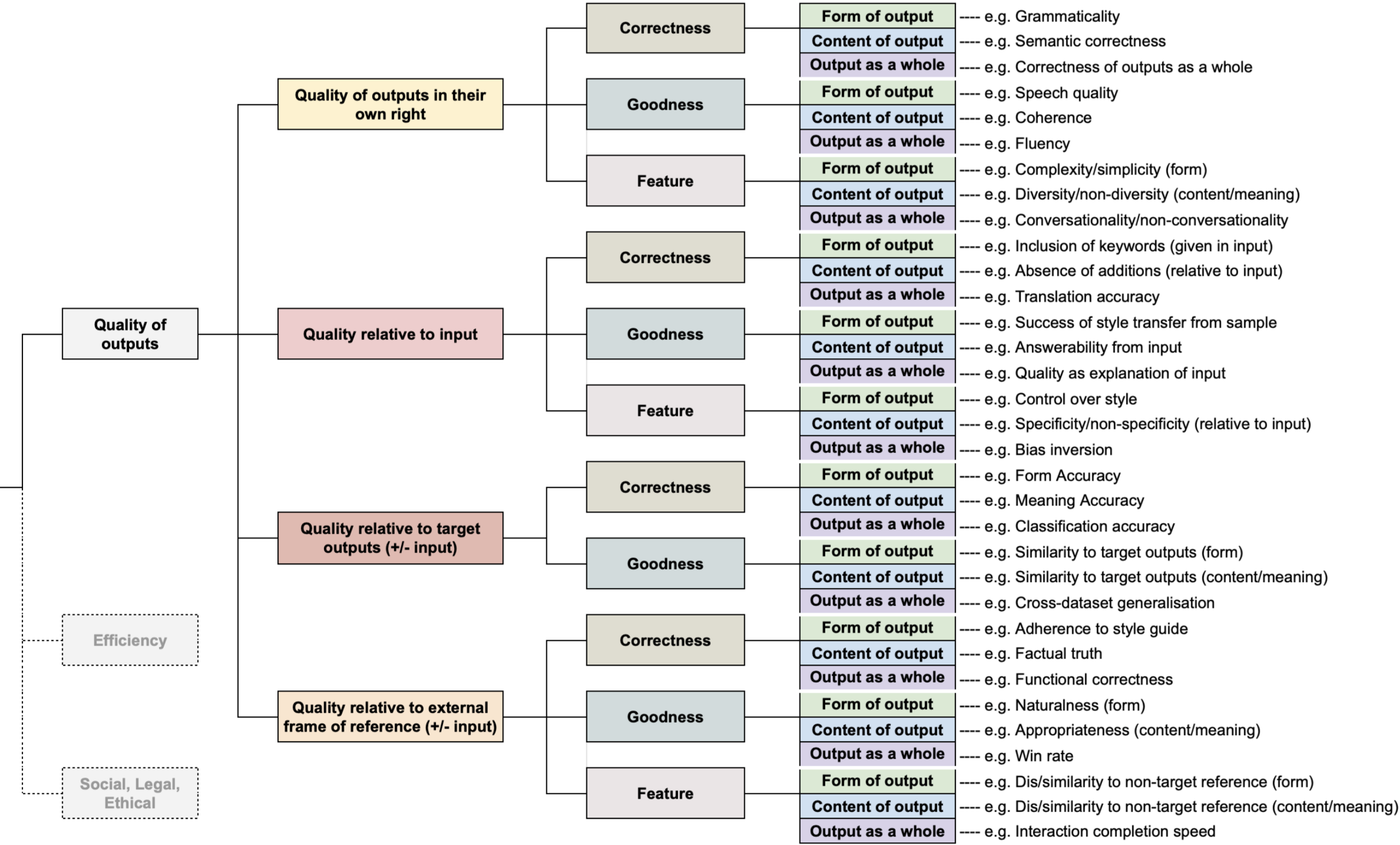}
    \caption{High-level view of QCET taxonomy structure with example leaf nodes.}
    \label{fig:high-level-taxonomy}
\end{figure*}

Figure~\ref{fig:high-level-taxonomy} provides a high-level view of the structure of the QCET taxonomy. The second, third and fourth levels (after the root) correspond to the main three branching factors frame of reference (Section~\ref{sec:for-branches}), type of quality (Section~\ref{ssec:type-of-quality-branches}), and aspect of quality (Section~\ref{sec:aspect-of-quality-branches}). Below these in the actual taxonomy are levels of specific quality criteria (for space reasons not shown in the figure, except for single, abbreviated examples), mostly at the one terminal-node level shown in the figure. However, in a small number of cases there are more specific QCs sitting on two further levels. All leaf nodes can be seen in Figure~\ref{fig:diag} in the Appendix.
We start below with conventions for node IDs and content, before describing each of the three branching factors.

\subsection{Node IDs and content}\label{sec:node-ids-names}

Figure~\ref{fig:high-level-taxonomy} shows just the quality criteria names at each node for a simple high-level view. In the actual taxonomy, each node additionally has (i) a definition of the QC, (ii) attestations in the literature, (iii) explanatory notes, and (iv) example questions that can be put to evaluators for the example evaluation modes absolute and relative (in both cases combined with intrinsic and subjective, by far the most common combination in NLP research).

Each node has a \textbf{node ID} which traces its path from the root via the first or other significant letters in the intervening node labels. E.g.\ Grammaticality (seen in the top right corner in Figure~\ref{fig:high-level-taxonomy}) has the ID QOC-f-1 (\textit{Q}uality, \textit{O}utput in its own right, \textit{C}orrectness, \textit{f}orm only, leaf node \textit{1}).

In the remainder of this paper, we use node ID plus node name to refer to QCs. Because these can be quite long, we use light grey highlight to indicate the span of the node ID/name, as in \qcetgrey{[QOC-f-1] Grammaticality}. Note that all nodes represent a subclass and most also represent a QC; when we refer to a node as a subclass we don't use the ID or the grey highlighting, instead boldfacing the node name. Some nodes only function as subclasses and not as QCs in their own right. E.g.\ this is the case for all internal nodes in \textbf{Features} subclasses. When we include the ID for such nodes, we mark it with an asterisk to indicate that it doesn't function as a QC.

\subsection{Frame-of-reference branches}\label{sec:for-branches}

The root node corresponds to the single most general QC class, {\textbf{Quality of outputs}}.  The next taxonomic level relates to the \textbf{Frame of Reference} if any that is used in an evaluation, and marks differences relating to what a QC is defined relative to. The distinction is between quality in a QC being defined relative to (i) just the \textbf{outputs}; (ii) outputs and \textbf{inputs} (only); (iii) outputs and \textbf{target outputs} drawn from the same distribution as the training data, plus optionally inputs as well; or (iv) outputs and an external \textbf{frame of reference}, plus optionally inputs as well. 

For example, in order to assess Fluency with human raters, we don't need to look at anything other than a sample of system outputs; to assess Meaning Preservation e.g.\ in paraphrasing, we need to look only at a sample of system inputs and outputs; to assess Similarity to Target Outputs e.g.\ with BLEU, we only need a sample of outputs and corresponding target outputs; and in order to assess Task Completion Speed by users of the system, we need to get users to interact with the system and complete a specified task (users and use context form part of the system-external frame of reference that the QC is defined relative to).

Figure~\ref{sfig:template} provides a diagrammatic overview of these possible inputs to evaluation methods, divided into external frames of reference in the dashed box, and the system, and samples of inputs, outputs and target outputs at the bottom left. By looking at which of these a given evaluation method uses, we can tell which of the frame-of-reference (FoR) branches of the taxonomy the QC being evaluated in the evaluation method belongs to:

\vspace{-.15cm}%
\begin{enumerate}\itemsep=-0.05cm
    \item \textbf{Output in its own right}: QCs that are defined relative to just the output (capture the quality of the output in its own right); an evaluation method of this type uses only the system outputs, as in e.g.\ Human Fluency Ratings. I.e.\ the measured value produced by this type of evaluation method can be obtained on the basis of the system outputs alone.
    \item \textbf{Output relative to input}: QCs that are defined relative to both and only the output and the input (capture the quality of the output relative to the input and nothing else); an evaluation method of this type uses only the system inputs and outputs, as in e.g.\ Meaning Preservation Ratings in Figure~\ref{fig:what-goes-into-eval-methods}.
    \item \textbf{Output relative to in-distribution target outputs (+/-- input)}: QCs that are defined relative to target outputs sampled from the same distribution as the training data, and optionally also the input (capture the quality of the output {relative to given target outputs}, optionally also taking the input into account). E.g.\ BLEU in Figure~\ref{fig:what-goes-into-eval-methods} uses (just) system outputs and target outputs. 
    \item \textbf{Output relative to a system-external frame of reference (+/-- input)}: QCs that are defined relative to a system-external frame of reference, and optionally also the input (capture the quality of the output {relative to an external frame of reference}, optionally also taking the output and/or input into account). An evaluation method of this type uses an explicit external frame of reference, as is the case when measuring time taken for task completion (here the FoR is the user interacting with the system), and Perplexity (here the FoR is the language model used for computing perplexity) in Figure~\ref{fig:what-goes-into-eval-methods}, both of which also take test set outputs into account. 
\end{enumerate}%
\vspace{-.15cm}

\subsection{Type-of-quality branches}\label{ssec:type-of-quality-branches}

The next level down has the three QC subclasses {{Correctness}}, {{Goodness}}, and {{Features}} \cite{belz-etal-2020-disentangling}. {\textbf{Correctness}} and {\textbf{Goodness}} QCs align with (i) what we train systems to be good at, and/or (ii) common-sense notions of desirable properties in NLP systems. For example, \qcetgrey{[QEG-w-6] User Satisfaction as Affected by Outputs} and \qcetgrey{[QIC-w-1] Translation Accuracy} refer to clearly desirable properties in this sense: it is hard to conceive of circumstances under which we would want to consider a system better if it has \textit{lower} user satisfaction or \textit{lower} translation accuracy. Another way to look at it is that there is a preferred end to the quality scale independently of evaluation context.

This is not the case in the {\textbf{Features}} subclass where either end of the scale might be the preferred end depending on evaluation context. For example, in the case of \qcetgrey{[QOF-w-3] Complexity/Non-complexity (outputs as a whole)}, sometimes better systems are those with simpler outputs (e.g.\ text simplification; 
\citet{angrosh-siddharthan-2014-text})
, and sometimes those with less simple outputs (e.g.\ story generation; \citeauthor{purdy2019reading}, \citeyear{purdy2019reading}). Or, alternatively, there is no notion of either end of the scale being the better one, and a better system is one that is better at generating texts of a required level of simplicity/complexity (e.g.\ graph summarisation at different reading levels; \citeauthor{moraes-etal-2016-enabling}, \citeyear{moraes-etal-2016-enabling}).

More generally, the three QC types are defined as follows:

\vspace{-.125cm}%
\begin{enumerate}\itemsep=-0.05cm
    \item 
    \textbf{Correctness}: Correctness type QCs are based on countable errors or deviations (e.g.\ spelling errors in \qcetgrey{[QOC-f-2] Spelling Accuracy}). This makes it possible to state formally and precisely the conditions under which an output is of maximal quality, namely when the output does not contain any of a finite number of known errors. %
    
    \item 
    \textbf{Goodness}: For goodness type QCs, it is not normally possible to state, in the general case, the conditions under which an output is of maximal quality. %
    E.g.\ for \qcetgrey{[QOG-w-3] Fluency}, even if an output contains no disfluencies as such, there may be other ways to improve its fluency, e.g.\ using a different word, or changing the syntactic structure. %
    Goodness type QCs are not primarily based on notions of countable errors, typically taking multiple factors into account without naming or distinguishing them. E.g.\ in \qcetgrey{[QOG-w-4] Humanlikeness}, many different factors will typically play into what makes an output more human-like. %
    
    \item %
    \textbf{Features}: For a feature-type QC {+X/--X}, outputs are not generally better either if they are more +X, or if they are more {--X}. Depending on evaluation context, either more {+X} may be better, or more --X may be better, worse, or neither is associated with a notion of better/worse. %
    E.g.\ in the case of \qcetgrey{[QEF-w-3] Effect on User Emotion}, a better system produces outputs that affect the user's emotions (a) more, (b) less, or (c) as specified in the input, in terms of a given range of possible emotions. %

    The \textbf{Features} subclass is the most open class of the three: any +X/--X property of outputs or systems can in principle be added as a leaf node to one of the \textbf{Features} subclasses. In contrast to Correctness/Goodness criteria, Feature-type criteria are often evaluated where systems are designed to have a given feature, e.g.\ via a control attribute.
\end{enumerate}%
\vspace{-.1cm}

\subsection{Aspect-of-quality branches}\label{sec:aspect-of-quality-branches}

\noindent At the next level again, classes further split into the below three subclasses capturing which aspect of an output is being assessed:

\vspace{-.15cm}%
\begin{enumerate}\itemsep=-0.05cm
    \item \textbf{Form}: The form of outputs alone is assessed, e.g.\ \qcetgrey{[QOC-f-1] Grammaticality} -- a sentence can be grammatical (form) yet wrong or nonsensical in terms of content. 
    \item \textbf{Content}: The content/meaning of outputs alone is assessed, e.g.\ \qcetgrey{[QIC-c-4] Coverage of Topics (given in input)} -- two sentences can have the same meaning, but differ in form.
    \item \textbf{Outputs as a whole}: Outputs are assessed not distinguishing form from content. E.g.\ \qcetgrey{[QOG-w-5.1] Clarity} is a property of outputs as a whole, either form or meaning can detract from it.
\end{enumerate}%
\vspace{-.15cm}

\noindent Except for a small number of edge cases, we have found it straightforward to distinguish \textbf{Form} QCs from \textbf{Content} QCs. The former refers to how something is said, whereas the latter refers to what is said. Style, level of formality, choice between near-synonyms, syntax, word order, typography, etc.\ are all part of \textbf{Form}. Sentiment, topic, factual truth, entailment, consistency, coherence, etc.\ are all part of \textbf{Content}.  However, when comparing the meaning of two representations (inputs, outputs,  references), the line between Form and Content has to be drawn explicitly. In the case of metrics, this is contained in the algorithmic definition of the metric, while in human evaluation, evaluators need to be instructed e.g.\ at what level of sameness of meaning an output can be said to preserve the meaning of the input: how close do two synonyms have to be to be taken to mean the same thing, does past tense vs.\ present make a difference, the order in which content is presented, etc.

\section{Use}\label{sec:use}

In this section, we look at the three main envisaged uses of the QCET taxonomy, (i) establishing comparability of existing evaluations, (ii) guiding the design of new evaluations, and (iii) assessing regulation compliance. For each, we discuss the steps involved, and present a worked example.

\subsection{Identifying the QC in an existing experiment and mapping it to the right QCET node}\label{sec:ident-qcs}

The first step is to locate all resources shared about a given experiment, then to identify (i) QC name, (ii) QC definition and (iii) the question and/or instructions put to evaluators. In many cases, \textit{i--iii} are not completely aligned in which case \textit{iii} takes priority as expressing what was actually evaluated. 

A complicating factor is that in the effort of explaining one QC, researchers often introduce terms associated with other QCs, e.g.\ in the second Fluency definition at the start of this paper, Fluency was explained (only) in terms of grammaticality which introduces another QC. The third definition in the same example introduced two other QCs. We would argue that in the former case, one QC is being evaluated (Grammaticality), and in the latter case two (Grammaticality and Logicality). Neither assesses Fluency.

To arrive at such conclusions, the taxonomy is perused via the QCET tool (Section~\ref{sec:resources}) top down, armed with the information in \textit{i--iii} above, until the correct node(s) is/are reached (corresponding to the specific individual quality criterion in question).

\paragraph{\textit{Worked example}:} \citet{novikova-etal-2018-rankme} collect human ratings for three data-to-text generation systems, including for a quality criterion they call Naturalness. A definition is not provided, but the question put to evaluators is given in Section~5: ``Could the utterance have been produced by a native speaker?'' Data input and two text outputs (utterances) were shown to evaluators side by side; evaluators rated each utterance on a 6-point scale. 

While the input and an output from another system are shown to evaluators in addition to the output being evaluated, the elicitation prompt does not refer to them, hence the QC cannot be said to be explicitly evaluated relative to either of them.\footnote{Evaluation outcomes may nevertheless be affected by their presence.} This means we are in the QCET branch \textbf{Quality of outputs in their own right}. 

At the next level down, we can see that we're not in the Correctness branch: the question does not refer to errors or correctness, and we couldn't state the conditions in the general case under which an utterance would be maximally native-speaker-like. This is a scale with just one good end, so it's not the Features branch either. This leaves us with the \textbf{Goodness} branch of QCET.

The elicitation prompt does not refer to either the form or the meaning of the utterance, nor is native-speaker-likeness associated exclusively with one or the other, so we go down the \textbf{Output as a whole} branch. 

Looking at Figure~\ref{fig:diag},
we can see \qcetgrey{[QOG-w-4] Humanlikeness} in this branch which fits except that the present elicitation prompt is more specific: it's not just about the utterance being like one produced by a human, but more specifically a human who is a native speaker of the language. This then gets us to the final leaf node which we chose as the QCET node to map to: \qcetgrey{[QOG-w-4.1] Native Speaker Likeness}.

Note that if we had done a keyword search of the taxonomy with the QC name used in the paper (Naturalness) as the keyword, we would have ended up in the following branch: \qcetgrey{[QE] Quality of outputs relative to a specified external frame of reference} $\rightarrow$ \qcetgrey{[QEG] GOODNESS} $\rightarrow$ \qcetgrey{[QEG-w] Outputs as a whole} $\rightarrow$ \qcetgrey{[QEG-w-1] Naturalness (outputs as a whole)}. However, that would have been wrong, because there is no external frame of reference involved in the question asked of evaluators in the \citeauthor{novikova-etal-2018-rankme} evaluation; nor does the question reference naturalness in any way. So on the basis that evaluators will evaluate by answering the question in the elicitation prompt, rather than looking at the name of the criterion (often not shown at individual points of evaluation) and ignoring the question, \qcetgrey{[QOG-w-4.1] Native Speaker Likeness} is the correct QCET node to map to.\footnote{Actual assessment of Naturalness needs a system-external frame of reference, because to decide if an utterance is a natural thing to say, you need to know \textit{in what context}; QCET has three versions of this, for form, content/meaning and utterances as a whole. E.g.\ \textit{Is your bank account overdrawn} is perfectly native-speaker-like, but not a natural utterance while giving cooking instructions, or when talking to 6-year-olds.}

\subsection{Selecting quality criteria for a new evaluation}\label{sec:select-qcs}

A good starting point for evaluation in system development (in both academic and industry contexts) is the following question: \textit{suppose we have two candidate systems, how do we know which one produces better outputs?}  A useful answer is unlikely to be \textit{the one with the higher BLEU score,} because that just means its outputs are more similar to the given sample of target outpus. Instead, the answer is likely to involve multiple dimensions of quality.

\paragraph{\textit{Worked example:}} Let's take accommodation highlights generation in an accommodation booking context as an example \cite{kamath-etal-2024-generating-hotel}. A company planning to develop such a system might  decide that the better accommodation description summariser of two candidate systems would produce summaries that have \textit{better quality of language}, \textit{cover the main selling points}, contain \textit{fewer mistakes or misrepresentations}, while also being closer \textit{to some given target length range}. These broad dimensions each correspond to one or more specific QCET \textit{branches}, but can each be covered by multiple QCET \textit{nodes} within those branches (except for output length), as explored further in the paragraphs below.

\vspace{0.15cm}\noindent\textit{Better quality of language:} Following the levels of the QCET taxonomy top down, the first question is, do we think that language quality in this context can be assessed by considering just the outputs? Or is it conditional on the inputs? Or should we assess it using an external frame of reference? The fourth option, comparison with target outputs, is not suitable for language quality (only) assessment as it will assess more than language quality, including e.g.\ the content/meaning. \citeauthor{kamath-etal-2024-generating-hotel}\ went for \qcetgrey{[QO] Quality of outputs in their own right}. 
    
Next, is it Correctness, Goodness, or a Feature we should assess?  \citeauthor{kamath-etal-2024-generating-hotel}\ went for Correctness, then Form and then \qcetgrey{[QOC-f-1] Grammaticality}. However, Goodness is also an option and offers QCs that have a more holistic view of language quality. One could argue %
that it is the output as a whole that should be assessed for language quality, leading us down the \qcetgrey{[QOG] GOODNESS} $\rightarrow$ \qcetgrey{[QOG-w] Goodness of outputs in their own right, Outputs as a whole} branch. %
Down this branch, we find \qcetgrey{[QOG-w-4] Humanlikeness} which seems a good choice in this context (other options exist and, depending on the system developers' view of quality, could be chosen instead).
    
\vspace{0.15cm}\noindent\textit{Cover the main selling points better:} This is a less fine-grained system requirement than the others, and needs refining further before identifying QCs. Suppose we have two candidate highlights summaries, how do we decide which covers the main selling points better? There are several distinct dimensions to this: (i) the highlights have to capture all the most important selling points, (ii) they shouldn't contain anything that is in the input but isn't a selling point (e.g.\ long distance from city centre), and (iii) they have to express the selling points clearly. For \textit{iii}, we could assess the overall clarity of the text (\qcetgrey{[QOG-w-5.1] Clarity}), and that is what \citeauthor{kamath-etal-2024-generating-hotel} did.

For \textit{i}, we need both inputs (the long-form accommodation descriptions) and outputs (the highlights), we're in the Goodness branch (we're not identifying errors or deviations, and there is a single good end to the quality spectrum), and we're looking at content/meaning, so we're in the \qcetgrey{[QIG-c]} branch. \citeauthor{kamath-etal-2024-generating-hotel} went for something slightly different and chose \qcetgrey{[QOG-c-2] Informativeness} which covers a bit of \textit{i} and \textit{iii} both. Alternatively, we could go for \qcetgrey{[QIG-c-3] Relevance to Input} to cover \textit{i}. %

For \textit{ii}, we could decide either that we do or that we do not want to take what's in the input into account. Suppose we go for the latter, then it's about Correctness (mentioning something that isn't a highlight, because negative, would be an error), and about the output as a whole (a selling point described in negative form would not be desirable), so it's the \qcetgrey{[QOC-w]} branch we want. Here we find \qcetgrey{[QOC-w-1] Correctness of Outputs (outputs as a whole)}. As the notes for this QCET node explain (see Appendix~\ref{qcet:QOC-w-1}), one of the uses of this QC is to test that the output is of the correct overall type for the task, so could be used here to check that the generated highlight is in fact a highlight. \citeauthor{kamath-etal-2024-generating-hotel} don't evaluate \textit{ii} separately.

\vspace{0.15cm}\noindent\textit{Fewer mistakes or misrepresentations:} Here too we quite clearly need both inputs (the long-form accommodation descriptions) and outputs (the highlights), we're in Correctness territory as we're identifying countable errors, and we need to take into account the content/meaning of outputs only as the errors are errors of meaning, not of form. This means we're in the \qcetgrey{[QIC-c]} branch, where we find \qcetgrey{[QIC-c-2] Absence of Additions (relative to input)} and \qcetgrey{[QIC-c-3] Consistency with Input} as candidate QCs,\footnote{\qcetgrey{[QIC-c-2] Absence of Omissions (relative to input)} is not suitable, because the output is not intended to express all the input, just highlights.} and these are indeed the two QCs chosen by \citeauthor{kamath-etal-2024-generating-hotel}, albeit under different names.  %

\vspace{0.15cm}\noindent\textit{Closer to target length range:} Here we only need to see the outputs, it's a feature (sometimes longer outputs are better, sometimes shorter), and we're assessing the form of outputs irrespective of the meaning, so we're in the \qcetgrey{[QOF-f]} branch, where we find \qcetgrey{[QOF-f-5] Output Length}. Results for this aspect weren't reported by \citeauthor{kamath-etal-2024-generating-hotel}, possibly because it was trivial to achieve.

\paragraph{\textit{Next steps:}} Once a set of QCs has been selected, they need to be completed with evaluation modes (extrinsic vs.\ extrinsic, subjective vs.\ objective, absolute vs.\ relative; see start of Section~\ref{sec:disentangling}, and for more details, \citeauthor{belz-etal-2020-disentangling}, \citeyear{belz-etal-2020-disentangling}) to yield a fully specified evaluation measure. The evaluation measure then needs to be completed with experimental design and implementation to yield an evaluation method that can be applied in practice (see start of Section~\ref{sec:disentangling}).

\subsection{AI regulation compliance assessment}

AI regulation is under active development in different countries, and implemented (at a coarse-grained level of specificity) in some. For the latter, interpretation and application in practice is underway. The EU's AI Act\footnote{\url{https://www.europarl.europa.eu/topics/en/article/20230601STO93804/eu-ai-act-first-regulation-on-artificial-intelligence}} for example came into force on 1 August 2024, with provisions coming into operation gradually over the following 6 to 36 months.  It requires high-impact general-purpose AI models that might pose systemic risk, such as  GPT-4 to undergo thorough evaluations under the Act's transparency requirements. What such evaluations will consist of, and what type of evaluations must be carried out, is currently not clear, but standardisation efforts are underway, led by ISO and European equivalents.

The European Commission has requested standards to be developed by ISO/IEC that can be used to assess and enforce compliance with the provisions of the AI Act, and these include assessment for different aspects of overall system `accuracy'.\footnote{ISO/IEC AWI 23282 Artificial Intelligence: Evaluation methods for accurate natural language processing systems: \url{https://www.iso.org/standard/87387.html}} 

There are two perspectives on compliance testing in this context: (i) how can developers ensure that their evaluations comply with regulatory requirements, and (ii) how can the oversight body test whether evaluations carried out by the developer comply with regulatory requirements.  In both cases, some way is clearly needed to determine objectively whether the evaluation required in the regulation, and the evaluation carried out by the developer, do in fact assess the same thing. 

For \textit{ii}, the standard used by the oversight body could directly incorporate the standardised QCET QCs in which case assessment boils down to applying the steps for identifying QCs from Section~\ref{sec:ident-qcs} to the evaluations carried out by the developer, then comparing the resulting QCET QCs with those identified in the standard. Note this is also the approach that would be used by a \textit{deployer} of technology developed by a third party to ensure they are compliant with regulations in using the technology.

For \textit{i}, the standard used by the oversight body would be used by the developer to identify the QCs that need to be covered (possibly among others) by their evaluations, otherwise following the steps for selecting QCs from Section~\ref{sec:select-qcs}.

\section{Conclusion}\label{sec:concl}

We have presented QCET, a taxonomy of standard quality criteria derived from existing NLP evaluations, designed to support (i)~assessments of the comparability of existing evaluations, (ii) guiding the design of new evaluations that are comparable by design, and (iii) assessment of regulation compliance. Rather than attempting to achieve complete coverage of QCs currently in use, 
QCET is designed to be arbitrarily extensible by adding new QCs to the appropriate branches of the taxonomy.

\appendix

\section{Diagrammatic View of QCET Taxonomy}
\label{app:diag}

\begin{figure*}[ht!]
     \centering
    \includegraphics[width=1.1\textwidth]{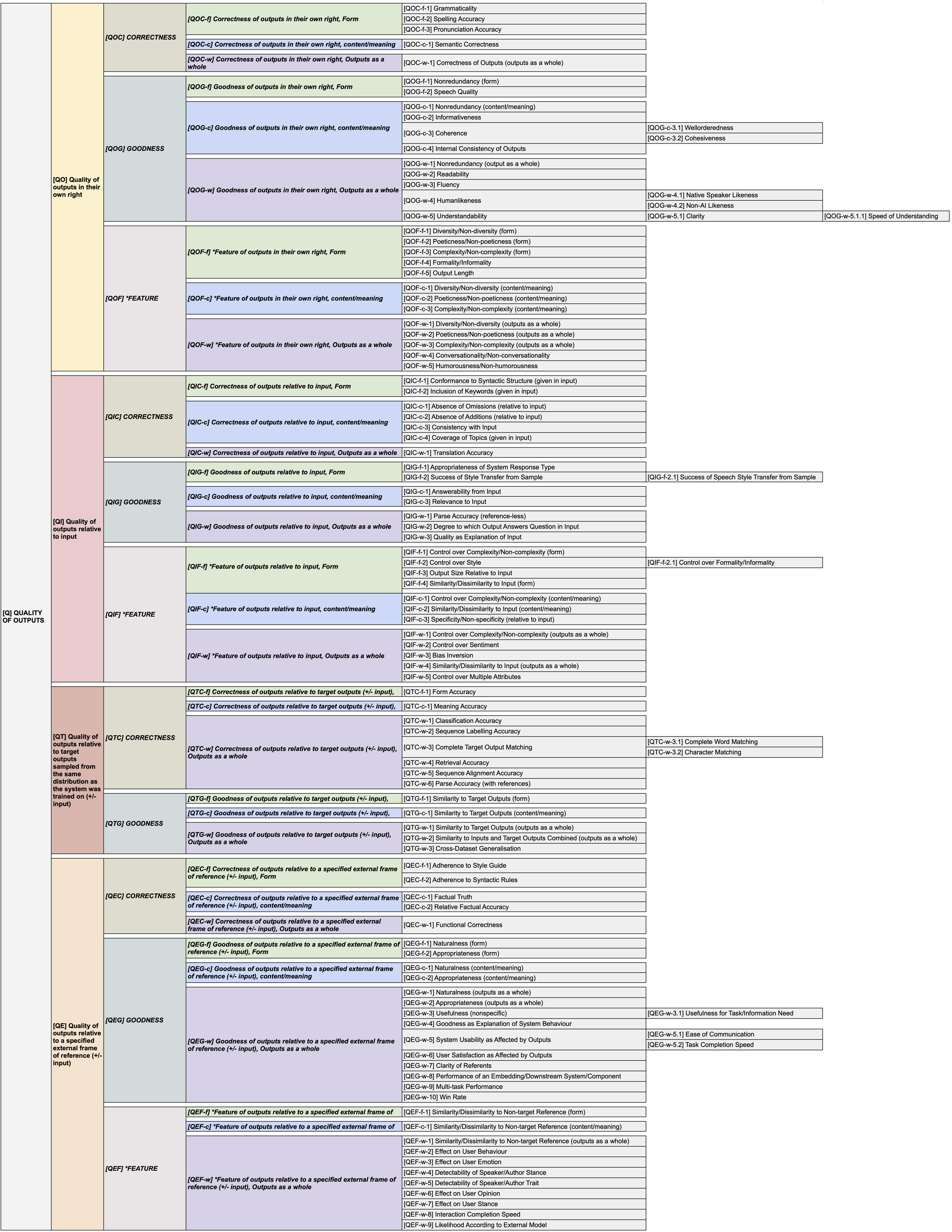}
    \caption{Diagrammatic overview of the QCET taxonomy, showing node IDs and quality criterion (QCs) names only. * = node is a class of QCs, but not a QC in its own right.}
    \label{fig:diag}
\end{figure*}

Figure~\ref{fig:diag} shows the whole of the QCET taxonomy in diagrammatic overview, displaying node IDs and QC names only.

\section{List of Quality Criteria with Definitions and Notes}\label{sec:list-qcs}
\label{app:list}

\subsection{QCs that define quality in terms of outputs only}\label{qcet:QO}

\begin{figure*}
        \centering
            \includegraphics[width=1.02\textwidth,trim={0cm 0cm 0.04cm 0cm},clip]{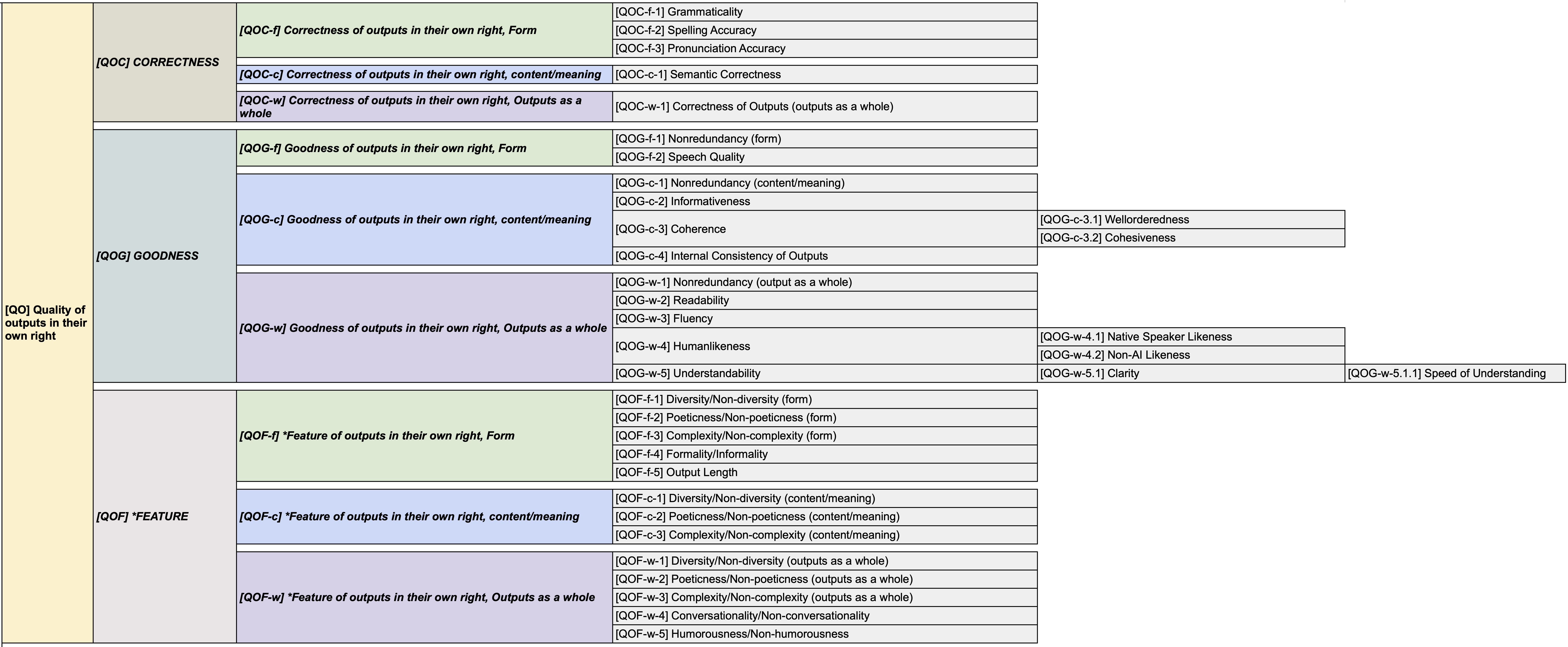}
            \caption{The \textit{Quality of outputs in their own right} branch of the QCET Taxonomy (QC IDs and names only).}
            \label{fig:output-only-branch-v3.4}
        \end{figure*}

        Figure~\ref{fig:output-only-branch-v3.4} shows a simplified view of the \textbf{Quality of outputs in their own right} branch of the QCET Taxonomy (only node IDs and Names are shown for each node). The top three levels were explained in the paper; below we list each QC and definition, with some additional explanatory notes in some cases, grouped according to subtrees (see Figure~\ref{fig:high-level-taxonomy}).

\subsubsection{Correctness}\label{qcet:QOC}

\subsubsection*{\textul{Form}}\label{qcet:QOC-f}

\label{qcet:QOC-f-1}

\vspace{0.1cm}\noindent\qcetgrey{[QOC-f-1] Grammaticality}: A better system produces texts with fewer grammatical errors.

\vspace{0.05cm}\noindent\textit{Example:} \citet{humphreys-etal-2001-reusing} informally evaluate 200 outputs manually for Grammaticality reporting 4\% of outputs with grammatical errors for their combined parser/generator.

\vspace{0.05cm}\noindent\textit{Notes:} [QOC-f-1] is Grammaticality as judged by native speakers, i.e.\ it's a human-assessable only QC. Cf.\ [QIC-f-1] Matching Syntactic Structure (given in input), and [QEC-f-2] Adherence to Syntactic Rules which can be assessed either with metrics or humans.

\label{qcet:QOC-f-2}

\vspace{0.1cm}\noindent\qcetgrey{[QOC-f-2] Spelling Accuracy}: A better system produces texts with fewer spelling errors.

\vspace{0.05cm}\noindent\textit{Example:} \citet{farrus2010linguistic} manually compare Google Translate and the language-pair specific N-II system, e.g.\ reporting 169 orthographic errors in 711 Spanish-Catalan translations for Google compared to 62 for N-II.

\vspace{0.05cm}\noindent\textit{Notes:} --

\label{qcet:QOC-f-3}

\vspace{0.1cm}\noindent\qcetgrey{[QOC-f-3] Pronunciation Accuracy}: A better system produces speech with fewer pronunciation errors.

\vspace{0.05cm}\noindent\textit{Example:} \citet{fong2024controlling} manually compare US and Scottish speech code inputs for their text-to-speech system, finding e.g.\ that the former lead to mispronunciations in 15\% of words, and the latter in 24\% (Table~7.1).

\vspace{0.05cm}\noindent\textit{Notes:} --

\subsubsection*{\textul{Content}}\label{qcet:QOC-c}

\label{qcet:QOC-c-1}

\vspace{0.1cm}\noindent\qcetgrey{[QOC-c-1] Semantic Correctness}: A better system produces outputs with fewer semantic errors.

\vspace{0.05cm}\noindent\textit{Example:} \citet{lindberg-etal-2013-generating} ask an education specialist to assess questions generated by their system in terms of semantic validity, finding e.g.\ that 66\% of them ``made sense.''

\vspace{0.05cm}\noindent\textit{Notes:} Semantic correctness is about the output being logically sound, but also obeying common-sense knowledge about the real world and events occurring in the right temporal order.

\subsubsection*{\textul{Outputs as a whole}}\label{qcet:QOC-w}

\label{qcet:QOC-w-1}

\vspace{0.1cm}\noindent\qcetgrey{[QOC-w-1] Correctness of Outputs (outputs as a whole)}: A better system produces outputs with fewer overall errors.

\vspace{0.05cm}\noindent\textit{Example:} \citet{rafatbakhsh2021development} manually determine the proportion of acceptable multiple-choice language learning items generated by their system; the assessment involved checking that question and answers matched, and that there were the right number of answers.

\vspace{0.05cm}\noindent\textit{Notes:} Correctness of Outputs is often about whether the output is of the correct type, e.g. in a question generator, \textit{is the output a question}, or if an LLM is prompted for class labels, \textit{is the output a class label}. The notion of correctness often derives simply from the system task (as in both these examples).

\subsubsection{Goodness}\label{qcet:QOG}

\subsubsection*{\textul{Form}}\label{qcet:QOG-f}

\label{qcet:QOG-f-1}

\vspace{0.1cm}\noindent\qcetgrey{[QOG-f-1] Nonredundancy (form)}: A better system produces outputs with less redundancy in their form.

\vspace{0.05cm}\noindent\textit{Example:} \citet{wang2024mitigating} assess outputs from LLM-based MT with a `repetition ratio' metric defined as the percentage of translations that have repetitions of substrings at the end.

\vspace{0.05cm}\noindent\textit{Notes:} Examples of redundancies of form include unnecessary repetitions of word or character strings, and extraneous brackets in code or mathematical expressions.

\label{qcet:QOG-f-2}

\vspace{0.1cm}\noindent\qcetgrey{[QOG-f-2] Speech Quality}: A better system produces speech that is of better quality.

\vspace{0.05cm}\noindent\textit{Example:} \citet{hu2007subjective} ask evaluators to rate speech enhancement outputs in terms of the level of distortion of the speech signal on a 5-point scale ranging from "very natural, no degradation" to "very unnatural, very degraded."

\vspace{0.05cm}\noindent\textit{Notes:} --

\subsubsection*{\textul{Content}}\label{qcet:QOG-c}

\label{qcet:QOG-c-1}

\vspace{0.1cm}\noindent\qcetgrey{[QOG-c-1] Nonredundancy (content/meaning)}: A better system produces outputs with less redundancy in their content/meaning.

\vspace{0.05cm}\noindent\textit{Example:} \citet{di-fabbrizio-etal-2014-hybrid} evaluate the `compactness' of review summaries on a 5-point scale via human evaluation on Amazon Mechanical Turk, where a compact summary is one that ``does not repeat information.''

\vspace{0.05cm}\noindent\textit{Notes:} Examples of redundancies of content/meaning include the same meaning being expressed more than once in different ways, use of full names when pronouns would suffice, and overly explanatory details (e.g.\ \textit{They closed the door behind them using the doorhandle which was affixed to the door.}).

\label{qcet:QOG-c-2}

\vspace{0.1cm}\noindent\qcetgrey{[QOG-c-2] Informativeness}: A better system produces outputs that are more informative.

\vspace{0.05cm}\noindent\textit{Example:} \citet{green-2006-generation} evaluates the outputs of a discourse generator that produces lay-oriented genetic counseling texts, by asking students to edit the outputs to ensure their contents provide the right level of information. The amount of editing done is viewed as indicative of informativeness.

\vspace{0.05cm}\noindent\textit{Notes:} Informativeness can be about information density and/or information sufficiency. E.g.\ a text that conveys a lot of information concisely, without using more words than necessary, is information dense. A text that provides the right level of information for a given scenario provides sufficient information.

\label{qcet:QOG-c-3}

\vspace{0.1cm}\noindent\qcetgrey{[QOG-c-3] Coherence}: A better system produces outputs whose contents/meaning hang(s) together better.

\vspace{0.05cm}\noindent\textit{Example:} In the 2005 DUC shared task on summarisation \cite{dang2005overview}, outputs are assessed in terms of the degree to which they meet the following statement: ``The summary should be well-structured and well-organized. The summary should not just be a heap of related information, but should build from sentence to sentence to a coherent body of information about a topic.''

\vspace{0.05cm}\noindent\textit{Notes:} --

\label{qcet:QOG-c-3.1}

\vspace{0.1cm}\noindent\qcetgrey{[QOG-c-3.1] Wellorderedness}: A better system produces outputs whose content/meaning is ordered better.

\vspace{0.05cm}\noindent\textit{Example:} In their content-planning enhanced approach to dialogue, \citet{xu2021enhancing} evaluate content ordering by first manually segmenting a dialogue by topic, and then rating each segment 1 if the ordering is appropriate, otherwise 0, and reporting the average.

\vspace{0.05cm}\noindent\textit{Notes:} [QOG-c-3.1] Wellorderedness captures whether content is ordered in a way that makes sense, e.g.\ that events are presented in the right order, that related points are made in the right place(s), etc. It is a necessary, but not sufficient, condition for [QOG-c-3] Coherence.

\label{qcet:QOG-c-3.2}

\vspace{0.1cm}\noindent\qcetgrey{[QOG-c-3.2] Cohesiveness}: A better system produces outputs whose content/meaning elements are linked better.

\vspace{0.05cm}\noindent\textit{Example:} \citet{eisentadt-elhadad-2020-neural} compare WebNLG outputs generated by (i) T5, and (ii) T5 plus a neural micro-planner, with the human target output texts in terms of the counts of cohesive devices they contain, finding that texts generated with a micro-planner are much more similar in this respect to human-written texts.

\vspace{0.05cm}\noindent\textit{Notes:} [QOG-c-3.2] Cohesiveness is all about linkage between content elements at every level of granularity, typically a property of text, it involves discourse connectives, anaphoric referencing, lexical congruity, etc. It is a necessary, but not sufficient, condition for [QOG-c-3] Coherence.

\label{qcet:QOG-c-4}

\vspace{0.1cm}\noindent\qcetgrey{[QOG-c-4] Internal Consistency of Outputs}: A better system produces outputs that are more consistent in their content/meaning.

\vspace{0.05cm}\noindent\textit{Example:} In their work on automatic alignment of images and text snippets from different sources, \citet{chen-etal-2023-weakly} evaluate output image-text pairs with the CLIPScore text-to-image similarity metric that computes the cosine similarity between the embeddings of the image and the text produced by a multimodal model.

\subsubsection*{\textul{Outputs as a whole}}\label{qcet:QOG-w}

\label{qcet:QOG-w-1}

\vspace{0.1cm}\noindent\qcetgrey{[QOG-w-1] Nonredundancy (output as a whole)}: A better system produces outputs with less overall redundancy.

\vspace{0.05cm}\noindent\textit{Example:} In the 2005 DUC shared task on summarisation \cite{dang2005overview}, outputs are assessed in terms of the degree to which they meet the following statement: ``There should be no unnecessary repetition in the summary. Unnecessary repetition might take the form of whole sentences that are repeated, or repeated facts, or the repeated use of a noun or noun phrase (e.g., `Bill Clinton') when a pronoun (`he') would suffice."

\vspace{0.05cm}\noindent\textit{Notes:} Nonredundancy of outputs as a whole captures redundancies of form and content both. The explanation from DUC 2005 quoted in the attestation for this node provides a good explanation of this QC.

\label{qcet:QOG-w-2}

\vspace{0.1cm}\noindent\qcetgrey{[QOG-w-2] Readability}: A better system produces outputs that are more readable.

\vspace{0.05cm}\noindent\textit{Example:} \citet{afsal-kuppusamy-2024-assessing-readability} compare the readability of texts generated with the Gemini LLM via different prompts with the Flesch Reading Ease (FRE) and Flesch-Kincaid Grade Level (FKGL) metrics.

\vspace{0.05cm}\noindent\textit{Notes:} Readability captures `reading ease' in the sense of text measures like Flesch and Flesch-Kincaid which aim to capture ability to easily read at different reading ages. Better readability is associated e.g. with more common words, shorter words, shorter sentences, and simpler sentence structure. Cf.\ [QOG-w-3] Fluency: very short sentences with repetitive structure would score highly on Readability, but not on Fluency.

\label{qcet:QOG-w-3}

\vspace{0.1cm}\noindent\qcetgrey{[QOG-w-3] Fluency}: A better system produces outputs that are more fluent.

\vspace{0.05cm}\noindent\textit{Example:} \citet{resendiz2025mopo} evaluate the fluency of affective text generation systems via LLM-prompting with the following prompt: ``Assess the text’s fluency, assigning a score from 1 to 5, with 5 representing the highest level of fluency. Do not give an explanation of the selection.''

\vspace{0.05cm}\noindent\textit{Notes:} Fluency captures how well text or speech flows, being absorbed readily without bringing the reader or listener up short, and without in the case of speech, hesitations, filler, or overly long pauses. For high fluency, language does not necessarily need to be simple, cf.\ [QOG-w-2] Readability.

\label{qcet:QOG-w-4}

\vspace{0.1cm}\noindent\qcetgrey{[QOG-w-4] Humanlikeness}: A better system produces outputs that are more human-like.

\vspace{0.05cm}\noindent\textit{Example:} \citet{cercas-curry-etal-2015-generating} assess the humanlikeness of navigation instructions by asking evaluators to rate the extent to which an instruction ``could have been produced by a human'' on a 4-point scale. 

\vspace{0.05cm}\noindent\textit{Notes:} --

\label{qcet:QOG-w-4.1}

\vspace{0.1cm}\noindent\qcetgrey{[QOG-w-4.1] Native Speaker Likeness}: A better system produces speech or text that is more like that of a native speaker.

\vspace{0.05cm}\noindent\textit{Example:} [QGO-f-2-1] \citet{novikova-etal-2018-rankme} assess system outputs e.g.\ by asking evaluators the question: ``Could the utterance have been produced by a native speaker?''

\vspace{0.05cm}\noindent\textit{Notes:} --

\label{qcet:QOG-w-4.2}

\vspace{0.1cm}\noindent\qcetgrey{[QOG-w-4.2] Non-AI Likeness}: A better system produces outputs that are less like those produced by AI.

\vspace{0.05cm}\noindent\textit{Example:} \citet{juraska-etal-2019-viggo} ask human judges to assess model outputs in terms of ``how much one would expect to encounter [this] utterance in a conversation with a human, as opposed to sounding robotic'.'

\vspace{0.05cm}\noindent\textit{Notes:} --

\label{qcet:QOG-w-5}

\vspace{0.1cm}\noindent\qcetgrey{[QOG-w-5] Understandability}: A better system produces outputs that are more understandable.

\vspace{0.05cm}\noindent\textit{Example:} \citet{hershenhouse2024accuracy} assess LLMs in terms of their ability to communicate medical information to the public by asking crowdworkers to demonstrate their understanding of generated texts through multiple-choice questions.

\vspace{0.05cm}\noindent\textit{Notes:} Understandability captures whether an output can be understood and is commonly evaluated in terms of whether it has been understood (via comphrehension questions). Cf.\ sub-QC [QOG-w-5.1] Clarity for which an output is assessed in terms of the higher-threshold criterion whether it can be \textit{easily} understood.

\label{qcet:QOG-w-5.1}

\vspace{0.1cm}\noindent\qcetgrey{[QOG-w-5.1] Clarity}: A better system produces outputs that are clearer.

\vspace{0.05cm}\noindent\textit{Example:} \citet{clinciu2021study} evaluate a system that generates explanations of Bayesian network graphs, e.g.\ in terms of clarity where evaluators are asked to indicate ``[h]ow clear the meaning of an explanation is” on a 7-point scale, where 1 = unclear and 7 = very clear.

\vspace{0.05cm}\noindent\textit{Notes:} --

\label{qcet:QOG-w-5.1.1}

\vspace{0.1cm}\noindent\qcetgrey{[QOG-w-5.1.1] Speed of Understanding}: A better system produces outputs that are faster to understand.

\vspace{0.05cm}\noindent\textit{Example:} \citet{mohanmodifying} assessed system explanations in terms of the time it took people to select a response when asked how well they understood the explanations generated by the system.

\vspace{0.05cm}\noindent\textit{Notes:} --

\subsubsection{Feature}\label{qcet:QOF}

\subsubsection*{\textul{Form}}\label{qcet:QOF-f}

\label{qcet:QOF-f-1}

\vspace{0.1cm}\noindent\qcetgrey{[QOF-f-1] Diversity/Non-diversity (form)}: A better system produces outputs that are in their form either (a) more diverse, or (b) less diverse.

\vspace{0.05cm}\noindent\textit{Example:} \citet{chen-etal-2024-evaluating-diversity} compare the average type-token ratio (ATTR) of poems generated with a range of models, finding e.g.\ that humans, poetry-specific models and general models have broadly similar ATTR scores. 

\vspace{0.05cm}\noindent\textit{Notes:} Diversity of form captures the variedness of the form of outputs either at the level of individual outputs where those are longer than a sentence, or at the level of a sample of outputs. Varied form could be diverse sentence structures, format, or speech patterns, etc.

\label{qcet:QOF-f-2}

\vspace{0.1cm}\noindent\qcetgrey{[QOF-f-2] Poeticness/Non-poeticness (form)}: A better system produces outputs that are in their form either (a) more poetic, or (b) less poetic.

\vspace{0.05cm}\noindent\textit{Example:} \citet{xie2017deep} manually compare four LSTM variants in terms of the quality of rhyme and metre, finding that gated and CNN-based LSTMs perform better than word and character LSTMs.

\vspace{0.05cm}\noindent\textit{Notes:} (Non)poeticness of form captures the degree to which outputs have the formal characteristics of a poem, including alliteration, rhythm, rhyming and specific syllable/metre patterns. For [QOF-f-2], this is assessed without taking the input or anything external to the system into account.

\label{qcet:QOF-f-3}

\vspace{0.1cm}\noindent\qcetgrey{[QOF-f-3] Complexity/Non-complexity (form)}: A better system produces outputs that are in their form either (a) more complex, or (b) less complex.

\vspace{0.05cm}\noindent\textit{Example:} In their evaluation of a text simplification system for Spanish, \citet{saggion2015making} measure the syntactic complexity of a sentence as the maximum distance between root and leaf nodes in the dependency tree of the sentence.

\vspace{0.05cm}\noindent\textit{Notes:} Complexity of form captures aspects of output complexity irrespective of meaning, e.g.\ longer word and sentence length, nested syntactic structure, and low frequency words can all be indicative of higher complexity of form.

\label{qcet:QOF-f-4}

\vspace{0.1cm}\noindent\qcetgrey{[QOF-f-4] Formality/Informality}: A better system produces outputs that are either (a) more formal, or (b) less formal.

\vspace{0.05cm}\noindent\textit{Example:} \citet{abu-sheikha-inkpen-2011-generation} asked human annotators to label system outputs from their template-based, formality-controlled text generator as either formal or informal.

\vspace{0.05cm}\noindent\textit{Notes:} (In)formality captures how relaxed the language of outputs is: in a conversation with friends, or on social media, language tends to be more informal, whereas in academic articles or legal contexts, it tends to be much more formal.

\label{qcet:QOF-f-5}

\vspace{0.1cm}\noindent\qcetgrey{[QOF-f-5] Output Length}: A better system produces outputs that are either (a) longer, or (b) shorter.

\vspace{0.05cm}\noindent\textit{Example:} \citet{ghalandari-etal-2022-efficient} report the mean length of the outputs from their unsupervised sentence compression system.

\subsubsection*{\textul{Content}}\label{qcet:QOF-c}

\label{qcet:QOF-c-1}

\vspace{0.1cm}\noindent\qcetgrey{[QOF-c-1] Diversity/Non-diversity (content/meaning)}: A better system produces outputs that are in their content/meaning either (a) more diverse, or (b) less diverse.

\vspace{0.05cm}\noindent\textit{Example:} \citet{chen-etal-2024-evaluating-diversity} use mean cosine similarity between SBERT embeddings to assess the semantic diversity of generated poems, finding e.g.\ that their German poems are on the whole substantially more diverse than the English ones.

\vspace{0.05cm}\noindent\textit{Notes:} (Non)diversity of content/meaning captures the variedness of the content/meaning of outputs either at the level of individual outputs where those are longer than a sentence, or at the level of a sample of outputs. Varied content/meaning could be diverse topics, information, or events, etc.

\label{qcet:QOF-c-2}

\vspace{0.1cm}\noindent\qcetgrey{[QOF-c-2] Poeticness/Non-poeticness (content/meaning)}: A better system produces outputs that are in their content/meaning either (a) more poetic, or (b) less poetic.

\vspace{0.05cm}\noindent\textit{Example:} \citet{wu2019evaluating} ask evaluators to assess ``whether some part of the poem is imaginative and/or moving'' in a 4-step assessment procedure which is then mapped to a single score.

\vspace{0.05cm}\noindent\textit{Notes:} (Non)poeticness of form captures the degree to which outputs have the semantic characteristics of a poem, including metaphor, topic and narrative structure. For [QOF-c-2], this is assessed without taking the input or anything external to the system into account.

\label{qcet:QOF-c-3}

\vspace{0.1cm}\noindent\qcetgrey{[QOF-c-3] Complexity/Non-complexity (content/meaning)}: A better system produces outputs that are in their content/meaning either (a) more complex, or (b) less complex.

\vspace{0.05cm}\noindent\textit{Example:} \citet{onoe2024doccidescriptionsconnectedcontrasting} measure semantic complexity as the number of nodes in the `scene graph' corresponding to an image description.

As part of the composite LLM Rater metric, \citet{luo2024laypersonsguidebiomedicine} ask evaluators to assess the extent to which a summary ``avoid[s] the use of technical details that would be difficult for non-expert readers to understand[, ... and] contains sufficient explanations of any complex terms and abbreviations.''

\vspace{0.05cm}\noindent\textit{Notes:} Complexity of content/meaning captures aspects of output complexity irrespective of form, e.g.\ complex logical structure or technical details can be indicative of higher semantic complexity.

\subsubsection*{\textul{Outputs as a whole}}\label{qcet:QOF-w}

\label{qcet:QOF-w-1}

\vspace{0.1cm}\noindent\qcetgrey{[QOF-w-1] Diversity/Non-diversity (outputs as a whole)}: A better system produces outputs that are either (a) more diverse, or (b) less diverse.

\vspace{0.05cm}\noindent\textit{Example:} \citet{jin-le-2016-selecting} ask evaluators to assess the overall diversity of a set of questions generated by a system from a given input text.

\citet{li2015diversity} assess the diversity of the responses produced by their conversational system with the distinct-1 and distinct-2 metrics computed as the number of distinct unigrams (bigrams) over the total number of generated tokens.

\vspace{0.05cm}\noindent\textit{Notes:} (Non)diversity of outputs as a whole captures diversity of form and content both, either at the level of individual outputs where those are longer than a sentence, or at the level of a sample of outputs. E.g.\ a diverse set of questions and answers generated for a given text passage would have little overlap in coverage of the text between them.

\label{qcet:QOF-w-2}

\vspace{0.1cm}\noindent\qcetgrey{[QOF-w-2] Poeticness/Non-poeticness (outputs as a whole)}: A better system produces outputs that are either (a) more poetic, or (b) less poetic.

\vspace{0.05cm}\noindent\textit{Example:} \citet{xie2017deep} ask evaluators to assess the overall poeticness of generated poems, finding e.g.\ that poems generated by more complex LSTMs are perceived as more poetic than those generated by simple LSTMs.

\vspace{0.05cm}\noindent\textit{Notes:} (Non)poeticness of outputs captures the degree to which outputs have the characteristics of a poem, including alliteration, rhythm, rhyming and specific syllable/metre patterns, metaphor, topic and narrative structure. For [QOF-w-2], this is assessed without taking the input or anything external to the system into account.

\label{qcet:QOF-w-3}

\vspace{0.1cm}\noindent\qcetgrey{[QOF-w-3] Complexity/Non-complexity (outputs as a whole)}: A better system produces outputs that are either (a) more complex, or (b) less complex.

\vspace{0.05cm}\noindent\textit{Example:} \citet{angrosh-siddharthan-2014-text} compare their rule-based simplification system (which combines handcrafted with automatically acquired rules) to a learned system and a human topline in terms of human-assessed text simplicity, finding that their system outperforms the learned system.

\vspace{0.05cm}\noindent\textit{Notes:} Complexity of outputs captures aspects of output complexity both of form and content/meaning, e.g.\ longer word and sentence length, nested syntactic structure, and low frequency words, complex logical structure or technical details can all be indicative of higher complexity.

\citet{zhang2024atlasimprovinglaysummarisation} assess the overall layness of generated lay summaries via human evaluation (``to what extent is the content of the model output comprehensible (or readable) to a non-expert, in terms of both structure and language?"), and three automatic metrics.

\label{qcet:QOF-w-4}

\vspace{0.1cm}\noindent\qcetgrey{[QOF-w-4] Conversationality/Non-conversationality}: A better system produces outputs that are either (a) more conversational, or (b) less conversational.

\vspace{0.05cm}\noindent\textit{Example:} \citet{cervone-etal-2019-natural} ask evaluators to assess how conversational the turns generated by different versions of a conversational agent are on a scale of 1--6.

\vspace{0.05cm}\noindent\textit{Notes:} Conversationality is typically assessed in dialogue scenarios where it captures the degree to which a series of turns between user and system resemble conversations between people, usually in the same situation. While it's difficult to conceive of situations where Nonconversationality would need to be assessed explicitly, it is certainly the case that in many situations user-facing generated text should not be conversational, in particular in single-turn scenarios such as question-answering.

\label{qcet:QOF-w-5}

\vspace{0.1cm}\noindent\qcetgrey{[QOF-w-5] Humorousness/Non-humorousness}: A better system produces outputs that are either (a) more humorous, or (b) less humorous.

\vspace{0.05cm}\noindent\textit{Example:} \citet{alnajjar-hamalainen-2018-master} assess whether generated film titles are humorous, on a 1--5 agreement scale.

\vspace{0.05cm}\noindent\textit{Notes:} --

\subsection{QCs that define quality relative to inputs}\label{qcet:QI}

        Figure~\ref{fig:high-level-taxonomy} shows a simplified view of the \textbf{Quality of outputs relative to input} branch of the QCET Taxonomy (only node IDs and Names are shown for each node). Below we list each QC and definition, with some additional explanatory notes in some cases, grouped according to subtrees (see Figure~\ref{fig:high-level-taxonomy}).

        \begin{figure*}[ht]
        \centering
            \includegraphics[width=1.02\textwidth,trim={0cm 0cm 0.04cm 0cm},clip]{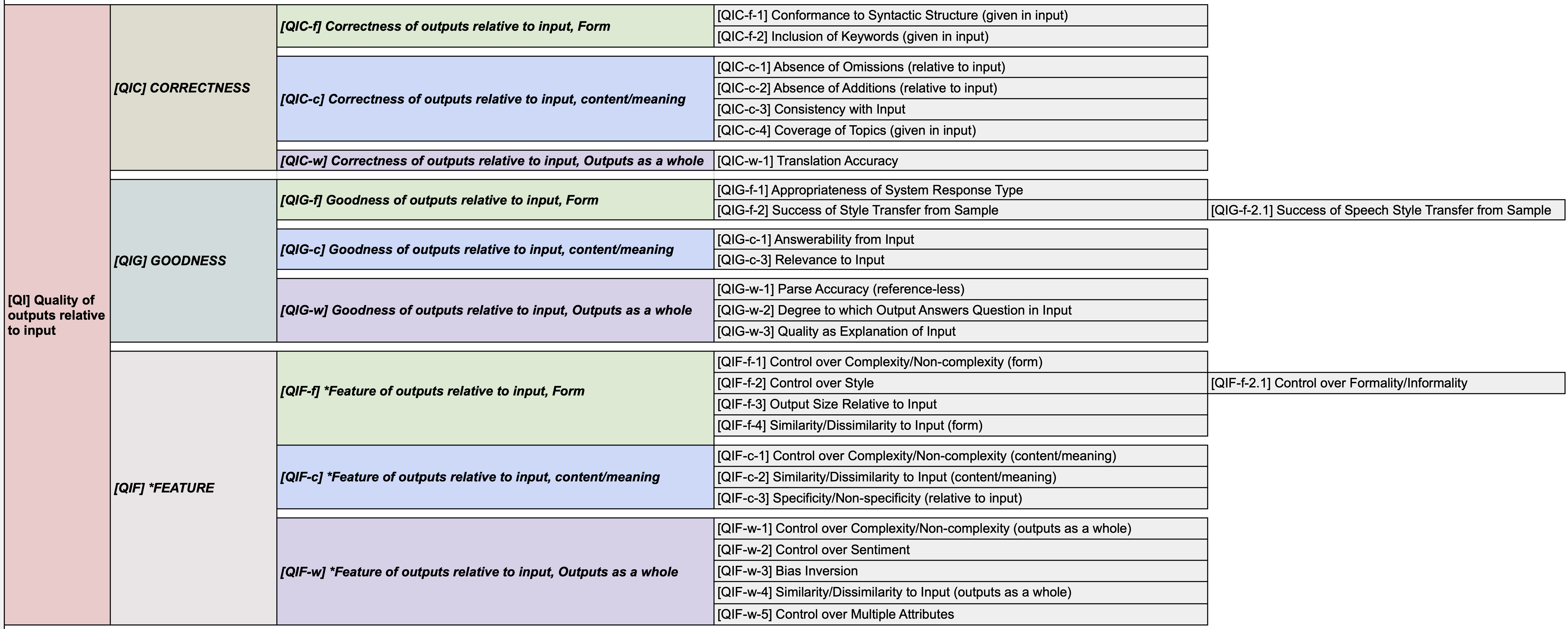}
            \caption{The \textit{Quality of outputs relative to input} branch of the QCET Taxonomy (QC IDs and names only).}
            \label{fig:rel-to-input-branch-v3.4}
        \end{figure*}

\subsubsection{Correctness}\label{qcet:QIC}

\subsubsection*{\textul{Form}}\label{qcet:QIC-f}

\label{qcet:QIC-f-1}

\vspace{0.1cm}\noindent\qcetgrey{[QIC-f-1] Conformance to Syntactic Structure (given in input)}: A better system produces texts with fewer deviations from the target syntactic structure provided in the input.

\vspace{0.05cm}\noindent\textit{Example:} \citet{kumar2020syntax} compute the tree-edit distance between the syntax exemplar given in the input and the output generated by their syntax-controlled paraphrasing system.

\vspace{0.05cm}\noindent\textit{Notes:} --

\label{qcet:QIC-f-2}

\vspace{0.1cm}\noindent\qcetgrey{[QIC-f-2] Inclusion of Keywords (given in input)}: A better system produces texts that lack fewer of the keywords provided in the input.

\vspace{0.05cm}\noindent\textit{Example:} \citet{sasazawa2023controllingkeywordspositionstext} compute the proportion of outputs that contain all the keywords given in the input in several keyword position controlled text generation systems.

\vspace{0.05cm}\noindent\textit{Notes:} --

\subsubsection*{\textul{Content}}\label{qcet:QIC-c}

\label{qcet:QIC-c-1}

\vspace{0.1cm}\noindent\qcetgrey{[QIC-c-1] Absence of Omissions (relative to input)}: A better system produces outputs that lack fewer of the content/meaning units provided in the input.

\vspace{0.05cm}\noindent\textit{Example:} \citet{gonzalez-corbelle-etal-2022-dealing} assess their transformer-based weather forecast generator e.g.\ with a metric that computes the number of output texts with omissions (input elements not literally contained in the output), finding omissions in 160 out of 272 texts (58\%).

\vspace{0.05cm}\noindent\textit{Notes:} Absence of Omissions is defined for cases where Absence of Additions, as its inverse, is also defined. This will typically hold in those cases where the output is supposed to contain/cover/verbalise all and only content/items/formal representations provided in the input. Cf. [QIC-f-2] Inclusion of Keywords where addition is not defined.

\label{qcet:QIC-c-2}

\vspace{0.1cm}\noindent\qcetgrey{[QIC-c-2] Absence of Additions (relative to input)}: A better system produces outputs that contain fewer content/meaning units not provided in the input.

\vspace{0.05cm}\noindent\textit{Example:} \citet{cripwell20232023} evaluate the semantic accuracy of the outputs of data-to-text systems in under-resourced languages using several criteria, one of them being the absence of additions. Human evaluators are shown the structured data input and an output text and are asked ``Looking at the Text, is all of its content expressed in the Data expression? (Allow duplication of content.)''

\vspace{0.05cm}\noindent\textit{Notes:} Absence of Additions is defined for cases where Absence of Omissions, as its inverse, is also defined. This will typically hold in those cases where the output is supposed to contain/cover/verbalise all and only content/items/formal representations provided in the input. Cf. the three Nonreduncy QGs, [QOG-f-1], [QOG-c-1], and [QOG-w-1], where omission is not defined.

\label{qcet:QIC-c-3}

\vspace{0.1cm}\noindent\qcetgrey{[QIC-c-3] Consistency with Input}: A better system produces outputs that have fewer inconsistencies with a given aspect of the input.

\vspace{0.05cm}\noindent\textit{Example:} \citet{shu2021logicconsistencytextgenerationsemantic} propose a metric called bi-directional logic evaluation of consistency (BLEC) for evaluating the consistency between database query logic inputs and textual questions in the output.

\vspace{0.05cm}\noindent\textit{Notes:} Cf. Similarity/Dissimilarity to Input (content/meaning); Consistency with Input is not about meaning similarity, but consistency in a task-specific sense, for which absence of contradictions may be sufficient.

\label{qcet:QIC-c-4}

\vspace{0.1cm}\noindent\qcetgrey{[QIC-c-4] Coverage of Topics (given in input)}: A better system produces outputs that lack fewer of the topics provided in the input.

\vspace{0.05cm}\noindent\textit{Example:} \citet{basu-roy-chowdhury-etal-2022-unsupervised} assess summarisation systems in terms of the average number of distinct aspects (fine-grained topics) covered in generated summaries, finding e.g.\ that the aspect-aware variant of their system increases aspect coverage by about 1 topic on average.

\vspace{0.05cm}\noindent\textit{Notes:} --

\subsubsection*{\textul{Outputs as a whole}}\label{qcet:QIC-w}

\label{qcet:QIC-w-1}

\vspace{0.1cm}\noindent\qcetgrey{[QIC-w-1] Translation Accuracy}: A better system produces translations of the input with fewer translation errors.

\vspace{0.05cm}\noindent\textit{Example:} \citet{popovic-2020-informative} asked annotators to identify and mark up word spans in machine-translated text whose meaning differed from the input text, then investigated different ways of numerically aggregating the annotations.

\vspace{0.05cm}\noindent\textit{Notes:} When assessed by human evaluators, Translation Accuracy is often broken down into different error types which are assessed (and sometimes reported) separately. However, such error taxonomies differ too widely in granularity and meaning between papers to be incorporated here as sub-QCs. Among metrics, BLEU (created for MT) works particularly well, especially when assessed against multiple target output translations per input.

\subsubsection{Goodness}\label{qcet:QIG}

\subsubsection*{\textul{Form}}\label{qcet:QIG-f}

\label{qcet:QIG-f-1}

\vspace{0.1cm}\noindent\qcetgrey{[QIG-f-1] Appropriateness of System Response Type}: A better system produces outputs that are of a more appropriate type relative to the input.

\vspace{0.05cm}\noindent\textit{Example:} \citet{webb-etal-2010-evaluating} evaluate whether or not it was appropriate (a) for the system to give a response when it responded, and (b) for the stystem not to respond when it didn't, regardless of the content of the response.

\vspace{0.05cm}\noindent\textit{Notes:} Appropriateness of System Response Type captures, at the level of entire responses, whether the system response was of the right type. Often used in dialogue, evaluation methods assessing this QC might take into account whether the system took initiative or handed over to a human when it should have done. In a question-answering scenario, whether the system response constituted an attempt to answer the question (as distinct from whether that answer was correct or informative) might be assessed.

\label{qcet:QIG-f-2}

\vspace{0.1cm}\noindent\qcetgrey{[QIG-f-2] Success of Style Transfer from Sample}: A better system produces outputs that are more like the style sample provided in the input.

\vspace{0.05cm}\noindent\textit{Example:} In their survey of text style transfer research, \citet{hu-etal-2022-text-style-transfer} identify two main ways in which previous work has measured style transfer success: transfer acccuracy measured as the accuracy achieved by style classifiers compared to the intended style; and earth mover distance between the style distributions of the input text and the transferred text.

\vspace{0.05cm}\noindent\textit{Notes:} Cf. Control over Alternative Styles; in Success of Style Transfer from Sample, content and style samples are provided in the input, whereas in `Control over' QCs, the task definition includes the controlled attribute and its finite set of possible values one of which is added to the input.

\label{qcet:QIG-f-2.1}

\vspace{0.1cm}\noindent\qcetgrey{[QIG-f-2.1] Success of Speech Style Transfer from Sample}: A better system produces speech that sounds more like the speech sample specified in the input.

\vspace{0.05cm}\noindent\textit{Example:} \citet{peng-etal-2024-voicecraft} evaluate speech LLMs on speech editing tasks where the input is a speech recording, corresponding transcript and an edited version of the transcript, and the output is a spoken rendition of the modified transcript that is intended to sound like the speech recording. The evaluation assesses if the speech sample and the spoken system output sound the same, via metric (source and target sound file comparion) and human evaluation (rate input/output similarity on a scale from 1 to 5).

\vspace{0.05cm}\noindent\textit{Notes:} As is the case for the parent node, [QIG-f-2] Success of Style Transfer from Sample, here the target style is provided in the form of a sample in the input. In contrast, in [QIF-f-2] Control over Style, typically the system is trained on a given number of different styles which are selected via control attributes in the input.

\subsubsection*{\textul{Content}}\label{qcet:QIG-c}

\label{qcet:QIG-c-1}

\vspace{0.1cm}\noindent\qcetgrey{[QIG-c-1] Answerability from Input}: A better system produces questions that are more answerable on the basis of information provided in the input.

\vspace{0.05cm}\noindent\textit{Example:} In their evaluation of a question generation system, \citet{harrison-walker-2018-neural} ask human assessors to evaluate on a 4-point scale how much of the information required to correctly answer the generated question is contained within the [input] text passage.

\vspace{0.05cm}\noindent\textit{Notes:} --

\label{qcet:QIG-c-3}

\vspace{0.1cm}\noindent\qcetgrey{[QIG-c-3] Relevance to Input}: A better system produces outputs that are in a given sense more relevant to the input.

\vspace{0.05cm}\noindent\textit{Example:} \citet{shen-etal-2022-knowledge} evaluate systems for the automatic generation of counseling dialogue turns by asking evaluators to indicate on a 3-point scale ``whether the response is on topic and relevant to the dialogue history" (the latter being provided in the input).

\vspace{0.05cm}\noindent\textit{Notes:} Relevance relative to the input is often assessed in a question-answering, instruction-generation or dialogue context, where it captures the relevance of the answer to the user's question, of the instruction to the user's need, and of the dialogue to the dialogue context. 'The given sense' in which relevance is assessed is often specified in human evaluations, e.g. whether the answer responds to the question.

\subsubsection*{\textul{Outputs as a whole}}\label{qcet:QIG-w}

\label{qcet:QIG-w-1}

\vspace{0.1cm}\noindent\qcetgrey{[QIG-w-1] Parse Accuracy (reference-less)}: A better system produces outputs that are more complete and accurate parses of the input.

\vspace{0.05cm}\noindent\textit{Example:} \citet{opitz-frank-2022-better} evaluate parsers by asking evaluators to asess the `parse acceptability' of parser outputs (graphs) given the input sentence.

\vspace{0.05cm}\noindent\textit{Notes:} Parse Accuracy is sometimes assessed by asking parsing experts whether e.g.\ output phrase-structure or depency parses represent complete and correct analyses of the input sequence.

\label{qcet:QIG-w-2}

\vspace{0.1cm}\noindent\qcetgrey{[QIG-w-2] Degree to which Output Answers Question in Input}: A better system produces outputs that are more complete and accurate answers to questions provided in the input.

\vspace{0.05cm}\noindent\textit{Example:} In the TREC-8 Question Answering shared task \citet{voorhees1999trec} ask human assessors whether answers generated by participating systems contain words that answer the input question.

\vspace{0.05cm}\noindent\textit{Notes:} --

\label{qcet:QIG-w-3}

\vspace{0.1cm}\noindent\qcetgrey{[QIG-w-3] Quality as Explanation of Input}: A better system produces outputs that are more complete and accurate explanations of the inputs.

\vspace{0.05cm}\noindent\textit{Example:} \citet{sarsa-etal-code-explanations-2022} evaluate automatically generated code explanations by answering the question ``Are all parts of the code explained?'' (Yes / No) and computing the proportion of correctly explained lines out of all the generated explanation lines.

\vspace{0.05cm}\noindent\textit{Notes:} Cf.\ [QEG-w-4] Goodness as Explanation of System Behaviour; Quality as Explanation of Input only takes the input and output into account, assessing how well the latter explains the former.

\subsubsection{Feature}\label{qcet:QIF}

\subsubsection*{\textul{Form}}\label{qcet:QIF-f}

\label{qcet:QIF-f-1}

\vspace{0.1cm}\noindent\qcetgrey{[QIF-f-1] Control over Complexity/Non-complexity (form)}: A better system produces outputs that are in their form more at the target level of complexity provided in the input.

\vspace{0.05cm}\noindent\textit{Example:} \citet{agrawal2023complexity} assesses the degree to which complexity-controlled machine translation systems succeed in generating translations at the target level of complexity by computing the correlation between the Automatic Readability Index (ARI) scores of the actual and target translations.

\vspace{0.05cm}\noindent\textit{Notes:} Complexity of form captures aspects of output complexity irrespective of meaning, e.g.\ longer word and sentence length, nested syntactic structure, and low frequency words can all be indicative of higher complexity of form. [QIF-f-1] Control over Complexity/Non-complexity (form) is for systems where target levels of (non)complexity of form are defined and the system is trained to achieve them.

\label{qcet:QIF-f-2}

\vspace{0.1cm}\noindent\qcetgrey{[QIF-f-2] Control over Style}: A better system produces outputs that are in their form more in the target style provided in the input.

\vspace{0.05cm}\noindent\textit{Example:} \citet{gero-etal-2019-low} evaluate whether automatically generated sentences exhibit the style specifed in the input, by asking human annotators to label output sentences with style labels and then calculating the accuracy compared to the input labels.

\vspace{0.05cm}\noindent\textit{Notes:} Style captures aspects of form, as opposed to meaning. It relates to the way something is said, rather than what is said. Examples include formal/informal, literary/non-fiction, and more fine-grained distinctions such as newspaper house style and personal writing style. [QIF-f-2] Control over Style typically is used to evaluate systems that are trained on a specific set of alternative styles, with a control attribute in the input indicating the style that outputs are supposed to be generated in.

\label{qcet:QIF-f-2.1}

\vspace{0.1cm}\noindent\qcetgrey{[QIF-f-2.1] Control over Formality/Informality}: A better system produces outputs that are in their form more at the target level of formality provided in the input.

\vspace{0.05cm}\noindent\textit{Example:} \citet{Yang_2021} assess the outputs of an informal-to-formal MT system in terms of the mean likelihood that the output is indeed formal according to a formality classifier trained on separate data.

\vspace{0.05cm}\noindent\textit{Notes:} (In)formality captures how relaxed the language of outputs is: in a conversation with friends, or on social media, language tends to be more informal, whereas in academic articles or legal contexts, it tends to be more formal. [QIF-f-2.1] Control over Formality/Informality typically is used to evaluate systems that are trained on data with different levels of formality, with a control attribute in the input indicating the level of formality that outputs are supposed to be generated in. Cf.\ [QOF-f-4] Formality/Informality which captures the (in)formality of the output when generating outputs with given levels of formality is not part of the system task.

\label{qcet:QIF-f-3}

\vspace{0.1cm}\noindent\qcetgrey{[QIF-f-3] Output Size Relative to Input}: A better system produces outputs that achieve, relative to the input, (a) a greater reduction in size, (b) a greater increase in size, or (c) size change at the target level given in the input.

\vspace{0.05cm}\noindent\textit{Example:} \citet{clarke-lapata-2006-models} compare human and machine-produced sentence-level summaries via their respective compression rates (number of tokens in output sentences over number of tokens in input sentences), finding that humans are more conservative than machines.

\vspace{0.05cm}\noindent\textit{Notes:} [QIG-f-3] Output Size Relative to Input is commonly assessed as the ratio of output size over input size, where size is e.g. measured as number of words or characters. Typical NLP tasks where this QC is assessed include sentence summarisation and simplification.

\label{qcet:QIF-f-4}

\vspace{0.1cm}\noindent\qcetgrey{[QIF-f-4] Similarity/Dissimilarity to Input (form)}: A better system produces outputs that in their form are (a) more similar to the input, (b) less similar to the input, or (c) more at the target level of similarity to the input, where that target level is provided in the input.

\vspace{0.05cm}\noindent\textit{Example:} \citet{yin-etal-2022-ingredients} measure the 'syntactic divergence' between text input and paraphrased output using Kendall's tau.

\subsubsection*{\textul{Content}}\label{qcet:QIF-c}

\label{qcet:QIF-c-1}

\vspace{0.1cm}\noindent\qcetgrey{[QIF-c-1] Control over Complexity/Non-complexity (content/meaning)}: A better system produces outputs that are in their content/meaning more at the target level of complexity provided in the input.

\vspace{0.05cm}\noindent\textit{Example:} \citet{imperial2024standardizealigninglanguagemodels} assess content generation which uses English language standards (CEFR, CCS) to control complexity by asking expert assessors to assign English standard levels to generated stories, then measuring the accuracy relative to target levels specified in inputs.

\vspace{0.05cm}\noindent\textit{Notes:} Complexity of content/meaning captures aspects of output complexity irrespective of form, e.g.\ complex logical structure or technical details can be indicative of higher semantic complexity. [QIF-c-1] Control over Complexity/Non-complexity (content/meaning) is for systems where target levels of (non)complexity of content/meaning are defined and the system is trained to achieve them.

\label{qcet:QIF-c-2}

\vspace{0.1cm}\noindent\qcetgrey{[QIF-c-2] Similarity/Dissimilarity to Input (content/meaning)}: A better system produces outputs that in their content/meaning are (a) more similar to the input, (b) less similar to the input, or (c) more at the target level of similarity to the input, where that target level is provided in the input.

\vspace{0.05cm}\noindent\textit{Example:} In their survey of text style transfer research, \citet{hu-etal-2022-text-style-transfer} identify some of the main ways in which previous work has measured the meaning (dis)similarity between source text and transferred text: cosine similarity between embeddings, word overlap (excluding style-related words), and human assessment of meaning (dis)similarity.

\vspace{0.05cm}\noindent\textit{Notes:} [QIF-c-2] Similarity/Dissimilarity to Input (content/meaning) is the same as [QEF-c-1] Similarity/Dissimilarity to Non-target Reference (content/meaning), but here the comparison is against the input, rather than a system-external reference. Typical NLP tasks where this QC is assessed include paraphrasing and style transfer.

\label{qcet:QIF-c-3}

\vspace{0.1cm}\noindent\qcetgrey{[QIF-c-3] Specificity/Non-specificity (relative to input)}: A better system produces outputs that, relative to some aspect of the input, are (a) more specific, (b) less specific, or (c) more at the target level of specificity provided in the input.

\vspace{0.05cm}\noindent\textit{Example:} In an evaluation of a content selection method for question generation, \citet{jin-le-2016-selecting} ask human judges to assess in a binary fashion whether or not a question is specific enough considering the input text.

\vspace{0.05cm}\noindent\textit{Notes:} [QIF-c-3] Specificity/Non-specificity (relative to input) is about the level of specificity with which the output addrsses a given aspect of the input. E.g.\ where the output answers an input question, does it do so with enough specificity, or is it too vague? Similarly if the output is instructions for a given task, are the instructions specific enough to solve the task?

\subsubsection*{\textul{Outputs as a whole}}\label{qcet:QIF-w}

\label{qcet:QIF-w-1}

\vspace{0.1cm}\noindent\qcetgrey{[QIF-w-1] Control over Complexity/Non-complexity (outputs as a whole)}: A better system produces outputs that are more at the target level of complexity provided in the input.

\vspace{0.05cm}\noindent\textit{Example:} \citet{moraes-etal-2016-enabling} assess their system's ability to generate graph summaries at the target level of complexity by first asking evaluators to rank systems in terms of complexity (=suitability for grade levels), then computing the proportion of pairwise ranks that match those of the target complexity levels, finding a match in 72\% of cases.

\vspace{0.05cm}\noindent\textit{Notes:} Complexity of outputs captures aspects of output complexity both of form and content/meaning, e.g.\ longer word and sentence length, nested syntactic structure, and low frequency words, complex logical structure or technical details can all be indicative of higher complexity.  [QIF-w-1] Control over Complexity/Non-complexity (outputs as a whole) is for systems where target levels of (non)complexity are defined and the system is trained to achieve them.

\citet{luo2024laypersonsguidebiomedicine} prompt an LLM to summarise articles ``for a lay audience,'' asking evaluators to assess summaries in terms of the following statement:``Layness: Compared to the original article, the summary should decrease the linguistic complexity, omit content that is too technical, and include sufficient background explanation of technical terms."

\label{qcet:QIF-w-2}

\vspace{0.1cm}\noindent\qcetgrey{[QIF-w-2] Control over Sentiment}: A better system produces outputs that are more at the target level of positivity/negativity provided in the input.

\vspace{0.05cm}\noindent\textit{Example:} \citet{kumar2024synthesizingsentimentcontrolledfeedbackmultimodal} assess the effectiveness of sentiment control (negative, positive, uncontrolled) in a feedback generator as the accuracy according to majority voting by four sentiment classifiers.

\vspace{0.05cm}\noindent\textit{Notes:} [QIF-w-2] Control over Positive/Negative Sentiment captures, usually at the whole-text level, the overall tone of a text reflecting positive/negative disposition either to its topic(s) overall, or towards a specific aspect. This QC is for systems where target levels of positivity/negativity are defined (commonly positive, neutral, negative) and the system is trained to achieve them (possibly among other things including in LLMs).

\label{qcet:QIF-w-3}

\vspace{0.1cm}\noindent\qcetgrey{[QIF-w-3] Bias Inversion}: A better system produces outputs that are more of the inverse bias relative to the input.

\vspace{0.05cm}\noindent\textit{Example:} \citet{chen-etal-2018-learning} evaluate a system that inverts the political bias of a news article (between left and right) by asking human evaluators to assess whether input/output pairs have fully or partially opposite bias.

\vspace{0.05cm}\noindent\textit{Notes:} Here, target bias is usually implicit in the system task, e.g.\ the system would be trained on input-output pairs where the output has the inverse bias of the input. Alternatively, both input and output could have (inverse) bias labels.

\label{qcet:QIF-w-4}

\vspace{0.1cm}\noindent\qcetgrey{[QIF-w-4] Similarity/Dissimilarity to Input (outputs as a whole)}: A better system produces outputs that overall are (a) more similar to the input, (b) less similar to the input, or (c) more at the target level of similarity to the input, where that target level is provided in the input.

\vspace{0.05cm}\noindent\textit{Example:} \citet{panthaplackel-etal-2022-updated} ask evaluators to assess whether an updated headline makes only minimal edits to the original headline, i.e. makes changes only to parts that warrant it.

\vspace{0.05cm}\noindent\textit{Notes:} [QIF-w-4] Similarity/Dissimilarity to Input (outputs as a whole) is the same as [QEF-w-1] Similarity/Dissimilarity to Non-target Reference (outputs as a whole), but here the comparison is against the input, rather than a system-external reference. Typical NLP tasks where this QC is assessed include paraphrasing and style transfer.

\label{qcet:QIF-w-5}

\vspace{0.1cm}\noindent\qcetgrey{[QIF-w-5] Control over Multiple Attributes}: A better system produces outputs that are more at the target levels of multiple attributes provided in the input.

\vspace{0.05cm}\noindent\textit{Example:} \citet{zhong-etal-2024-benchmarking} report results for 'controllability' defined as the average classifier identification success for the controlled attributes.

\subsection{QCs that define quality in terms of one or more target outputs}\label{qcet:QT}

\begin{figure*}[ht]
        \centering
            \includegraphics[width=1.02\textwidth,trim={0cm 0cm 0.04cm 0cm},clip]{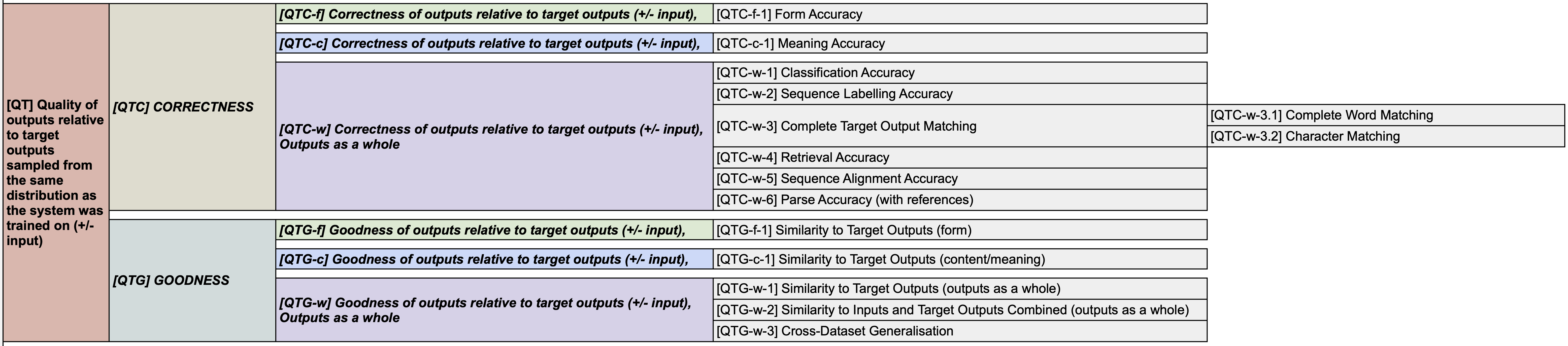}
            \caption{The \textit{Quality relative to in-distribution target outputs} branch of the QCET Taxonomy (QC IDs and names only).}
            \label{fig:rel-to-target-branch-v3.4}
        \end{figure*}

        Figure~\ref{fig:rel-to-target-branch-v3.4} shows a simplified view of the \textbf{Quality relative to in-distribution target outputs} branch of the QCET Taxonomy (only node IDs and Names are shown for each node). Below we list each QC and definition, with some additional explanatory notes in some cases, grouped according to subtrees (see Figure~\ref{fig:high-level-taxonomy}).

\subsubsection{Correctness}\label{qcet:QTC}

\subsubsection*{\textul{Form}}\label{qcet:QTC-f}

\label{qcet:QTC-f-1}

\vspace{0.1cm}\noindent\qcetgrey{[QTC-f-1] Form Accuracy}: A better system produces outputs that in their form less often differ from the given target outputs.

\vspace{0.05cm}\noindent\textit{Example:} \citet{kasner-dusek-2022-neural} report the accuracy score of their ordering model on WebNLG against the human-generated plans from Ferreira et al. (2018).

\subsubsection*{\textul{Content}}\label{qcet:QTC-c}

\label{qcet:QTC-c-1}

\vspace{0.1cm}\noindent\qcetgrey{[QTC-c-1] Meaning Accuracy}: A better system produces outputs that in their content/meaning less often differ from the given target outputs.

\vspace{0.05cm}\noindent\textit{Example:} \citet{zheng-etal-2024-neo} evaluate free definition generation in terms of the percentage of correct definitions generated, where a generated definition is correct if it is judged to be semantically equivalent to the given target output by GPT-4.

\subsubsection*{\textul{Outputs as a whole}}\label{qcet:QTC-w}

\label{qcet:QTC-w-1}

\vspace{0.1cm}\noindent\qcetgrey{[QTC-w-1] Classification Accuracy}: A better system produces output classes that less often differ from the given target output class (from a given finite set of classes).

\vspace{0.05cm}\noindent\textit{Example:} \citet{jarvis-etal-2013-maximizing} evaluate a classifier that predicts the native language of participants by computing class Recall on 11,000 texts (in 11 languages).

\vspace{0.05cm}\noindent\textit{Notes:} The notion of accuracy in this QC is wider than just the Accuracy metric, encompassing also e.g. Recall, Precision, F-score, and other combinations of true/false positives/negatives.

\label{qcet:QTC-w-2}

\vspace{0.1cm}\noindent\qcetgrey{[QTC-w-2] Sequence Labelling Accuracy}: A better system produces sequences of output labels that less often differ from the given target output label (from a given finite set of labels).

\vspace{0.05cm}\noindent\textit{Example:} \citet{de-vries-etal-2022-make} compute the part-of-speech (POS) tagging accuracy achieved by a task-tuned model on pairs of languages where one was seen during task-tuning and the other was not. They report POS tagging accuracy for a large number of pairs of languages some of which were seen during model pretraining, some were not.

\label{qcet:QTC-w-3}

\vspace{0.1cm}\noindent\qcetgrey{[QTC-w-3] Complete Target Output Matching}: A better system produces outputs that less often differ from a given target output (where the set of possible outputs is not given, and is not necessarily finite).

\vspace{0.05cm}\noindent\textit{Example:} \citet{yue-etal-2022-synthetic} report the exact match (EM) rate for question-answer generation systems, where an exact match is a question that appears in the set of target output questions verbatim.

\vspace{0.05cm}\noindent\textit{Notes:} In contrast to Classification Accuracy, this QC is defined for cases where the system does \textit{not} choose between a set of posible outputs that is known a priori. Instead the output is typically generated in some way from the input, and is often quantified as the exact match rate.

\label{qcet:QTC-w-3.1}

\vspace{0.1cm}\noindent\qcetgrey{[QTC-w-3.1] Complete Word Matching}: A better system produces a words that less often differ from corresponding given target words (where the set of possible output words is not given, and is not necessarily finite).

\vspace{0.05cm}\noindent\textit{Example:} \citet{pupier-etal-2024-growing} report the word error rate (WER) of systems in a speech recognition task.

\label{qcet:QTC-w-3.2}

\vspace{0.1cm}\noindent\qcetgrey{[QTC-w-3.2] Character Matching}: A better system produces a words that less often differ from corresponding given target words (where the set of possible output characters is not given, and is not necessarily finite).

\vspace{0.05cm}\noindent\textit{Example:} \citet{pupier-etal-2024-growing} report the character error rate of systems in a speech recognition task.

\label{qcet:QTC-w-4}

\vspace{0.1cm}\noindent\qcetgrey{[QTC-w-4] Retrieval Accuracy}: A better system produces query results that less often differ from those in a given set of target query results.

\vspace{0.05cm}\noindent\textit{Example:} \citet{cheng-etal-2024-interpreting} report Mean Reciprocal Rank for their conversational retrieval system.

\label{qcet:QTC-w-5}

\vspace{0.1cm}\noindent\qcetgrey{[QTC-w-5] Sequence Alignment Accuracy}: A better system produces alignments between two input sequences that have fewer errors compared to a given target alignment.

\vspace{0.05cm}\noindent\textit{Example:} \citet{latouche-etal-2024-binaryalign} report two word alignment metrics to assess their approach to word sequence alignment: Alignment Error Rate and the percentage of correctly aligned words.

\label{qcet:QTC-w-6}

\vspace{0.1cm}\noindent\qcetgrey{[QTC-w-6] Parse Accuracy (with references)}: A better system produces parses for input texts that have fewer errors compared to a given target parse.

\vspace{0.05cm}\noindent\textit{Example:} \citet{pupier-etal-2024-growing} report two parse accuracy metrics for their approach to end-to-end dependency parsing of speech: labelled attachment score (LAS) and unlabelled attachment score (UAS).

\subsubsection{Goodness}\label{qcet:QTG}

\subsubsection*{\textul{Form}}\label{qcet:QTG-f}

\label{qcet:QTG-f-1}

\vspace{0.1cm}\noindent\qcetgrey{[QTG-f-1] Similarity to Target Outputs (form)}: A better system produces outputs that are in their form more similar to given target outputs.

\vspace{0.05cm}\noindent\textit{Example:} \citet{gero-etal-2019-low} use BLEU to compute the similarity between (a) reconstructions of test sentences from content and feature representations, and (b) the original test sentences. Because the content is constrained to be the same, this assesses similarity of form.

\vspace{0.05cm}\noindent\textit{Notes:} Similarity to target outputs is a very common form of evaluation in NLP where one or more target outputs (often called gold outputs or references) are provided as part of a test set, and the degree of similarity between actual system output and target output(s) is measured. Note this is different from binary same/not same assessments made e.g.\ in Classification Accuracy. [QTG-f-1] Similarity to Target Outputs (form) assesses similarity in terms of form, covering aspects such as morphology, syntax, document structure, style, etc.

\subsubsection*{\textul{Content}}\label{qcet:QTG-c}

\label{qcet:QTG-c-1}

\vspace{0.1cm}\noindent\qcetgrey{[QTG-c-1] Similarity to Target Outputs (content/meaning)}: A better system produces outputs that are in their content/meaning more similar to given target outputs.

\vspace{0.05cm}\noindent\textit{Example:} In their study about experiment design for the evaluation of dialogue system ouptuts, \citet{santhanam-shaikh-2019-towards} compute the cosine similarity between embeddings of (a) system responses and (b) target system responses'.

\citet{mille-etal-2018-first} evaluate multilingual surface realisers that take syntactic or semantic trees as input by asking raters to assess the meaning similarity between system outputs and the target outputs (i.e.\ the original sentences previously parsed to get the inputs).

\vspace{0.05cm}\noindent\textit{Notes:} Similarity to target outputs is a very common form of evaluation in NLP where one or more target outputs (often called gold outputs or references) are provided as part of a test set, and the degree of similarity between actual system output and target output(s) is measured. Note this is different from binary same/not same assessments made e.g.\ in Classification Accuracy. [QTG-c-1] Similarity to Target Outputs (content/meaning) assesses similarity in terms of content units or semantic representations.

\subsubsection*{\textul{Outputs as a whole}}\label{qcet:QTG-w}

\label{qcet:QTG-w-1}

\vspace{0.1cm}\noindent\qcetgrey{[QTG-w-1] Similarity to Target Outputs (outputs as a whole)}: A better system produces outputs that are overall more similar to given target outputs.

\vspace{0.05cm}\noindent\textit{Example:} In the WebNLG shared tasks \citet{gardent-etal-2017-webnlg}, the similarity between outputs of data-to-text generators and target system outputs is evaluated using BLEU (strict n-gram matching) and METEOR (allowing synonyms and morphological variation).

\vspace{0.05cm}\noindent\textit{Notes:} Similarity to target outputs is a very common form of evaluation in NLP where one or more target outputs (often called gold outputs or references) are provided as part of a test set, and the degree of similarity between actual system output and target output(s) is measured. Note this is different from binary same/not same assessments made e.g.\ in Classification Accuracy. [QTG-w-1] Similarity to Target Outputs (outputs as a whole) assesses overall similarity, not distinguishing form or content.

\label{qcet:QTG-w-2}

\vspace{0.1cm}\noindent\qcetgrey{[QTG-w-2] Similarity to Inputs and Target Outputs Combined (outputs as a whole)}: A better system produces outputs that are overall more similar to given target outputs and more similar to the input.

\vspace{0.05cm}\noindent\textit{Example:} \citet{ingolfsdottir-etal-2023-byte} use the GLEU metric in their work on grammatical error correction to asses that their system's outputs both make the right corrections (similarity to target outputs), and do so in a way that minimally changes the text (similarity to input texts).

\label{qcet:QTG-w-3}

\vspace{0.1cm}\noindent\qcetgrey{[QTG-w-3] Cross-Dataset Generalisation}: A better system produces outputs that obtain higher scores on a given out-of-distribution task dataset.

\vspace{0.05cm}\noindent\textit{Example:} \citet{huang-etal-2024-mitigating} report work where a system is trained on a series of tasks, and after each new task training round, the system's performance (i) on the last task it was trained on, and (ii) on the next task it will be trained on, is assessed with ROUGE-L.

\vspace{0.05cm}\noindent\textit{Notes:} [QTG-w-3] captures the extent to which a system generalises beyond the data distribution on a sample from which it was trained (and in rare cases, created manually). It can be used e.g. to assess transfer learning (without any additional training). 

\subsection{QCs that define quality in terms of one or more external frames of reference}\label{qcet:QE}

\begin{figure*}[ht]
            \centering
                \includegraphics[width=1.02\textwidth,trim={0cm 0cm 0.04cm 0cm},clip]{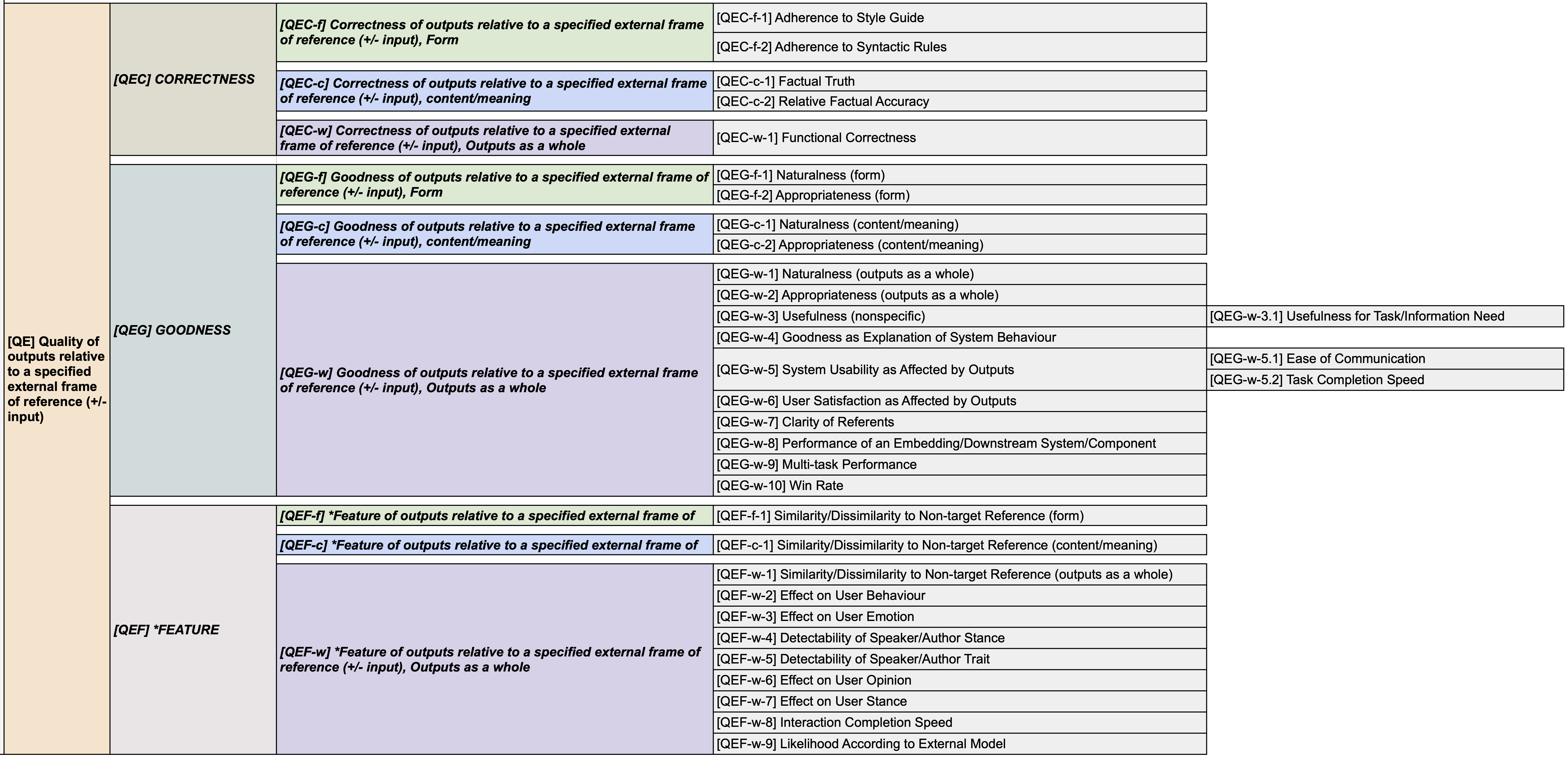}
                \caption{The \textit{Quality relative to an external frame of reference} branch of the QCET Taxonomy (QC IDs and names only).}
                \label{fig:rel-to-ext-ref-branch-v3.4}
        \end{figure*}

        Figure~\ref{fig:rel-to-ext-ref-branch-v3.4} shows a simplified view of the \textbf{Quality relative to an external frame of reference} branch of the QCET Taxonomy (only node IDs and Names are shown for each node). Below we list each QC and definition, with some additional explanatory notes in some cases, grouped according to subtrees (see Figure~\ref{fig:high-level-taxonomy}).

\subsubsection{Correctness}\label{qcet:QEC}

\subsubsection*{\textul{Form}}\label{qcet:QEC-f}

\label{qcet:QEC-f-1}

\vspace{0.1cm}\noindent\qcetgrey{[QEC-f-1] Adherence to Style Guide}: A better system produces texts with fewer deviations from a given style guide.

\vspace{0.05cm}\noindent\textit{Example:} \citet{zhang-et-al-2022-making} report an automatic code refactoring system which transforms Python code into more python-idiomatic, functionally equivalent code on the basis of a given set of idioms. They assess each refactored output e.g.\ by manually checking that it conforms with Pythonic idiom, reporting an accuracy of 0.998. 

\vspace{0.05cm}\noindent\textit{Notes:} When assessing [QEC-f-1] Adherence to Style Guide, outputs are normally compared with a system-external style guide onn the basis of which style errors can be identified in outputs, and then counted and aggregated.

\label{qcet:QEC-f-2}

\vspace{0.1cm}\noindent\qcetgrey{[QEC-f-2] Adherence to Syntactic Rules}: A better system produces outputs with fewer syntactic errors as defined by a given grammar.

\vspace{0.05cm}\noindent\textit{Example:} \citet{pratapa2021evaluatingmorphosyntacticwellformednessgenerated} present a metric to evaluate the morphosyntactic well-formedness of generated text using dependency parsing and morphosyntactic rule checking.

\vspace{0.05cm}\noindent\textit{Notes:} [QEC-f-2] Adherence to Syntactic Rules captures grammaticality, or perhaps more accurately parsability, as defined by a given formal grammar, either computational or described in a text resource. In the former case, automatic assessment can establish whether a text can be parsed with a given grammar; in the latter case, evaluators can be asked if a text conforms with the rules described in the grammar. Cf.\ [QOC-f-1] Grammaticality

\subsubsection*{\textul{Content}}\label{qcet:QEC-c}

\label{qcet:QEC-c-1}

\vspace{0.1cm}\noindent\qcetgrey{[QEC-c-1] Factual Truth}: A better system produces texts with fewer real-world untruths.

\vspace{0.05cm}\noindent\textit{Example:} \citet{thomson-reiter-2020-gold} evaluate the outputs of their sports summarisation system by asking participants (a) to mark up factual errors as determined by open web search as non-overlapping word spans, then (b) to categorise the word spans.  They report an average of 19 erorrs per summary.

\vspace{0.05cm}\noindent\textit{Notes:} In assessing [QEC-c-1] Factual Truth, the aim is to establish the real-world truth or untruth of output content. In contrast to  [QEC-c-2] Relative Factual Accuracy, specific information sources (not expected to contain contradictory information) are not normally provided in evaluation. More typically, a process is described whereby truth is to be established for the purposes of the evaluation which may involve resolving any amount of contradictory information.

\label{qcet:QEC-c-2}

\vspace{0.1cm}\noindent\qcetgrey{[QEC-c-2] Relative Factual Accuracy}: A better system produces texts with fewer untruths according to a given source of information.

\vspace{0.05cm}\noindent\textit{Example:} \citet{min2023factscore} present FACTSCORE, an estimator that decomposes generated text into atomic facts, then validates each based on a given knowledge source.

\vspace{0.05cm}\noindent\textit{Notes:} Assessing [QEC-c-2] Relative Factual Accuracy involves consulting a given source of information that is not expected to contain contradictory information (a website, a work of fiction, a database etc.), and checking if outputs are factually accurate relative to the closed world of the information source. I.e.\ unlike in [QEC-c-1] Factual Truth, there is no attempt to get at real-world truth of output content; whatever the information source states is taken as fact.

\subsubsection*{\textul{Outputs as a whole}}\label{qcet:QEC-w}

\label{qcet:QEC-w-1}

\vspace{0.1cm}\noindent\qcetgrey{[QEC-w-1] Functional Correctness}: A better system produces outputs that result in functional behaviour that less often differs from the target behaviour when applied to a given set of tests.

\vspace{0.05cm}\noindent\textit{Example:} \citet{lee-etal-2024-wrote} evaluate the functional quality of generated source code by executing the code and calculating the proportion of times generated code performs correctly on a set of unit tests, reported as pass@1 in tables.

\subsubsection{Goodness}\label{qcet:QEG}

\subsubsection*{\textul{Form}}\label{qcet:QEG-f}

\label{qcet:QEG-f-1}

\vspace{0.1cm}\noindent\qcetgrey{[QEG-f-1] Naturalness (form)}: A better system produces texts that are in their form more natural in a given context and/or for a given subset of speakers.

\vspace{0.05cm}\noindent\textit{Example:} \citet{mir2019evaluatingstyletransfertext} evaluate the naturalness of style-transferred texts via classifiers that have been adversarially trained to distinguish transferred from non-transferred texts in the same style domain. The idea is that outputs that are natural texts in the given style will be classified as non-transferred by the classifier. As the meaning is constrained to be the same, this assesses the naturalness of the form of (non-)transferred texts.

\vspace{0.05cm}\noindent\textit{Notes:} If an output is not natural in form, then it isn't likely to be encountered in this form in the given scenario. Cf. [QEG-f-2] Appropriateness: if a text is not appropriate in form then it shouldn't be in this form in the scenario.

\label{qcet:QEG-f-2}

\vspace{0.1cm}\noindent\qcetgrey{[QEG-f-2] Appropriateness (form)}: A better system produces texts that are in their form more appropriate in a given context and/or for a given subset of speakers or audience.

\vspace{0.05cm}\noindent\textit{Example:} \citet{sripada-etal-2014-case} asked experts from the UK national weather agency to assess whether individual generated forecast texts are of appropriate length. 

\vspace{0.05cm}\noindent\textit{Notes:} If an output is not appropriate in form, then it shouldn't be in this form in the given scenario. Cf. [QEG-f-1] Naturalness: if a text is not natural in form then it isn't likely to be encountered in this form in the scenario, but it's not the case that it shouldn't be.

\subsubsection*{\textul{Content}}\label{qcet:QEG-c}

\label{qcet:QEG-c-1}

\vspace{0.1cm}\noindent\qcetgrey{[QEG-c-1] Naturalness (content/meaning)}: A better system produces texts that are in their content/meaning more natural in a given context and/or for a given subset of speakers.

\vspace{0.05cm}\noindent\textit{Example:} \citet{xu2021enhancing} extract event chains from narrative texts and connect them as a graph. To evaluate the quality of the graph, they randomly sample 500 edges and calculate the proportion of edges which are ``suitable for chatting," finding that to be the case for 73.6%

\vspace{0.05cm}\noindent\textit{Notes:} If an output is not natural in content, then it isn't likely to be encountered with this content in the given scenario. Cf. [QEG-f-2] Appropriateness: if a text is not appropriate in content then it shouldn't have this content in the scenario.

\label{qcet:QEG-c-2}

\vspace{0.1cm}\noindent\qcetgrey{[QEG-c-2] Appropriateness (content/meaning)}: A better system produces texts that are in their content/meaning more appropriate in a given context and/or for a given subset of speakers or audience.

\vspace{0.05cm}\noindent\textit{Example:} \citet{mahamood-reiter-2011-generating} evaluate the impact of adding reassurance statements to automatically generated texts giving medical information to parents of pre-term newborns. Parent users rate on a scale of 1--5 the extent to which the text ``appropriately considers the parents’ emotional state in the given scenario''.

\vspace{0.05cm}\noindent\textit{Notes:} If an output is not appropriate in content, then it shouldn't have this content in the given scenario. Cf. [QEG-f-1] Naturalness: if a text is not natural in content then it isn't likely to be encountered with this content in the scenario, but it's not the case that it shouldn't be.

\subsubsection*{\textul{Outputs as a whole}}\label{qcet:QEG-w}

\label{qcet:QEG-w-1}

\vspace{0.1cm}\noindent\qcetgrey{[QEG-w-1] Naturalness (outputs as a whole)}: A better system produces texts that are more natural in a given context and/or for a given subset of speakers.

\vspace{0.05cm}\noindent\textit{Example:} In the E2E shared task \cite{dusek-etal-2018-findings}, systems generated short restaurant descriptions from a meaning representation; these were manually evaluated on a quasi-continuous scale in terms of naturalness.

\vspace{0.05cm}\noindent\textit{Notes:} If an output is not natural, then it isn't likely to be encountered in the given scenario. Cf. [QEG-f-2] Appropriateness: if an output is not appropriate then it shouldn't be used in the scenario.

\label{qcet:QEG-w-2}

\vspace{0.1cm}\noindent\qcetgrey{[QEG-w-2] Appropriateness (outputs as a whole)}: A better system produces texts that are more appropriate in a given context and/or for a given subset of speakers or audience.

\vspace{0.05cm}\noindent\textit{Example:} \citet{macdonald-siddharthan-2016-summarising} compare the outputs of two news summarisers, one producing basic summaries and the other producing summaries suitable for children. They ask human participants the following question: ``Overall, which of these summaries do you believe is more suitable for a child?''

\vspace{0.05cm}\noindent\textit{Notes:} If an output is not appropriate, then it shouldn't be used in the given scenario. Cf. [QEG-f-1] Naturalness: if an output is not natural then it isn't likely to be encountered in the scenario, but it's not the case that it shouldn't be.

\label{qcet:QEG-w-3}

\vspace{0.1cm}\noindent\qcetgrey{[QEG-w-3] Usefulness (nonspecific)}: A better system produces outputs that more useful to the user.

\vspace{0.05cm}\noindent\textit{Example:} \citet{colineau-etal-2002-evaluation} evaluate a system that generates interactive instructions for technical writers using a text editor by asking users to indicate how they would ``evaluate the usefulness of the help'' on a 6-point scale.

\vspace{0.05cm}\noindent\textit{Notes:} [QEG-w-3] Usefulness (nonspecific) is often assessed simply by asking users how useful they find the system. Cf.\ User Satisfaction as Affected by Outputs: users can be satisfied with a system (e.g.\ one whose primary purpose is entertainment) without also finding it useful. See also the more specific sub-QC [QEG-w-3.1] Usefulness for Task/Information Need.

\label{qcet:QEG-w-3.1}

\vspace{0.1cm}\noindent\qcetgrey{[QEG-w-3.1] Usefulness for Task/Information Need}: A better system produces outputs that are more useful for the user's task and/or information need.

\vspace{0.05cm}\noindent\textit{Example:} \citet{qu-green-2002-constraint} assess a cooperative mixed-initiative dialogue system for information-seeking dialogue via a user study where they measure the agreement between ``the user's recorded solution for each task'' and ``the user’s original information need'' with the kappa statistic (Carletta, 1992).

\vspace{0.05cm}\noindent\textit{Notes:} [QEG-w-3.1] Usefulness for Task/Information Need shares the same characteristics of its parent QC[QEG-w-3] Usefulness (nonspecific), but is more specific, assessing usefulness for a given task, such as following generated instructions to trouble-shoot a malfunctioning app, or for a given information need, e.g.\ are the accommodation descriptions on a website useful in selecting a holiday rental.

\label{qcet:QEG-w-4}

\vspace{0.1cm}\noindent\qcetgrey{[QEG-w-4] Goodness as Explanation of System Behaviour}: A better system produces explanations that better help the user understand its behaviour.

\vspace{0.05cm}\noindent\textit{Example:} \citet{chiyah-garcia-etal-2018-explainable} evaluate a system that generates natural language explanations for an autonomous underwater vehicle by asking evaluators to indicate (dis)agreement with the following statement on a 7-point scale: ``'Worth it' question: It would be worth reading the explanations to understand how the system is behaving.''

\vspace{0.05cm}\noindent\textit{Notes:} Cf.\ [QIG-w-3] Quality as Explanation of Input; Quality as Explanation of System Behaviour assesses how well the system can explain its own outputs or other aspects of its behaviour, e.g. what the evidence was on the basis of which it rejects an application for job or a loan.

\label{qcet:QEG-w-5}

\vspace{0.1cm}\noindent\qcetgrey{[QEG-w-5] System Usability as Affected by Outputs}: A better system is more usable by the user, where compared systems differ only in their outputs.

\vspace{0.05cm}\noindent\textit{Example:} \citet{zheng2015ease} evaluate five varied NLP systems using a software usability tool called TURF which records interaction streams and user audio; they also evaluate ease of use on a scale from -1 to 2.

\vspace{0.05cm}\noindent\textit{Notes:} In software development, Usability is commonly defined in terms of the effort required to use a system. The scope of the QCET taxonomy is evaluation of NLP systems, so usability evaluation in this context would vary just the NLP components, not e.g.\  the user interface or animation design.

\label{qcet:QEG-w-5.1}

\vspace{0.1cm}\noindent\qcetgrey{[QEG-w-5.1] Ease of Communication}: A better system produces outputs that make communication with the user easier.

\vspace{0.05cm}\noindent\textit{Example:} In their evaluation of ease of communication with a new dialogue system, \citet{rieser-etal-2011-adaptive} ask evaluators for a rating on a scale from 1 to 6 for how well users perceived they were understood by the system.

\vspace{0.05cm}\noindent\textit{Notes:} [QEG-w-5.1] Ease of Communication captures the perceived or measured ease with which users (i) convey their communicative goals to the system, and (ii) understand what the system tells them.

\label{qcet:QEG-w-5.2}

\vspace{0.1cm}\noindent\qcetgrey{[QEG-w-5.2] Task Completion Speed}: A better system produces outputs that result in faster task completion by the user.

\vspace{0.05cm}\noindent\textit{Example:} \citet{qu-green-2002-constraint} give users flight reservation tasks to evaluate two versions of a dialogue system in terms of task completion time, finding that the system-initiative version allows users to solve tasks more quickly.

\vspace{0.05cm}\noindent\textit{Notes:} This QC is assessed with a system, a user and a task that the user has to complete with the system. The speed with which the user completes the task is measured. Task Completion Speed is independ o overall interaction duration. Cf.\ [QEF-w-8] Interaction Completion Speed.

\label{qcet:QEG-w-6}

\vspace{0.1cm}\noindent\qcetgrey{[QEG-w-6] User Satisfaction as Affected by Outputs}: A better system is one that users are more satisfied with, where compared systems differ only in their outputs.

\vspace{0.05cm}\noindent\textit{Example:} \citet{mulia-etal-2023-usability} evaluate the satisfaction of users with ChatGPT via (dis)agreement with the statement ``I think that I would like to use this system frequently," finding a mean agreement score of 3.95 out of 5.

\vspace{0.05cm}\noindent\textit{Notes:} User Satisfaction is about how happy users are with their use of a system, and is commonly assessed by asking users directly how satisfied they are, or via tracking repeat use. In the present context of NLP system evaluation, evaluations would take into account variations of the NLP component(s), but not other aspects like other functionality, user interface, etc.

\label{qcet:QEG-w-7}

\vspace{0.1cm}\noindent\qcetgrey{[QEG-w-7] Clarity of Referents}: A better system produces referring expressions that more clearly identify their referents.

\vspace{0.05cm}\noindent\textit{Example:} In the 2005 DUC shared task on summarisation, \citet{dang2005overview} manually assess systems in terms of `Referential clarity,' defined as follows: ``It should be easy to identify who or what the pronouns and noun phrases in the summary are referring to. If a person or other entity is mentioned, it should be clear what their role in the story is. So, a reference would be unclear if an entity is referenced but its identity or relation to the story remains unclear.''

\vspace{0.05cm}\noindent\textit{Notes:} Clarity of referents is about how readily intended referents can be identified from referring expressions in texts or other representations. The explanation included in the DUC 2005 attestation provides a good explanation for the case of texts.

\label{qcet:QEG-w-8}

\vspace{0.1cm}\noindent\qcetgrey{[QEG-w-8] Performance of an Embedding/Downstream System/Component}: A better system produces outputs that result in outputs with better performance when used by another system or component.

\vspace{0.05cm}\noindent\textit{Example:} \citet{reddy-etal-2017-generating} evaluate a system that generates question-answer pairs from keywords by assessing whether its outputs can improve the performance of a semantic parser when added to its training data. Comparing performance when (a) training on manual question-answer pairs only with (b) training on the manual data plus the system's question-answer pairs, they find a 5.5\% improvement for the augmented training data.

\vspace{0.05cm}\noindent\textit{Notes:} Assessing a system in terms of the impact its outputs have on the performance of the bigger system it is part of, or in terms of another system that uses its outputs, is often called an extrinsic form of evaluation. Extrinsic evaluation is especially suitable for NLP components embedded in a larger system, such as a TTS component that is part of an interactive system.

\label{qcet:QEG-w-9}

\vspace{0.1cm}\noindent\qcetgrey{[QEG-w-9] Multi-task Performance}: A better system produces outputs that obtain higher aggregated scores on a given set of task datasets and metrics.

\vspace{0.05cm}\noindent\textit{Example:} \citet{zhou2023dontmakellmevaluation} explore data leakage in LLM assessment, evaluating different models on the MMLU benchmark of 57 different tasks that require real-world knowledge and problem-solving abilities.

\vspace{0.05cm}\noindent\textit{Notes:} Multi-task benchmarks have become increasingly common in NLP, particularly in LLM evaluation. [QEG-w-10] Multi-task Performance covers any case where aggregated results expressing performance at multiple tasks is reported.

\label{qcet:QEG-w-10}

\vspace{0.1cm}\noindent\qcetgrey{[QEG-w-10] Win Rate}: A better system produces outputs that more often beat a given comparitor system or system stand-in.

\vspace{0.05cm}\noindent\textit{Example:} \citet{wang-etal-2024-uncertainty} evaluate a model by measuring the fraction of times a powerful LLM (e.g. GPT-4) prefers the outputs from that model over outputs from a reference model.

\subsubsection{Feature}\label{qcet:QEF}

\subsubsection*{\textul{Form}}\label{qcet:QEF-f}

\label{qcet:QEF-f-1}

\vspace{0.1cm}\noindent\qcetgrey{[QEF-f-1] Similarity/Dissimilarity to Non-target Reference (form)}: A better system produces outputs that are in their form (a) more similar, (b) less similar, or (c) more at the target level of similarity, compared to given references that are not target system outputs.

\vspace{0.05cm}\noindent\textit{Example:} \citet{chim2024evaluating} assess various types of similarity of synthetic user-generated text with `known source data' that was included in training data for the generating model. One type of similarity they look at is idiolect preservation where they measure the cosine similarity between idiolect embeddings that reflect stylistic idiosyncrasies to capture style preservation relative to the source data. 

\vspace{0.05cm}\noindent\textit{Notes:} We use the term `non-target references' here to refer to texts, speech, or structured representations which system outputs are compared against, but which do not have the status of a target system output for the given input. In the case of [QEF-f-1] Similarity/Dissimilarity to Non-target Reference (form), such comparisons assess form similarity. This is the case e.g.\ when the similarity of outputs to a sample of multiple texts, speech, or structured representations (that are not target outputs) is measured as an indication of stylistic or other form alignment.

\subsubsection*{\textul{Content}}\label{qcet:QEF-c}

\label{qcet:QEF-c-1}

\vspace{0.1cm}\noindent\qcetgrey{[QEF-c-1] Similarity/Dissimilarity to Non-target Reference (content/meaning)}: A better system produces outputs that are in their content/meaning (a) more similar, (b) less similar, or (c) more at the target level of similarity, compared to given references that are not target system outputs.

\vspace{0.05cm}\noindent\textit{Example:} \citet{10.1371/journal.pone.0254034} propose a new metric to assess the degree of novelty of scientific articles, which uses the distance relative to embeddings of cited articles, and show that a larger average distance correlates with a higher level of novelty.

\vspace{0.05cm}\noindent\textit{Notes:} We use the term `non-target references' here to refer to texts, speech, or structured representations which system outputs are compared against, but which do not have the status of a target system output for the given input. In the case of [QEF-f-1] Similarity/Dissimilarity to Non-target Reference (content/meaning), such comparisons assess content/meaning similarity. This is the case e.g.\ when the similarity of outputs to a sample of multiple texts, speech, or structured representations (that are not target outputs) is measured as an indication of topic or other semantic alignment.

\subsubsection*{\textul{Outputs as a whole}}\label{qcet:QEF-w}

\label{qcet:QEF-w-1}

\vspace{0.1cm}\noindent\qcetgrey{[QEF-w-1] Similarity/Dissimilarity to Non-target Reference (outputs as a whole)}: A better system produces outputs that are overall (a) more similar, (b) less similar, or (c) more at the target level of similarity, compared to given references that are not target system outputs.

\vspace{0.05cm}\noindent\textit{Example:} \citet{chim2024evaluating} assess various types of similarity of synthetic user-generated text with `known source data' that was included in training data for the generating model. One type of similarity they look at is divergence (text overlap) as an intrinsic proxy for privacy preservation, for which they compute the BLEU score between source-data text and synthetic text, reporting divergence per data point as 1--BLEU(s, t).

\vspace{0.05cm}\noindent\textit{Notes:} We use the term `non-target references' here to refer to texts, speech, or structured representations which system outputs are compared against, but which do not have the status of a target system output for the given input. In the case of [QEF-f-1] Similarity/Dissimilarity to Non-target Reference (outputs as a whole), such comparisons assess overall output similarity. This is the case e.g.\ when the similarity of outputs to a sample of multiple texts, speech, or structured representations (that are not target outputs) is measured as an indication of overall alignment.

\label{qcet:QEF-w-2}

\vspace{0.1cm}\noindent\qcetgrey{[QEF-w-2] Effect on User Behaviour}: A better system produces outputs that affect the user's behaviour (a) more, (b) less, or (c) as specified in the input, in terms of a given range of possible behaviours.

\vspace{0.05cm}\noindent\textit{Example:} \citet{davis2020process} evaluate a virtual health assistant \textit{inter alia} in terms of user diet adherence and overall goal achievement, finding that mean dietary adherence was 91\% and was lowest for discretionary foods, grains, red meat, and vegetables. Participants met their step goal 59\% of the time.

\vspace{0.05cm}\noindent\textit{Notes:} [QEF-w-2] Effect on User Behaviour captures changes in the behaviour of users as a result of using the system. Cf.\ [QEF-w-3] Effect on User Emotion; [QEF-w-6] Effect on User Opinion; and [QEF-w-7] Effect on User Stance. Examples of behaviours include driving behaviour, diet adherence, smoking, and exercise.

\label{qcet:QEF-w-3}

\vspace{0.1cm}\noindent\qcetgrey{[QEF-w-3] Effect on User Emotion}: A better system produces outputs that affect the user's emotions (a) more, (b) less, or (c) as specified in the input, in terms of a given range of possible emotions.

\vspace{0.05cm}\noindent\textit{Example:} \citet{van-der-sluis-mellish-2009-towards} evaluate text generation strategies aimed at emphasising positive feedback in mixed-feedback texts via lexical and syntactic choice in terms of the effect on users' emotions. One type of measure asks users to rate the strength with which they are feeling different emotions before and after receiving feedback (in the form of human-written stand-ins); they find that the system version using positive emphasis strategies has a distinct positive effect, increasing mean ratings for all tested emotions.

\vspace{0.05cm}\noindent\textit{Notes:} [QEF-w-3] Effect on User Emotion captures changes in the emotional state of users as a result of using the system. Cf.\ [QEF-w-2] Effect on User Behaviour; [QEF-w-6] Effect on User Opinion; and [QEF-w-7] Effect on User Stance. An example of a set of emotion classes is Ekman et al.'s six cross-cultural basic emotions: anger, disgust, fear, happiness, sadness and surprise.

\label{qcet:QEF-w-4}

\vspace{0.1cm}\noindent\qcetgrey{[QEF-w-4] Detectability of Speaker/Author Stance}: A better system produces outputs that make the entity producing the output come across to an observer as having one of a range of given stances (a) more, (b) less, or (c) to the degree specified in the input.

\vspace{0.05cm}\noindent\textit{Example:} \citet{van-der-lee-etal-2017-pass} evaluate a system that generates football game summaries for (a) supporters of the home team, and (b) supporters of the away team, by asking evaluators who they they think the summaries are intended for, finding that evaluators identified the correct team in 91\% of all cases.

\vspace{0.05cm}\noindent\textit{Notes:} [QEF-w-4] Detectability of Speaker/Author Stance is about the degree to which the user perceives the entity producing the outputs (which may be perceived as an interlocutor) as having a given stance towards a given object. In contrast to trait ([QEF-w-5] Detectability of Speaker/Author Trait), a stance cannot be expressed in a single adjective. An example is (strength of) support for a sports team, as in the example attestation.

\label{qcet:QEF-w-5}

\vspace{0.1cm}\noindent\qcetgrey{[QEF-w-5] Detectability of Speaker/Author Trait}: A better system produces outputs that make the entity producing the output come across to an observer as having one of a range of given traits (a) more, (b) less, or (c) to the degree specified in the input.

\vspace{0.05cm}\noindent\textit{Example:} \citet{glas-pelachaud-2015-topic} evaluate different strategies for dialogue agents to introduce new topics into conversation, assessing which makes the user perceive the agent as more (a) competent, (b) friendly, (c) fun, and (d) informed, i.e.\ four different traits. 

\vspace{0.05cm}\noindent\textit{Notes:} [QEF-w-5] Detectability of Speaker/Author Trait is about the degree to which the user perceives the entity producing the outputs (which may be perceived as an interlocutor) as having a given trait. A trait in this context is usually something that can be captured in a single adjective, as in the example attestation.

\label{qcet:QEF-w-6}

\vspace{0.1cm}\noindent\qcetgrey{[QEF-w-6] Effect on User Opinion}: A better system produces outputs that affect the user's opinions (a) more, (b) less, or (c) as specified in the input, in terms of a given range of possible opinions.

\vspace{0.05cm}\noindent\textit{Example:} \citet{wu2023explaining} evaluated six versions of a web search system in terms of the extent to which user opinions change on several topics after system use. Opinions for and against are captured at the start and the end on a +6 to -6 scale and opinion change is computed as the difference between the value at the start and the value at the end.

\vspace{0.05cm}\noindent\textit{Notes:} [QEF-w-6] Effect on User Opinion captures changes in user opinion about a given topic as a result of using the system. Cf.\ [QEF-w-2] Effect on User Behaviour; [QEF-w-3] Effect on User Emotion; and [QEF-w-7] Effect on User Stance. Opinion is more fine-grained than stance, and while an opinion can be typical of a stance, it would take more than one opinion to identify a stance. E.g.\ favouring the introduction of a universal income is a political opinion, not an overall political stance.

\label{qcet:QEF-w-7}

\vspace{0.1cm}\noindent\qcetgrey{[QEF-w-7] Effect on User Stance}: A better system produces outputs that affect the user's stance (a) more, (b) less, or (c) as specified in the input, in terms of a given range of possible stances.

\vspace{0.05cm}\noindent\textit{Example:} \citet{forrest-etal-2018-towards} compare explainability components for use with machine learning tools where components use either (i) language generation, or (ii) more numbers-based forms of explanation, by asking users if they ``would trust a decision with this explanation," finding that the components that use language-generation inspire more trust in users.

\vspace{0.05cm}\noindent\textit{Notes:} [QEF-w-7] Effect on User Stance captures changes in user stance as a result of using the system. Cf.\ [QEF-w-2] Effect on User Behaviour; [QEF-w-3] Effect on User Emotion; and [QEF-w-6] Effect on User Opinion. Stance is more coarse-grained than opinion, typically comprising multiple related opinions. E.g.\ Left-leaning and right-leaning are political stances, not in themselves political opinions.

\label{qcet:QEF-w-8}

\vspace{0.1cm}\noindent\qcetgrey{[QEF-w-8] Interaction Completion Speed}: A better system produces outputs that result in interactions with the user completing in (a) more time, (b) less time, or (c) as much time as specified in the input.

\vspace{0.05cm}\noindent\textit{Example:} \citet{peng2017composite} evaluate four dialogue systems in terms of the average number of turns in simulated user interactions with three user types.

\vspace{0.05cm}\noindent\textit{Notes:} [QEF-w-8] Interaction Completion Speed captures how quickly user-system interactions are completed. This can be measured in different ways, including number of turns, number of system turns, length of turns altogether, or clock time. Note that in many cases, a better system will have faster interaction completion speed, but in other cases the aim may be to keep the user interacting for as long as possible.

\label{qcet:QEF-w-9}

\vspace{0.1cm}\noindent\qcetgrey{[QEF-w-9] Likelihood According to External Model}: A better system produces outputs that are estimated to be more likely by a given external model.

\vspace{0.05cm}\noindent\textit{Example:} \citet{yedetore2023poor} train various models (5-gram model, LSTMs, Transformers) on child-directed language data, and use perplexity (a standard formulation as well as the word-frequency normalised SLOR metric) to evaluate how well each model captures the basic structure of the training domain, finding that Transformers have the lowest perplexity, the 5-gram model the highest.

\vspace{0.05cm}\noindent\textit{Notes:} The more common use of perplexity is in evaluations where \textit{low} perplexity as computed with a given model is desirable, where it's seen as indicative of `natural' output. However, it can equally be desirable for outputs to have high perplexity, e.g.\ in situation where a different style from that encapsulated by the model is intended.  [QEF-w-9] Model Perplexity is a Feature-type QC, hence captures both possibilities. Note that various metrics exist for measuring model perplexity including normalised ones such as SLOR.

\bibliography{latex/bib/custom,latex/bib/first_60,latex/bib/second_60,latex/bib/extracted_citations}

\begin{thebibliography}{122}
\providecommand{\natexlab}[1]{#1}

\bibitem[{Abu~Sheikha and Inkpen(2011)}]{abu-sheikha-inkpen-2011-generation}
Fadi Abu~Sheikha and Diana Inkpen. 2011.
\newblock \href {https://aclanthology.org/W11-2826/} {Generation of formal and informal sentences}.
\newblock In \emph{Proceedings of the 13th {E}uropean Workshop on Natural Language Generation}, pages 187--193, Nancy, France. Association for Computational Linguistics.

\bibitem[{Afsal and Kuppusamy(2024)}]{afsal-kuppusamy-2024-assessing-readability}
C.P Afsal and K~S Kuppusamy. 2024.
\newblock \href {https://doi.org/10.1109/ICCCNT61001.2024.10724250} {Assessing the readability and coherence in gemini’s triple draft generation: A multi-metric approach}.
\newblock In \emph{2024 15th International Conference on Computing Communication and Networking Technologies (ICCCNT)}, pages 1--7.

\bibitem[{Agrawal(2023)}]{agrawal2023complexity}
Sweta Agrawal. 2023.
\newblock \emph{Complexity Controlled Natural Language Generation}.
\newblock Ph.D. thesis, University of Maryland, College Park.

\bibitem[{Alnajjar and H{\"a}m{\"a}l{\"a}inen(2018)}]{alnajjar-hamalainen-2018-master}
Khalid Alnajjar and Mika H{\"a}m{\"a}l{\"a}inen. 2018.
\newblock \href {https://doi.org/10.18653/v1/W18-6534} {A master-apprentice approach to automatic creation of culturally satirical movie titles}.
\newblock In \emph{Proceedings of the 11th International Conference on Natural Language Generation}, pages 274--283, Tilburg University, The Netherlands. Association for Computational Linguistics.

\bibitem[{Angrosh and Siddharthan(2014)}]{angrosh-siddharthan-2014-text}
Mandya Angrosh and Advaith Siddharthan. 2014.
\newblock \href {https://doi.org/10.3115/v1/W14-4404} {Text simplification using synchronous dependency grammars: Generalising automatically harvested rules}.
\newblock In \emph{Proceedings of the 8th International Natural Language Generation Conference ({INLG})}, pages 16--25, Philadelphia, Pennsylvania, U.S.A. Association for Computational Linguistics.

\bibitem[{Basu Roy~Chowdhury et~al.(2022)Basu Roy~Chowdhury, Zhao, and Chaturvedi}]{basu-roy-chowdhury-etal-2022-unsupervised}
Somnath Basu Roy~Chowdhury, Chao Zhao, and Snigdha Chaturvedi. 2022.
\newblock \href {https://doi.org/10.18653/v1/2022.acl-long.86} {Unsupervised extractive opinion summarization using sparse coding}.
\newblock In \emph{Proceedings of the 60th Annual Meeting of the Association for Computational Linguistics (Volume 1: Long Papers)}, pages 1209--1225, Dublin, Ireland. Association for Computational Linguistics.

\bibitem[{Belz et~al.(2020)Belz, Mille, and Howcroft}]{belz-etal-2020-disentangling}
Anya Belz, Simon Mille, and David~M. Howcroft. 2020.
\newblock \href {https://www.aclweb.org/anthology/2020.inlg-1.24} {Disentangling the properties of human evaluation methods: A classification system to support comparability, meta-evaluation and reproducibility testing}.
\newblock In \emph{Proceedings of the 13th International Conference on Natural Language Generation}, pages 183--194, Dublin, Ireland. Association for Computational Linguistics.

\bibitem[{Belz et~al.(to appear)Belz, Mille, and Thomson}]{belz-et-al-2025-120papers}
Anya Belz, Simon Mille, and Craig Thomson. to appear.
\newblock Standardising evaluation criterion names and definitions in nlp via systematic surveys.

\bibitem[{Belz et~al.(2024)Belz, Mille, Thomson, and Huidrom}]{belz2024qcet}
Anya Belz, Simon Mille, Craig Thomson, and Rudali Huidrom. 2024.
\newblock {QCET}: An interactive taxonomy of quality criteria for comparable and repeatable evaluation of nlp systems.
\newblock In \emph{Proceedings of the 17th International Natural Language Generation Conference: System Demonstrations}, pages 9--12.

\bibitem[{Belz et~al.(2023{\natexlab{a}})Belz, Thomson, Reiter, Abercrombie, Alonso-Moral, Arvan, Braggaar, Cieliebak, Clark, van Deemter, Dinkar, Du{\v{s}}ek, Eger, Fang, Gao, Gatt, Gkatzia, Gonz{\'a}lez-Corbelle, Hovy, H{\"u}rlimann, Ito, Kelleher, Klubicka, Krahmer, Lai, van~der Lee, Li, Mahamood, Mieskes, van Miltenburg, Mosteiro, Nissim, Parde, Pl{\'a}tek, Rieser, Ruan, Tetreault, Toral, Wan, Wanner, Watson, and Yang}]{belz-et-al-2023-missing}
Anya Belz, Craig Thomson, Ehud Reiter, Gavin Abercrombie, Jose~M. Alonso-Moral, Mohammad Arvan, Anouck Braggaar, Mark Cieliebak, Elizabeth Clark, Kees van Deemter, Tanvi Dinkar, Ond{\v{r}}ej Du{\v{s}}ek, Steffen Eger, Qixiang Fang, Mingqi Gao, Albert Gatt, Dimitra Gkatzia, Javier Gonz{\'a}lez-Corbelle, Dirk Hovy, Manuela H{\"u}rlimann, Takumi Ito, John~D. Kelleher, Filip Klubicka, Emiel Krahmer, Huiyuan Lai, Chris van~der Lee, Yiru Li, Saad Mahamood, Margot Mieskes, Emiel van Miltenburg, Pablo Mosteiro, Malvina Nissim, Natalie Parde, Ond{\v{r}}ej Pl{\'a}tek, Verena Rieser, Jie Ruan, Joel Tetreault, Antonio Toral, Xiaojun Wan, Leo Wanner, Lewis Watson, and Diyi Yang. 2023{\natexlab{a}}.
\newblock \href {https://doi.org/10.18653/v1/2023.insights-1.1} {Missing information, unresponsive authors, experimental flaws: The impossibility of assessing the reproducibility of previous human evaluations in {NLP}}.
\newblock In \emph{The Fourth Workshop on Insights from Negative Results in NLP}, pages 1--10, Dubrovnik, Croatia. Association for Computational Linguistics.

\bibitem[{Belz et~al.(2023{\natexlab{b}})Belz, Thomson, Reiter, and Mille}]{belz2023non}
Anya Belz, Craig Thomson, Ehud Reiter, and Simon Mille. 2023{\natexlab{b}}.
\newblock Non-repeatable experiments and non-reproducible results: The reproducibility crisis in human evaluation in {NLP}.
\newblock In \emph{Findings of the Association for Computational Linguistics: ACL 2023}, pages 3676--3687.

\bibitem[{Cercas~Curry et~al.(2015)Cercas~Curry, Gkatzia, and Rieser}]{cercas-curry-etal-2015-generating}
Amanda Cercas~Curry, Dimitra Gkatzia, and Verena Rieser. 2015.
\newblock \href {https://doi.org/10.18653/v1/W15-4715} {Generating and evaluating landmark-based navigation instructions in virtual environments}.
\newblock In \emph{Proceedings of the 15th {E}uropean Workshop on Natural Language Generation ({ENLG})}, pages 90--94, Brighton, UK. Association for Computational Linguistics.

\bibitem[{Cervone et~al.(2019)Cervone, Khatri, Goel, Hedayatnia, Venkatesh, Hakkani-Tur, and Gabriel}]{cervone-etal-2019-natural}
Alessandra Cervone, Chandra Khatri, Rahul Goel, Behnam Hedayatnia, Anu Venkatesh, Dilek Hakkani-Tur, and Raefer Gabriel. 2019.
\newblock \href {https://doi.org/10.18653/v1/W19-8657} {Natural language generation at scale: A case study for open domain question answering}.
\newblock In \emph{Proceedings of the 12th International Conference on Natural Language Generation}, pages 453--462, Tokyo, Japan. Association for Computational Linguistics.

\bibitem[{Chen et~al.(2023)Chen, Li, Sun, and Liu}]{chen-etal-2023-weakly}
Chi Chen, Peng Li, Maosong Sun, and Yang Liu. 2023.
\newblock \href {https://doi.org/10.18653/v1/2023.acl-long.464} {Weakly supervised vision-and-language pre-training with relative representations}.
\newblock In \emph{Proceedings of the 61st Annual Meeting of the Association for Computational Linguistics (Volume 1: Long Papers)}, pages 8341--8355, Toronto, Canada. Association for Computational Linguistics.

\bibitem[{Chen et~al.(2018)Chen, Wachsmuth, Al-Khatib, and Stein}]{chen-etal-2018-learning}
Wei-Fan Chen, Henning Wachsmuth, Khalid Al-Khatib, and Benno Stein. 2018.
\newblock \href {https://doi.org/10.18653/v1/W18-6509} {Learning to flip the bias of news headlines}.
\newblock In \emph{Proceedings of the 11th International Conference on Natural Language Generation}, pages 79--88, Tilburg University, The Netherlands. Association for Computational Linguistics.

\bibitem[{Chen et~al.(2024)Chen, Gr{\"o}ner, Zarrie{\ss}, and Eger}]{chen-etal-2024-evaluating-diversity}
Yanran Chen, Hannes Gr{\"o}ner, Sina Zarrie{\ss}, and Steffen Eger. 2024.
\newblock \href {https://doi.org/10.18653/v1/2024.emnlp-main.1097} {Evaluating diversity in automatic poetry generation}.
\newblock In \emph{Proceedings of the 2024 Conference on Empirical Methods in Natural Language Processing}, pages 19671--19692, Miami, Florida, USA. Association for Computational Linguistics.

\bibitem[{Cheng et~al.(2024)Cheng, Mao, and Dou}]{cheng-etal-2024-interpreting}
Yiruo Cheng, Kelong Mao, and Zhicheng Dou. 2024.
\newblock \href {https://doi.org/10.18653/v1/2024.acl-long.159} {Interpreting conversational dense retrieval by rewriting-enhanced inversion of session embedding}.
\newblock In \emph{Proceedings of the 62nd Annual Meeting of the Association for Computational Linguistics (Volume 1: Long Papers)}, pages 2879--2893, Bangkok, Thailand. Association for Computational Linguistics.

\bibitem[{Chim et~al.(2024)Chim, Ive, and Liakata}]{chim2024evaluating}
Jenny Chim, Julia Ive, and Maria Liakata. 2024.
\newblock Evaluating synthetic data generation from user generated text.
\newblock \emph{Computational Linguistics}, pages 1--44.

\bibitem[{Chiyah~Garcia et~al.(2018)Chiyah~Garcia, Robb, Liu, Laskov, Patron, and Hastie}]{chiyah-garcia-etal-2018-explainable}
Francisco~Javier Chiyah~Garcia, David~A. Robb, Xingkun Liu, Atanas Laskov, Pedro Patron, and Helen Hastie. 2018.
\newblock \href {https://doi.org/10.18653/v1/W18-6511} {Explainable autonomy: A study of explanation styles for building clear mental models}.
\newblock In \emph{Proceedings of the 11th International Conference on Natural Language Generation}, pages 99--108, Tilburg University, The Netherlands. Association for Computational Linguistics.

\bibitem[{Clarke and Lapata(2006)}]{clarke-lapata-2006-models}
James Clarke and Mirella Lapata. 2006.
\newblock \href {https://doi.org/10.3115/1220175.1220223} {Models for sentence compression: A comparison across domains, training requirements and evaluation measures}.
\newblock In \emph{Proceedings of the 21st International Conference on Computational Linguistics and 44th Annual Meeting of the Association for Computational Linguistics}, pages 377--384, Sydney, Australia. Association for Computational Linguistics.

\bibitem[{Clinciu et~al.(2021)Clinciu, Eshghi, and Hastie}]{clinciu2021study}
Miruna Clinciu, Arash Eshghi, and Helen Hastie. 2021.
\newblock A study of automatic metrics for the evaluation of natural language explanations.
\newblock In \emph{Proceedings of the 16th Conference of the European Chapter of the Association for Computational Linguistics: Main Volume}, pages 2376--2387.

\bibitem[{Cohen et~al.(2018)Cohen, Xia, Zweigenbaum, Callahan, Hargraves, Goss, Ide, N{\'e}v{\'e}ol, Grouin, and Hunter}]{cohen-etal-2018-thee-dimensions}
K~Bretonnel Cohen, Jingbo Xia, Pierre Zweigenbaum, Tiffany~J Callahan, Orin Hargraves, Foster Goss, Nancy Ide, Aur{\'e}lie N{\'e}v{\'e}ol, Cyril Grouin, and Lawrence~E Hunter. 2018.
\newblock Three dimensions of reproducibility in natural language processing.
\newblock \emph{LREC Int Conf Lang Resour Eval}, 2018:156--165.

\bibitem[{Colineau et~al.(2002)Colineau, Paris, and Vander~Linden}]{colineau-etal-2002-evaluation}
Nathalie Colineau, Cecile Paris, and Keith Vander~Linden. 2002.
\newblock \href {https://aclanthology.org/W02-2117/} {An evaluation of procedural instructional text}.
\newblock In \emph{Proceedings of the International Natural Language Generation Conference}, pages 128--135, Harriman, New York, USA. Association for Computational Linguistics.

\bibitem[{Cripwell et~al.(2023)Cripwell, Belz, Gardent, Gatt, Borg, Borg, Judge, Lorandi, Nikiforovskaya, Soto-Martinez et~al.}]{cripwell20232023}
Liam Cripwell, Anya Belz, Claire Gardent, Albert Gatt, Claudia Borg, Marthese Borg, John Judge, Michela Lorandi, Anna Nikiforovskaya, William Soto-Martinez, et~al. 2023.
\newblock The 2023 webnlg shared task on low resource languages overview and evaluation results (webnlg 2023).
\newblock In \emph{Proceedings of the Workshop on Multimodal, Multilingual Natural Language Generation and Multilingual WebNLG Challenge (MM-NLG 2023)}.

\bibitem[{Dang(2005)}]{dang2005overview}
Hoa~Trang Dang. 2005.
\newblock Overview of duc 2005.
\newblock In \emph{Proceedings of the document understanding conference}, volume 2005, pages 1--12. Citeseer.

\bibitem[{Davis et~al.(2020)Davis, Murphy, Curtis, and Maher}]{davis2020process}
Courtney~R Davis, Karen~J Murphy, Rachel~G Curtis, and Carol~A Maher. 2020.
\newblock A process evaluation examining the performance, adherence, and acceptability of a physical activity and diet artificial intelligence virtual health assistant.
\newblock \emph{International journal of environmental research and public health}, 17(23):9137.

\bibitem[{de~Vries et~al.(2022)de~Vries, Wieling, and Nissim}]{de-vries-etal-2022-make}
Wietse de~Vries, Martijn Wieling, and Malvina Nissim. 2022.
\newblock \href {https://doi.org/10.18653/v1/2022.acl-long.529} {Make the best of cross-lingual transfer: Evidence from {POS} tagging with over 100 languages}.
\newblock In \emph{Proceedings of the 60th Annual Meeting of the Association for Computational Linguistics (Volume 1: Long Papers)}, pages 7676--7685, Dublin, Ireland. Association for Computational Linguistics.

\bibitem[{Di~Fabbrizio et~al.(2014)Di~Fabbrizio, Stent, and Gaizauskas}]{di-fabbrizio-etal-2014-hybrid}
Giuseppe Di~Fabbrizio, Amanda Stent, and Robert Gaizauskas. 2014.
\newblock \href {https://doi.org/10.3115/v1/W14-4408} {A hybrid approach to multi-document summarization of opinions in reviews}.
\newblock In \emph{Proceedings of the 8th International Natural Language Generation Conference ({INLG})}, pages 54--63, Philadelphia, Pennsylvania, U.S.A. Association for Computational Linguistics.

\bibitem[{Du{\v{s}}ek et~al.(2018)Du{\v{s}}ek, Novikova, and Rieser}]{dusek-etal-2018-findings}
Ond{\v{r}}ej Du{\v{s}}ek, Jekaterina Novikova, and Verena Rieser. 2018.
\newblock \href {https://doi.org/10.18653/v1/W18-6539} {Findings of the {E}2{E} {NLG} challenge}.
\newblock In \emph{Proceedings of the 11th International Conference on Natural Language Generation}, pages 322--328, Tilburg University, The Netherlands. Association for Computational Linguistics.

\bibitem[{Eisenstadt and Elhadad(2020)}]{eisentadt-elhadad-2020-neural}
Roy Eisenstadt and Michael Elhadad. 2020.
\newblock \href {https://aclanthology.org/2020.dt4tp-1.2/} {Neural micro-planning for data to text generation produces more cohesive text}.
\newblock In \emph{Proceedings of the Workshop on Discourse Theories for Text Planning}, pages 6--9, Dublin, Ireland. Association for Computational Linguistics.

\bibitem[{Farr{\'u}s~Cabeceran et~al.(2010)Farr{\'u}s~Cabeceran, Ruiz Costa-Juss{\`a}, Mari{\~n}o~Acebal, and Rodr{\'\i}guez~Fonollosa}]{farrus2010linguistic}
Mireia Farr{\'u}s~Cabeceran, Marta Ruiz Costa-Juss{\`a}, Jos{\'e}~Bernardo Mari{\~n}o~Acebal, and Jos{\'e}~Adri{\'a}n Rodr{\'\i}guez~Fonollosa. 2010.
\newblock Linguistic-based evaluation criteria to identify statistical machine translation errors.
\newblock In \emph{14th Annual Conference of the European Association for Machine Translation}, pages 167--173.

\bibitem[{Fong(2024)}]{fong2024controlling}
Jason Fong. 2024.
\newblock \emph{Controlling text-to-speech pronunciation using limited linguistic resources}.
\newblock Ph.D. thesis, The University of Edinburgh.

\bibitem[{Forrest et~al.(2018)Forrest, Sripada, Pang, and Coghill}]{forrest-etal-2018-towards}
James Forrest, Somayajulu Sripada, Wei Pang, and George Coghill. 2018.
\newblock \href {https://doi.org/10.18653/v1/W18-6522} {Towards making {NLG} a voice for interpretable machine learning}.
\newblock In \emph{Proceedings of the 11th International Conference on Natural Language Generation}, pages 177--182, Tilburg University, The Netherlands. Association for Computational Linguistics.

\bibitem[{Gardent et~al.(2017)Gardent, Shimorina, Narayan, and Perez-Beltrachini}]{gardent-etal-2017-webnlg}
Claire Gardent, Anastasia Shimorina, Shashi Narayan, and Laura Perez-Beltrachini. 2017.
\newblock \href {https://doi.org/10.18653/v1/W17-3518} {The {W}eb{NLG} challenge: Generating text from {RDF} data}.
\newblock In \emph{Proceedings of the 10th International Conference on Natural Language Generation}, pages 124--133, Santiago de Compostela, Spain. Association for Computational Linguistics.

\bibitem[{Gehrmann et~al.(2023)Gehrmann, Clark, and Sellam}]{gehrmann2023repairing}
Sebastian Gehrmann, Elizabeth Clark, and Thibault Sellam. 2023.
\newblock Repairing the cracked foundation: A survey of obstacles in evaluation practices for generated text.
\newblock \emph{Journal of Artificial Intelligence Research}, 77:103--166.

\bibitem[{Gero et~al.(2019)Gero, Kedzie, Reeve, and Chilton}]{gero-etal-2019-low}
Katy Gero, Chris Kedzie, Jonathan Reeve, and Lydia Chilton. 2019.
\newblock \href {https://doi.org/10.18653/v1/W19-8628} {Low level linguistic controls for style transfer and content preservation}.
\newblock In \emph{Proceedings of the 12th International Conference on Natural Language Generation}, pages 208--218, Tokyo, Japan. Association for Computational Linguistics.

\bibitem[{Ghalandari et~al.(2022)Ghalandari, Hokamp, and Ifrim}]{ghalandari-etal-2022-efficient}
Demian Ghalandari, Chris Hokamp, and Georgiana Ifrim. 2022.
\newblock \href {https://doi.org/10.18653/v1/2022.acl-long.90} {Efficient unsupervised sentence compression by fine-tuning transformers with reinforcement learning}.
\newblock In \emph{Proceedings of the 60th Annual Meeting of the Association for Computational Linguistics (Volume 1: Long Papers)}, pages 1267--1280, Dublin, Ireland. Association for Computational Linguistics.

\bibitem[{Glas and Pelachaud(2015)}]{glas-pelachaud-2015-topic}
Nadine Glas and Catherine Pelachaud. 2015.
\newblock \href {https://doi.org/10.18653/v1/W15-4725} {Topic transition strategies for an information-giving agent}.
\newblock In \emph{Proceedings of the 15th {E}uropean Workshop on Natural Language Generation ({ENLG})}, pages 146--155, Brighton, UK. Association for Computational Linguistics.

\bibitem[{Gonz{\'a}lez~Corbelle et~al.(2022)Gonz{\'a}lez~Corbelle, Bugar{\'i}n-Diz, Alonso-Moral, and Taboada}]{gonzalez-corbelle-etal-2022-dealing}
Javier Gonz{\'a}lez~Corbelle, Alberto Bugar{\'i}n-Diz, Jose Alonso-Moral, and Juan Taboada. 2022.
\newblock \href {https://doi.org/10.18653/v1/2022.inlg-main.10} {Dealing with hallucination and omission in neural natural language generation: A use case on meteorology.}
\newblock In \emph{Proceedings of the 15th International Conference on Natural Language Generation}, pages 121--130, Waterville, Maine, USA and virtual meeting. Association for Computational Linguistics.

\bibitem[{Green(2006)}]{green-2006-generation}
Nancy Green. 2006.
\newblock \href {https://aclanthology.org/W06-1417/} {Generation of biomedical arguments for lay readers}.
\newblock In \emph{Proceedings of the Fourth International Natural Language Generation Conference}, pages 114--121, Sydney, Australia. Association for Computational Linguistics.

\bibitem[{Harrison and Walker(2018)}]{harrison-walker-2018-neural}
Vrindavan Harrison and Marilyn Walker. 2018.
\newblock \href {https://doi.org/10.18653/v1/W18-6536} {Neural generation of diverse questions using answer focus, contextual and linguistic features}.
\newblock In \emph{Proceedings of the 11th International Conference on Natural Language Generation}, pages 296--306, Tilburg University, The Netherlands. Association for Computational Linguistics.

\bibitem[{Hershenhouse et~al.(2024)Hershenhouse, Mokhtar, Eppler, Rodler, Storino~Ramacciotti, Ganjavi, Hom, Davis, Tran, Russo et~al.}]{hershenhouse2024accuracy}
Jacob~S Hershenhouse, Daniel Mokhtar, Michael~B Eppler, Severin Rodler, Lorenzo Storino~Ramacciotti, Conner Ganjavi, Brian Hom, Ryan~J Davis, John Tran, Giorgio~Ivan Russo, et~al. 2024.
\newblock Accuracy, readability, and understandability of large language models for prostate cancer information to the public.
\newblock \emph{Prostate Cancer and Prostatic Diseases}, pages 1--6.

\bibitem[{Howcroft et~al.(2020)Howcroft, Belz, Clinciu, Gkatzia, Hasan, Mahamood, Mille, Santhanam, {van~Miltenburg}, and Rieser}]{howcroft:etal:2020}
David Howcroft, Anya Belz, Miruna Clinciu, Dimitra Gkatzia, Sadid Hasan, Saad Mahamood, Simon Mille, Sashank Santhanam, Emiel {van~Miltenburg}, and Verena Rieser. 2020.
\newblock Twenty years of confusion in human evaluation: {NLG} needs evaluation sheets and standardised definitions.
\newblock In \emph{Proceedings of the 13th International Natural Language Generation Conference}.

\bibitem[{Hu and Loizou(2007)}]{hu2007subjective}
Yi~Hu and Philipos~C Loizou. 2007.
\newblock Subjective comparison and evaluation of speech enhancement algorithms.
\newblock \emph{Speech communication}, 49(7-8):588--601.

\bibitem[{Hu et~al.(2022)Hu, Lee, Aggarwal, and Zhang}]{hu-etal-2022-text-style-transfer}
Zhiqiang Hu, Roy Ka-Wei Lee, Charu~C. Aggarwal, and Aston Zhang. 2022.
\newblock \href {https://doi.org/10.1145/3544903.3544906} {Text style transfer: A review and experimental evaluation}.
\newblock \emph{SIGKDD Explor. Newsl.}, 24(1):14–45.

\bibitem[{Huang et~al.(2024)Huang, Cui, Wang, Yang, Liao, Song, Yao, and Su}]{huang-etal-2024-mitigating}
Jianheng Huang, Leyang Cui, Ante Wang, Chengyi Yang, Xinting Liao, Linfeng Song, Junfeng Yao, and Jinsong Su. 2024.
\newblock \href {https://doi.org/10.18653/v1/2024.acl-long.77} {Mitigating catastrophic forgetting in large language models with self-synthesized rehearsal}.
\newblock In \emph{Proceedings of the 62nd Annual Meeting of the Association for Computational Linguistics (Volume 1: Long Papers)}, pages 1416--1428, Bangkok, Thailand. Association for Computational Linguistics.

\bibitem[{Humphreys et~al.(2001)Humphreys, Calcagno, and Weise}]{humphreys-etal-2001-reusing}
Kevin Humphreys, Mike Calcagno, and David Weise. 2001.
\newblock \href {https://aclanthology.org/W01-0812/} {Reusing a statistical language model for generation}.
\newblock In \emph{Proceedings of the {ACL} 2001 Eighth {E}uropean Workshop on Natural Language Generation ({EWNLG})}, Toulouse, France. Association for Computational Linguistics.

\bibitem[{Imperial et~al.(2024)Imperial, Forey, and Madabushi}]{imperial2024standardizealigninglanguagemodels}
Joseph~Marvin Imperial, Gail Forey, and Harish~Tayyar Madabushi. 2024.
\newblock \href {https://arxiv.org/abs/2402.12593} {Standardize: Aligning language models with expert-defined standards for content generation}.
\newblock \emph{Preprint}, arXiv:2402.12593.

\bibitem[{Ing{\'o}lfsd{\'o}ttir et~al.(2023)Ing{\'o}lfsd{\'o}ttir, Ragnarsson, J{\'o}nsson, Simonarson, Thorsteinsson, and Sn{\ae}bjarnarson}]{ingolfsdottir-etal-2023-byte}
Svanhv{\'i}t~Lilja Ing{\'o}lfsd{\'o}ttir, Petur Ragnarsson, Haukur J{\'o}nsson, Haukur Simonarson, Vilhjalmur Thorsteinsson, and V{\'e}steinn Sn{\ae}bjarnarson. 2023.
\newblock \href {https://doi.org/10.18653/v1/2023.acl-long.402} {Byte-level grammatical error correction using synthetic and curated corpora}.
\newblock In \emph{Proceedings of the 61st Annual Meeting of the Association for Computational Linguistics (Volume 1: Long Papers)}, pages 7299--7316, Toronto, Canada. Association for Computational Linguistics.

\bibitem[{Jarvis et~al.(2013)Jarvis, Bestgen, and Pepper}]{jarvis-etal-2013-maximizing}
Scott Jarvis, Yves Bestgen, and Steve Pepper. 2013.
\newblock \href {https://aclanthology.org/W13-1714/} {Maximizing classification accuracy in native language identification}.
\newblock In \emph{Proceedings of the Eighth Workshop on Innovative Use of {NLP} for Building Educational Applications}, pages 111--118, Atlanta, Georgia. Association for Computational Linguistics.

\bibitem[{Jin and Le(2016)}]{jin-le-2016-selecting}
Yiping Jin and Phu Le. 2016.
\newblock \href {https://doi.org/10.18653/v1/W16-6623} {Selecting domain-specific concepts for question generation with lightly-supervised methods}.
\newblock In \emph{Proceedings of the 9th International Natural Language Generation conference}, pages 133--142, Edinburgh, UK. Association for Computational Linguistics.

\bibitem[{Juraska et~al.(2019)Juraska, Bowden, and Walker}]{juraska-etal-2019-viggo}
Juraj Juraska, Kevin Bowden, and Marilyn Walker. 2019.
\newblock \href {https://doi.org/10.18653/v1/W19-8623} {{V}i{GGO}: A video game corpus for data-to-text generation in open-domain conversation}.
\newblock In \emph{Proceedings of the 12th International Conference on Natural Language Generation}, pages 164--172, Tokyo, Japan. Association for Computational Linguistics.

\bibitem[{Kamath et~al.(2024)Kamath, Same, and Mahamood}]{kamath-etal-2024-generating-hotel}
Srinivas~Ramesh Kamath, Fahime Same, and Saad Mahamood. 2024.
\newblock \href {https://aclanthology.org/2024.inlg-main.23/} {Generating hotel highlights from unstructured text using {LLM}s}.
\newblock In \emph{Proceedings of the 17th International Natural Language Generation Conference}, pages 280--288, Tokyo, Japan. Association for Computational Linguistics.

\bibitem[{Kasner and Dusek(2022)}]{kasner-dusek-2022-neural}
Zden{\v{e}}k Kasner and Ondrej Dusek. 2022.
\newblock \href {https://doi.org/10.18653/v1/2022.acl-long.271} {Neural pipeline for zero-shot data-to-text generation}.
\newblock In \emph{Proceedings of the 60th Annual Meeting of the Association for Computational Linguistics (Volume 1: Long Papers)}, pages 3914--3932, Dublin, Ireland. Association for Computational Linguistics.

\bibitem[{Kumar et~al.(2020)Kumar, Ahuja, Vadapalli, and Talukdar}]{kumar2020syntax}
Ashutosh Kumar, Kabir Ahuja, Raghuram Vadapalli, and Partha Talukdar. 2020.
\newblock Syntax-guided controlled generation of paraphrases.
\newblock \emph{Transactions of the Association for Computational Linguistics}, 8:330--345.

\bibitem[{Kumar et~al.(2024)Kumar, Malik, Raman, and Li}]{kumar2024synthesizingsentimentcontrolledfeedbackmultimodal}
Puneet Kumar, Sarthak Malik, Balasubramanian Raman, and Xiaobai Li. 2024.
\newblock \href {https://arxiv.org/abs/2402.07640} {Synthesizing sentiment-controlled feedback for multimodal text and image data}.
\newblock \emph{Preprint}, arXiv:2402.07640.

\bibitem[{Latouche et~al.(2024)Latouche, Carbonneau, and Swanson}]{latouche-etal-2024-binaryalign}
Gaetan Latouche, Marc-Andr{\'e} Carbonneau, and Benjamin Swanson. 2024.
\newblock \href {https://doi.org/10.18653/v1/2024.acl-long.553} {{B}inary{A}lign: Word alignment as binary sequence labeling}.
\newblock In \emph{Proceedings of the 62nd Annual Meeting of the Association for Computational Linguistics (Volume 1: Long Papers)}, pages 10277--10288, Bangkok, Thailand. Association for Computational Linguistics.

\bibitem[{Lee et~al.(2024)Lee, Hong, Ahn, Hong, Lee, Yun, Shin, and Kim}]{lee-etal-2024-wrote}
Taehyun Lee, Seokhee Hong, Jaewoo Ahn, Ilgee Hong, Hwaran Lee, Sangdoo Yun, Jamin Shin, and Gunhee Kim. 2024.
\newblock \href {https://doi.org/10.18653/v1/2024.acl-long.268} {Who wrote this code? watermarking for code generation}.
\newblock In \emph{Proceedings of the 62nd Annual Meeting of the Association for Computational Linguistics (Volume 1: Long Papers)}, pages 4890--4911, Bangkok, Thailand. Association for Computational Linguistics.

\bibitem[{Li et~al.(2015)Li, Galley, Brockett, Gao, and Dolan}]{li2015diversity}
Jiwei Li, Michel Galley, Chris Brockett, Jianfeng Gao, and Bill Dolan. 2015.
\newblock A diversity-promoting objective function for neural conversation models.
\newblock \emph{arXiv preprint arXiv:1510.03055}.

\bibitem[{Lindberg et~al.(2013)Lindberg, Popowich, Nesbit, and Winne}]{lindberg-etal-2013-generating}
David Lindberg, Fred Popowich, John Nesbit, and Phil Winne. 2013.
\newblock \href {https://aclanthology.org/W13-2114/} {Generating natural language questions to support learning on-line}.
\newblock In \emph{Proceedings of the 14th {E}uropean Workshop on Natural Language Generation}, pages 105--114, Sofia, Bulgaria. Association for Computational Linguistics.

\bibitem[{Luo et~al.(2024)Luo, Xie, and Ananiadou}]{luo2024laypersonsguidebiomedicine}
Zheheng Luo, Qianqian Xie, and Sophia Ananiadou. 2024.
\newblock \href {https://arxiv.org/abs/2402.13498} {The lay person's guide to biomedicine: Orchestrating large language models}.
\newblock \emph{Preprint}, arXiv:2402.13498.

\bibitem[{Macdonald and Siddharthan(2016)}]{macdonald-siddharthan-2016-summarising}
Iain Macdonald and Advaith Siddharthan. 2016.
\newblock \href {https://doi.org/10.18653/v1/W16-6601} {Summarising news stories for children}.
\newblock In \emph{Proceedings of the 9th International Natural Language Generation conference}, pages 1--10, Edinburgh, UK. Association for Computational Linguistics.

\bibitem[{Mahamood and Reiter(2011)}]{mahamood-reiter-2011-generating}
Saad Mahamood and Ehud Reiter. 2011.
\newblock \href {https://aclanthology.org/W11-2803/} {Generating affective natural language for parents of neonatal infants}.
\newblock In \emph{Proceedings of the 13th {E}uropean Workshop on Natural Language Generation}, pages 12--21, Nancy, France. Association for Computational Linguistics.

\bibitem[{Mille et~al.(2018)Mille, Belz, Bohnet, Graham, Pitler, and Wanner}]{mille-etal-2018-first}
Simon Mille, Anya Belz, Bernd Bohnet, Yvette Graham, Emily Pitler, and Leo Wanner. 2018.
\newblock \href {https://doi.org/10.18653/v1/W18-3601} {The first multilingual surface realisation shared task ({SR}`18): Overview and evaluation results}.
\newblock In \emph{Proceedings of the First Workshop on Multilingual Surface Realisation}, pages 1--12, Melbourne, Australia. Association for Computational Linguistics.

\bibitem[{Min et~al.(2023)Min, Krishna, Lyu, Lewis, Yih, Koh, Iyyer, Zettlemoyer, and Hajishirzi}]{min2023factscore}
Sewon Min, Kalpesh Krishna, Xinxi Lyu, Mike Lewis, Wen-tau Yih, Pang~Wei Koh, Mohit Iyyer, Luke Zettlemoyer, and Hannaneh Hajishirzi. 2023.
\newblock Factscore: Fine-grained atomic evaluation of factual precision in long form text generation.
\newblock \emph{arXiv preprint arXiv:2305.14251}.

\bibitem[{Mir et~al.(2019)Mir, Felbo, Obradovich, and Rahwan}]{mir2019evaluatingstyletransfertext}
Remi Mir, Bjarke Felbo, Nick Obradovich, and Iyad Rahwan. 2019.
\newblock \href {https://arxiv.org/abs/1904.02295} {Evaluating style transfer for text}.
\newblock \emph{Preprint}, arXiv:1904.02295.

\bibitem[{Mohan(2021)}]{mohanmodifying}
Raghav Mohan. 2021.
\newblock Modifying visual explanations to improve the user understand-ability of explainable artificial intelligence systems. master's thesis.

\bibitem[{Moraes et~al.(2016)Moraes, McCoy, and Carberry}]{moraes-etal-2016-enabling}
Priscilla Moraes, Kathleen McCoy, and Sandra Carberry. 2016.
\newblock \href {https://doi.org/10.18653/v1/W16-6621} {Enabling text readability awareness during the micro planning phase of {NLG} applications}.
\newblock In \emph{Proceedings of the 9th International Natural Language Generation conference}, pages 121--131, Edinburgh, UK. Association for Computational Linguistics.

\bibitem[{Mulia et~al.(2023)Mulia, Piri, and Tho}]{mulia-etal-2023-usability}
Angelina~Patience Mulia, Pirelli~Rahelya Piri, and Cuk Tho. 2023.
\newblock \href {https://doi.org/10.1016/j.procs.2023.10.537} {Usability analysis of text generation by chatgpt openai using system usability scale method}.
\newblock \emph{Procedia Comput. Sci.}, 227(C):381–388.

\bibitem[{Novikova et~al.(2018)Novikova, Du{\v{s}}ek, and Rieser}]{novikova-etal-2018-rankme}
Jekaterina Novikova, Ond{\v{r}}ej Du{\v{s}}ek, and Verena Rieser. 2018.
\newblock \href {https://doi.org/10.18653/v1/N18-2012} {{R}ank{ME}: Reliable human ratings for natural language generation}.
\newblock In \emph{Proceedings of the 2018 Conference of the North {A}merican Chapter of the Association for Computational Linguistics: Human Language Technologies, Volume 2 (Short Papers)}, pages 72--78, New Orleans, Louisiana. Association for Computational Linguistics.

\bibitem[{Onoe et~al.(2024)Onoe, Rane, Berger, Bitton, Cho, Garg, Ku, Parekh, Pont-Tuset, Tanzer, Wang, and Baldridge}]{onoe2024doccidescriptionsconnectedcontrasting}
Yasumasa Onoe, Sunayana Rane, Zachary Berger, Yonatan Bitton, Jaemin Cho, Roopal Garg, Alexander Ku, Zarana Parekh, Jordi Pont-Tuset, Garrett Tanzer, Su~Wang, and Jason Baldridge. 2024.
\newblock \href {https://arxiv.org/abs/2404.19753} {Docci: Descriptions of connected and contrasting images}.
\newblock \emph{Preprint}, arXiv:2404.19753.

\bibitem[{Opitz and Frank(2022)}]{opitz-frank-2022-better}
Juri Opitz and Anette Frank. 2022.
\newblock \href {https://doi.org/10.18653/v1/2022.eval4nlp-1.4} {Better {S}match = better parser? {AMR} evaluation is not so simple anymore}.
\newblock In \emph{Proceedings of the 3rd Workshop on Evaluation and Comparison of NLP Systems}, pages 32--43, Online. Association for Computational Linguistics.

\bibitem[{Pan et~al.(2020)Pan, Xie, Feng, Chua, and Kan}]{pan-etal-2020-semantic}
Liangming Pan, Yuxi Xie, Yansong Feng, Tat-Seng Chua, and Min-Yen Kan. 2020.
\newblock \href {https://doi.org/10.18653/v1/2020.acl-main.135} {Semantic graphs for generating deep questions}.
\newblock In \emph{Proceedings of the 58th Annual Meeting of the Association for Computational Linguistics}, pages 1463--1475, Online. Association for Computational Linguistics.

\bibitem[{Panthaplackel et~al.(2022)Panthaplackel, Benton, and Dredze}]{panthaplackel-etal-2022-updated}
Sheena Panthaplackel, Adrian Benton, and Mark Dredze. 2022.
\newblock \href {https://doi.org/10.18653/v1/2022.acl-long.446} {Updated headline generation: Creating updated summaries for evolving news stories}.
\newblock In \emph{Proceedings of the 60th Annual Meeting of the Association for Computational Linguistics (Volume 1: Long Papers)}, pages 6438--6461, Dublin, Ireland. Association for Computational Linguistics.

\bibitem[{Peng et~al.(2017)Peng, Li, Li, Gao, Celikyilmaz, Lee, and Wong}]{peng2017composite}
Baolin Peng, Xiujun Li, Lihong Li, Jianfeng Gao, Asli Celikyilmaz, Sungjin Lee, and Kam-Fai Wong. 2017.
\newblock Composite task-completion dialogue policy learning via hierarchical deep reinforcement learning.
\newblock \emph{arXiv preprint arXiv:1704.03084}.

\bibitem[{Peng et~al.(2024)Peng, Huang, Li, Mohamed, and Harwath}]{peng-etal-2024-voicecraft}
Puyuan Peng, Po-Yao Huang, Shang-Wen Li, Abdelrahman Mohamed, and David Harwath. 2024.
\newblock \href {https://doi.org/10.18653/v1/2024.acl-long.673} {{V}oice{C}raft: Zero-shot speech editing and text-to-speech in the wild}.
\newblock In \emph{Proceedings of the 62nd Annual Meeting of the Association for Computational Linguistics (Volume 1: Long Papers)}, pages 12442--12462, Bangkok, Thailand. Association for Computational Linguistics.

\bibitem[{Popovi{\'c}(2020)}]{popovic-2020-informative}
Maja Popovi{\'c}. 2020.
\newblock \href {https://doi.org/10.18653/v1/2020.coling-main.444} {Informative manual evaluation of machine translation output}.
\newblock In \emph{Proceedings of the 28th International Conference on Computational Linguistics}, pages 5059--5069, Barcelona, Spain (Online). International Committee on Computational Linguistics.

\bibitem[{Pratapa et~al.(2021)Pratapa, Anastasopoulos, Rijhwani, Chaudhary, Mortensen, Neubig, and Tsvetkov}]{pratapa2021evaluatingmorphosyntacticwellformednessgenerated}
Adithya Pratapa, Antonios Anastasopoulos, Shruti Rijhwani, Aditi Chaudhary, David~R. Mortensen, Graham Neubig, and Yulia Tsvetkov. 2021.
\newblock \href {https://arxiv.org/abs/2103.16590} {Evaluating the morphosyntactic well-formedness of generated texts}.
\newblock \emph{Preprint}, arXiv:2103.16590.

\bibitem[{Pupier et~al.(2024)Pupier, Coavoux, Goulian, and Lecouteux}]{pupier-etal-2024-growing}
Adrien Pupier, Maximin Coavoux, J{\'e}r{\^o}me Goulian, and Benjamin Lecouteux. 2024.
\newblock \href {https://doi.org/10.18653/v1/2024.acl-short.22} {Growing trees on sounds: Assessing strategies for end-to-end dependency parsing of speech}.
\newblock In \emph{Proceedings of the 62nd Annual Meeting of the Association for Computational Linguistics (Volume 2: Short Papers)}, pages 225--233, Bangkok, Thailand. Association for Computational Linguistics.

\bibitem[{Purdy et~al.(2019)Purdy, Coppens, Madden, Mozeiko, Patterson, Wallace, and Freed}]{purdy2019reading}
Mary Purdy, Patrick Coppens, Elizabeth~Brookshire Madden, Jennifer Mozeiko, Janet Patterson, Sarah~E Wallace, and Donald Freed. 2019.
\newblock Reading comprehension treatment in aphasia: A systematic review.
\newblock \emph{Aphasiology}, 33(6):629--651.

\bibitem[{Qu and Green(2002)}]{qu-green-2002-constraint}
Yan Qu and Nancy Green. 2002.
\newblock \href {https://www.aclweb.org/anthology/W02-2118} {A constraint-based approach for cooperative information-seeking dialogue}.
\newblock In \emph{Proceedings of the International Natural Language Generation Conference}, pages 136--143, Harriman, New York, USA. Association for Computational Linguistics.

\bibitem[{Rafatbakhsh et~al.(2021)Rafatbakhsh, Ahmadi, Moloodi, and Mehrpour}]{rafatbakhsh2021development}
Elaheh Rafatbakhsh, Alireza Ahmadi, Amirsaeid Moloodi, and Saeed Mehrpour. 2021.
\newblock Development and validation of an automatic item generation system for english idioms.
\newblock \emph{Educational Measurement: Issues and Practice}, 40(2):49--59.

\bibitem[{Reddy et~al.(2017)Reddy, Raghu, Khapra, and Joshi}]{reddy-etal-2017-generating}
Sathish Reddy, Dinesh Raghu, Mitesh~M. Khapra, and Sachindra Joshi. 2017.
\newblock \href {https://aclanthology.org/E17-1036/} {Generating natural language question-answer pairs from a knowledge graph using a {RNN} based question generation model}.
\newblock In \emph{Proceedings of the 15th Conference of the {E}uropean Chapter of the Association for Computational Linguistics: Volume 1, Long Papers}, pages 376--385, Valencia, Spain. Association for Computational Linguistics.

\bibitem[{Resendiz and Klinger(2025)}]{resendiz2025mopo}
Yarik~Menchaca Resendiz and Roman Klinger. 2025.
\newblock Mopo: Multi-objective prompt optimization for affective text generation.
\newblock In \emph{Proceedings of the 31st International Conference on Computational Linguistics}, pages 5588--5606.

\bibitem[{Rieser et~al.(2011)Rieser, Keizer, Lemon, and Liu}]{rieser-etal-2011-adaptive}
Verena Rieser, Simon Keizer, Oliver Lemon, and Xingkun Liu. 2011.
\newblock \href {https://aclanthology.org/W11-2813/} {Adaptive information presentation for spoken dialogue systems: Evaluation with real users}.
\newblock In \emph{Proceedings of the 13th {E}uropean Workshop on Natural Language Generation}, pages 102--109, Nancy, France. Association for Computational Linguistics.

\bibitem[{Ruan et~al.(2024)Ruan, Wang, and Wan}]{ruan-etal-2024-defining}
Jie Ruan, Wenqing Wang, and Xiaojun Wan. 2024.
\newblock \href {https://doi.org/10.18653/v1/2024.naacl-long.441} {Defining and detecting vulnerability in human evaluation guidelines: A preliminary study towards reliable {NLG} evaluation}.
\newblock In \emph{Proceedings of the 2024 Conference of the North American Chapter of the Association for Computational Linguistics: Human Language Technologies (Volume 1: Long Papers)}, pages 7965--7989, Mexico City, Mexico. Association for Computational Linguistics.

\bibitem[{Saggion et~al.(2015)Saggion, {\v{S}}tajner, Bott, Mille, Rello, and Drndarevic}]{saggion2015making}
Horacio Saggion, Sanja {\v{S}}tajner, Stefan Bott, Simon Mille, Luz Rello, and Biljana Drndarevic. 2015.
\newblock Making it simplext: Implementation and evaluation of a text simplification system for spanish.
\newblock \emph{ACM Transactions on Accessible Computing (TACCESS)}, 6(4):1--36.

\bibitem[{Santhanam and Shaikh(2019)}]{santhanam-shaikh-2019-towards}
Sashank Santhanam and Samira Shaikh. 2019.
\newblock \href {https://doi.org/10.18653/v1/W19-8610} {Towards best experiment design for evaluating dialogue system output}.
\newblock In \emph{Proceedings of the 12th International Conference on Natural Language Generation}, pages 88--94, Tokyo, Japan. Association for Computational Linguistics.

\bibitem[{Sarsa et~al.(2022)Sarsa, Denny, Hellas, and Leinonen}]{sarsa-etal-code-explanations-2022}
Sami Sarsa, Paul Denny, Arto Hellas, and Juho Leinonen. 2022.
\newblock \href {https://doi.org/10.1145/3501385.3543957} {Automatic generation of programming exercises and code explanations using large language models}.
\newblock In \emph{Proceedings of the 2022 ACM Conference on International Computing Education Research - Volume 1}, ICER '22, page 27–43, New York, NY, USA. Association for Computing Machinery.

\bibitem[{Sasazawa et~al.(2023)Sasazawa, Morishita, Ozaki, Imaichi, and Sogawa}]{sasazawa2023controllingkeywordspositionstext}
Yuichi Sasazawa, Terufumi Morishita, Hiroaki Ozaki, Osamu Imaichi, and Yasuhiro Sogawa. 2023.
\newblock \href {https://arxiv.org/abs/2304.09516} {Controlling keywords and their positions in text generation}.
\newblock \emph{Preprint}, arXiv:2304.09516.

\bibitem[{Schmidtova et~al.(2024)Schmidtova, Mahamood, Balloccu, Dusek, Gatt, Gkatzia, Howcroft, Platek, and Sivaprasad}]{schmidtova-etal-2024-automatic-metrics}
Patricia Schmidtova, Saad Mahamood, Simone Balloccu, Ondrej Dusek, Albert Gatt, Dimitra Gkatzia, David~M. Howcroft, Ondrej Platek, and Adarsa Sivaprasad. 2024.
\newblock \href {https://aclanthology.org/2024.inlg-main.44/} {Automatic metrics in natural language generation: A survey of current evaluation practices}.
\newblock In \emph{Proceedings of the 17th International Natural Language Generation Conference}, pages 557--583, Tokyo, Japan. Association for Computational Linguistics.

\bibitem[{Shen et~al.(2022)Shen, Perez-Rosas, Welch, Poria, and Mihalcea}]{shen-etal-2022-knowledge}
Siqi Shen, Veronica Perez-Rosas, Charles Welch, Soujanya Poria, and Rada Mihalcea. 2022.
\newblock \href {https://doi.org/10.18653/v1/2022.acl-long.221} {Knowledge enhanced reflection generation for counseling dialogues}.
\newblock In \emph{Proceedings of the 60th Annual Meeting of the Association for Computational Linguistics (Volume 1: Long Papers)}, pages 3096--3107, Dublin, Ireland. Association for Computational Linguistics.

\bibitem[{Shibayama et~al.(2021)Shibayama, Yin, and Matsumoto}]{10.1371/journal.pone.0254034}
Sotaro Shibayama, Deyun Yin, and Kuniko Matsumoto. 2021.
\newblock \href {https://doi.org/10.1371/journal.pone.0254034} {Measuring novelty in science with word embedding}.
\newblock \emph{PLOS ONE}, 16(7):1--16.

\bibitem[{Shu et~al.(2021)Shu, Zhang, Dong, Shi, Yu, and Zhang}]{shu2021logicconsistencytextgenerationsemantic}
Chang Shu, Yusen Zhang, Xiangyu Dong, Peng Shi, Tao Yu, and Rui Zhang. 2021.
\newblock \href {https://arxiv.org/abs/2108.00577} {Logic-consistency text generation from semantic parses}.
\newblock \emph{Preprint}, arXiv:2108.00577.

\bibitem[{{Sparck~Jones}(1981)}]{sparkjones:k:1981}
Karen {Sparck~Jones}. 1981.
\newblock \emph{Information Retrieval Experiment}, illustrated edition.
\newblock Butterworths.

\bibitem[{Sripada et~al.(2014)Sripada, Burnett, Turner, Mastin, and Evans}]{sripada-etal-2014-case}
Somayajulu Sripada, Neil Burnett, Ross Turner, John Mastin, and Dave Evans. 2014.
\newblock \href {https://doi.org/10.3115/v1/W14-4401} {A case study: {NLG} meeting weather industry demand for quality and quantity of textual weather forecasts}.
\newblock In \emph{Proceedings of the 8th International Natural Language Generation Conference ({INLG})}, pages 1--5, Philadelphia, Pennsylvania, U.S.A. Association for Computational Linguistics.

\bibitem[{Thomson and Reiter(2020)}]{thomson-reiter-2020-gold}
Craig Thomson and Ehud Reiter. 2020.
\newblock \href {https://doi.org/10.18653/v1/2020.inlg-1.22} {A gold standard methodology for evaluating accuracy in data-to-text systems}.
\newblock In \emph{Proceedings of the 13th International Conference on Natural Language Generation}, pages 158--168, Dublin, Ireland. Association for Computational Linguistics.

\bibitem[{Van~de Cruys(2020)}]{van2020automatic}
Tim Van~de Cruys. 2020.
\newblock Automatic poetry generation from prosaic text.
\newblock In \emph{Proceedings of the 58th Annual Meeting of the Association for Computational Linguistics}, pages 2471--2480.

\bibitem[{van~der Lee et~al.(2021)van~der Lee, Gatt, van Miltenburg, and Krahmer}]{van2021human}
Chris van~der Lee, Albert Gatt, Emiel van Miltenburg, and Emiel Krahmer. 2021.
\newblock Human evaluation of automatically generated text: Current trends and best practice guidelines.
\newblock \emph{Computer Speech \& Language}, 67:101151.

\bibitem[{van~der Lee et~al.(2019)van~der Lee, Gatt, van Miltenburg, Wubben, and Krahmer}]{van-der-lee-etal-2019-best}
Chris van~der Lee, Albert Gatt, Emiel van Miltenburg, Sander Wubben, and Emiel Krahmer. 2019.
\newblock \href {https://doi.org/10.18653/v1/W19-8643} {Best practices for the human evaluation of automatically generated text}.
\newblock In \emph{Proceedings of the 12th International Conference on Natural Language Generation}, pages 355--368, Tokyo, Japan. Association for Computational Linguistics.

\bibitem[{van~der Lee et~al.(2017)van~der Lee, Krahmer, and Wubben}]{van-der-lee-etal-2017-pass}
Chris van~der Lee, Emiel Krahmer, and Sander Wubben. 2017.
\newblock \href {https://doi.org/10.18653/v1/W17-3513} {{PASS}: A {D}utch data-to-text system for soccer, targeted towards specific audiences}.
\newblock In \emph{Proceedings of the 10th International Conference on Natural Language Generation}, pages 95--104, Santiago de Compostela, Spain. Association for Computational Linguistics.

\bibitem[{van~der Sluis and Mellish(2009)}]{van-der-sluis-mellish-2009-towards}
Ielka van~der Sluis and Chris Mellish. 2009.
\newblock \href {https://aclanthology.org/W09-0625/} {Towards empirical evaluation of affective tactical {NLG}}.
\newblock In \emph{Proceedings of the 12th {E}uropean Workshop on Natural Language Generation ({ENLG} 2009)}, pages 146--153, Athens, Greece. Association for Computational Linguistics.

\bibitem[{Voorhees et~al.(1999)Voorhees, Tice et~al.}]{voorhees1999trec}
Ellen~M Voorhees, Dawn~M Tice, et~al. 1999.
\newblock The trec-8 question answering track evaluation.
\newblock In \emph{TREC}, volume 1999, page~82.

\bibitem[{Wang et~al.(2024{\natexlab{a}})Wang, Li, Lian, Ma, Song, and Wei}]{wang2024mitigating}
Weichuan Wang, Zhaoyi Li, Defu Lian, Chen Ma, Linqi Song, and Ying Wei. 2024{\natexlab{a}}.
\newblock Mitigating the language mismatch and repetition issues in llm-based machine translation via model editing.
\newblock \emph{arXiv preprint arXiv:2410.07054}.

\bibitem[{Wang et~al.(2024{\natexlab{b}})Wang, Zheng, Ding, Zhang, Lin, and Tao}]{wang-etal-2024-uncertainty}
Yikun Wang, Rui Zheng, Liang Ding, Qi~Zhang, Dahua Lin, and Dacheng Tao. 2024{\natexlab{b}}.
\newblock \href {https://doi.org/10.18653/v1/2024.acl-long.597} {Uncertainty aware learning for language model alignment}.
\newblock In \emph{Proceedings of the 62nd Annual Meeting of the Association for Computational Linguistics (Volume 1: Long Papers)}, pages 11087--11099, Bangkok, Thailand. Association for Computational Linguistics.

\bibitem[{Webb et~al.(2010)Webb, Benyon, Hansen, and Mival}]{webb-etal-2010-evaluating}
Nick Webb, David Benyon, Preben Hansen, and Oil Mival. 2010.
\newblock \href {https://aclanthology.org/L10-1071/} {Evaluating human-machine conversation for appropriateness}.
\newblock In \emph{Proceedings of the Seventh International Conference on Language Resources and Evaluation ({LREC}`10)}, Valletta, Malta. European Language Resources Association (ELRA).

\bibitem[{Wu et~al.(2019)Wu, Song, Sakai, Cheng, Xie, and Lin}]{wu2019evaluating}
Chao-Chung Wu, Ruihua Song, Tetsuya Sakai, Wen-Feng Cheng, Xing Xie, and Shou-De Lin. 2019.
\newblock Evaluating image-inspired poetry generation.
\newblock In \emph{CCF international conference on natural language processing and chinese computing}, pages 539--551. Springer.

\bibitem[{Wu et~al.(2023)Wu, Draws, Cau, Barile, Rieger, and Tintarev}]{wu2023explaining}
Zhangyi Wu, Tim Draws, Federico Cau, Francesco Barile, Alisa Rieger, and Nava Tintarev. 2023.
\newblock Explaining search result stances to opinionated people.
\newblock In \emph{World Conference on Explainable Artificial Intelligence}, pages 573--596. Springer.

\bibitem[{Xie et~al.(2017)Xie, Rastogi, and Chang}]{xie2017deep}
Stanley Xie, Ruchir Rastogi, and Max Chang. 2017.
\newblock Deep poetry: Word-level and character-level language models for shakespearean sonnet generation.
\newblock \emph{Natural Lang. Process. Deep Learn}.

\bibitem[{Xu et~al.(2021)Xu, Lei, Wang, Niu, Wu, and Che}]{xu2021enhancing}
Jun Xu, Zeyang Lei, Haifeng Wang, Zheng-Yu Niu, Hua Wu, and Wanxiang Che. 2021.
\newblock Enhancing dialog coherence with event graph grounded content planning.
\newblock In \emph{Proceedings of the Twenty-Ninth International Conference on International Joint Conferences on Artificial Intelligence}, pages 3941--3947.

\bibitem[{Xu and Mease(2009)}]{xu2009evaluating}
Ya~Xu and David Mease. 2009.
\newblock Evaluating web search using task completion time.
\newblock In \emph{Proceedings of the 32nd international ACM SIGIR conference on Research and development in information retrieval}, pages 676--677.

\bibitem[{Yang and Klein(2021)}]{Yang_2021}
Kevin Yang and Dan Klein. 2021.
\newblock \href {https://doi.org/10.18653/v1/2021.naacl-main.276} {Fudge: Controlled text generation with future discriminators}.
\newblock In \emph{Proceedings of the 2021 Conference of the North American Chapter of the Association for Computational Linguistics: Human Language Technologies}. Association for Computational Linguistics.

\bibitem[{Yedetore et~al.(2023)Yedetore, Linzen, Frank, and McCoy}]{yedetore2023poor}
Aditya Yedetore, Tal Linzen, Robert Frank, and R~Thomas McCoy. 2023.
\newblock How poor is the stimulus? evaluating hierarchical generalization in neural networks trained on child-directed speech.
\newblock In \emph{The 61st Annual Meeting Of The Association For Computational Linguistics}.

\bibitem[{Yin et~al.(2022)Yin, Wieting, Sil, and Neubig}]{yin-etal-2022-ingredients}
Pengcheng Yin, John Wieting, Avirup Sil, and Graham Neubig. 2022.
\newblock \href {https://doi.org/10.18653/v1/2022.acl-long.103} {On the ingredients of an effective zero-shot semantic parser}.
\newblock In \emph{Proceedings of the 60th Annual Meeting of the Association for Computational Linguistics (Volume 1: Long Papers)}, pages 1455--1474, Dublin, Ireland. Association for Computational Linguistics.

\bibitem[{Yu et~al.(2020)Yu, Bing, Zhang, Lam, and Si}]{yu-etal-2020-review}
Qian Yu, Lidong Bing, Qiong Zhang, Wai Lam, and Luo Si. 2020.
\newblock \href {https://doi.org/10.18653/v1/2020.acl-main.26} {Review-based question generation with adaptive instance transfer and augmentation}.
\newblock In \emph{Proceedings of the 58th Annual Meeting of the Association for Computational Linguistics}, pages 280--290, Online. Association for Computational Linguistics.

\bibitem[{Yue et~al.(2022)Yue, Yao, and Sun}]{yue-etal-2022-synthetic}
Xiang Yue, Ziyu Yao, and Huan Sun. 2022.
\newblock \href {https://doi.org/10.18653/v1/2022.acl-long.95} {Synthetic question value estimation for domain adaptation of question answering}.
\newblock In \emph{Proceedings of the 60th Annual Meeting of the Association for Computational Linguistics (Volume 1: Long Papers)}, pages 1340--1351, Dublin, Ireland. Association for Computational Linguistics.

\bibitem[{Zhang et~al.(2022)Zhang, Xing, Xia, Xu, and Zhu}]{zhang-et-al-2022-making}
Zejun Zhang, Zhenchang Xing, Xin Xia, Xiwei Xu, and Liming Zhu. 2022.
\newblock \href {https://doi.org/10.1145/3540250.3549143} {Making python code idiomatic by automatic refactoring non-idiomatic python code with pythonic idioms}.
\newblock In \emph{Proceedings of the 30th ACM Joint European Software Engineering Conference and Symposium on the Foundations of Software Engineering}, ESEC/FSE 2022, page 696–708, New York, NY, USA. Association for Computing Machinery.

\bibitem[{Zhang et~al.(2024)Zhang, Goldsack, Scarton, and Lin}]{zhang2024atlasimprovinglaysummarisation}
Zhihao Zhang, Tomas Goldsack, Carolina Scarton, and Chenghua Lin. 2024.
\newblock \href {https://arxiv.org/abs/2406.05625} {Atlas: Improving lay summarisation with attribute-based control}.
\newblock \emph{Preprint}, arXiv:2406.05625.

\bibitem[{Zheng et~al.(2024)Zheng, Ritter, and Xu}]{zheng-etal-2024-neo}
Jonathan Zheng, Alan Ritter, and Wei Xu. 2024.
\newblock \href {https://doi.org/10.18653/v1/2024.acl-long.749} {{NEO}-{BENCH}: Evaluating robustness of large language models with neologisms}.
\newblock In \emph{Proceedings of the 62nd Annual Meeting of the Association for Computational Linguistics (Volume 1: Long Papers)}, pages 13885--13906, Bangkok, Thailand. Association for Computational Linguistics.

\bibitem[{Zheng et~al.(2015)Zheng, Vydiswaran, Liu, Wang, Stubbs, Uzuner, Gururaj, Bayer, Aberdeen, Rumshisky et~al.}]{zheng2015ease}
Kai Zheng, VG~Vinod Vydiswaran, Yang Liu, Yue Wang, Amber Stubbs, {\"O}zlem Uzuner, Anupama~E Gururaj, Samuel Bayer, John Aberdeen, Anna Rumshisky, et~al. 2015.
\newblock Ease of adoption of clinical natural language processing software: an evaluation of five systems.
\newblock \emph{Journal of biomedical informatics}, 58:S189--S196.

\bibitem[{Zhong et~al.(2024)Zhong, Li, Wang, Song, Wei, Lian, and Mao}]{zhong-etal-2024-benchmarking}
Tianqi Zhong, Zhaoyi Li, Quan Wang, Linqi Song, Ying Wei, Defu Lian, and Zhendong Mao. 2024.
\newblock \href {https://doi.org/10.18653/v1/2024.acl-long.351} {Benchmarking and improving compositional generalization of multi-aspect controllable text generation}.
\newblock In \emph{Proceedings of the 62nd Annual Meeting of the Association for Computational Linguistics (Volume 1: Long Papers)}, pages 6486--6517, Bangkok, Thailand. Association for Computational Linguistics.

\bibitem[{Zhou et~al.(2023)Zhou, Zhu, Chen, Chen, Zhao, Chen, Lin, Wen, and Han}]{zhou2023dontmakellmevaluation}
Kun Zhou, Yutao Zhu, Zhipeng Chen, Wentong Chen, Wayne~Xin Zhao, Xu~Chen, Yankai Lin, Ji-Rong Wen, and Jiawei Han. 2023.
\newblock \href {https://arxiv.org/abs/2311.01964} {Don't make your llm an evaluation benchmark cheater}.
\newblock \emph{Preprint}, arXiv:2311.01964.

\end{thebibliography}

\end{document}